\documentclass[11pt]{article}
\pdfoutput=1

\usepackage{mathrsfs}
\usepackage{amsmath,amssymb}
\usepackage{bm}
\usepackage{natbib}
\usepackage[usenames]{color}
\usepackage{amsthm}

\usepackage{multirow} 
\usepackage{enumitem}

\allowdisplaybreaks

\usepackage[colorlinks,
linkcolor=red,
anchorcolor=blue,
citecolor=blue
]{hyperref}

\usepackage{mylatexstyle}

\usepackage{setspace}
\usepackage[left=1in, right=1in, top=1in, bottom=1in]{geometry}

\usepackage[textsize=tiny]{todonotes}
\title{\huge The Benefits of Mixup for Feature Learning}

\author
{
    Difan Zou\thanks{Department of Computer Science and Institute of Data Science, The University of Hong Kong, Hong Kong; e-mail:  {\tt dzou@cs.hku.hk}}
    ~~~and~~~
	Yuan Cao\thanks{Department of Statistics and Actuarial Science and Department of Mathematics, The University of Hong Kong, Hong Kong; e-mail:  {\tt yuancao@hku.hk}} 
	 ~~~and~~~
	Yuanzhi Li\thanks{Machine Learning Department, Carnegie Mellon University, Pittsburgh, PA, USA; e-mail: {\tt yuanzhil@andrew.cmu.edu}} 
	~~~and~~~
	Quanquan Gu\thanks{Department of Computer Science, University of California, Los Angeles, CA, USA; e-mail: {\tt qgu@cs.ucla.edu}}
}
\date{}

\date{}

\def\poly{\mathrm{poly}}

\def\polylog{\mathrm{polylog}}
\def\logit{\mathrm{Logit}}

\newtheorem{hypothesis}{Hypothesis}

\newcommand{\la}{\langle}
\newcommand{\ra}{\rangle}

\begin{document}

\maketitle

\begin{abstract}
Mixup, a simple data augmentation method that randomly mixes two data points via linear interpolation, has been extensively applied in various deep learning applications to gain better generalization. However, the theoretical underpinnings of its efficacy are not yet fully understood. In this paper, we aim to seek a fundamental understanding of the benefits of Mixup. We first show that Mixup using different linear interpolation parameters for features and labels can still achieve similar performance to the standard Mixup. This indicates that the intuitive linearity explanation in \citet{zhang2018mixup} may not fully explain the success of Mixup. Then we perform a theoretical study of Mixup from the feature learning perspective. We consider a feature-noise data model and show that Mixup training can effectively learn the rare features (appearing in a small fraction of data) from its mixture with the common features (appearing in a large fraction of data). In contrast, standard training can only learn the common features but fails to learn the rare features, thus suffering from bad generalization performance. Moreover, our theoretical analysis also shows that the benefits of Mixup for feature learning are mostly gained in the early training phase, based on which we propose to apply early stopping  in Mixup. Experimental results verify our theoretical findings and demonstrate the effectiveness of the early-stopped Mixup training.

\end{abstract}

\section{Introduction}

The Mixup method \citep{zhang2018mixup} is a popular data augmentation technique in deep learning, known to yield notable improvements in generalization and robustness across multiple domains, such as image recognition \citep{berthelot2019mixmatch}, natural language processing \citep{guo2019augmenting,chen2020mixtext}, and graph learning \citep{Han2022mixupgraph}. Unlike traditional data augmentation approaches that require domain knowledge of the dataset (e.g., random rotation and cropping for image data, and randomly modifying edges for graph data), Mixup relies on convex combinations of both features and labels from a pair of randomly selected training data points. As a result, this technique does not require any specialized knowledge or expertise to be performed.

Despite the remarkable empirical success of Mixup, there is a considerable gap in the theoretical understanding of this technique.  In the original work of Mixup \citep{zhang2018mixup}, it has been argued that the efficacy of Mixup can be attributed to its inductive bias, which encourages the trained model to behave linearly, leading to (relatively) simple decision boundaries. This inductive bias has been further supported by a series of works \citep{guo2019Mixup,zhang2020does,zhang2022and,chidambaram2021towards}, which prove that the Mixup behaves similarly to standard training for linear models.
In particular, Mixup applies the same linear interpolation on the features and labels of a pair of training data points $(\xb_1,y_1)$ and $(\xb_2,y_2)$: denoted by $\lambda\xb_1+(1-\lambda)\xb_2$ and labels $\lambda y_1 + (1-\lambda)y_2$, where $\lambda\in[0.5, 1]$ is randomly chosen. Then, the trained neural network (NN) model $F$ is naturally encouraged to conduct the mapping $F(\lambda\xb_1+(1-\lambda)\xb_2) \rightarrow \lambda y_1 + (1-\lambda)y_2$ for all $\lambda\in[0.5 ,1]$, $(\xb_1,y_1)$ and $(\xb_2,y_2)$, implying that $F$ tends to behave linearly at least within the line segments between all training data pairs. 

\begin{figure}[!t]
\vskip -0.1in
     \centering
     \subfigure[ResNet18]{\includegraphics[width=0.49\columnwidth]{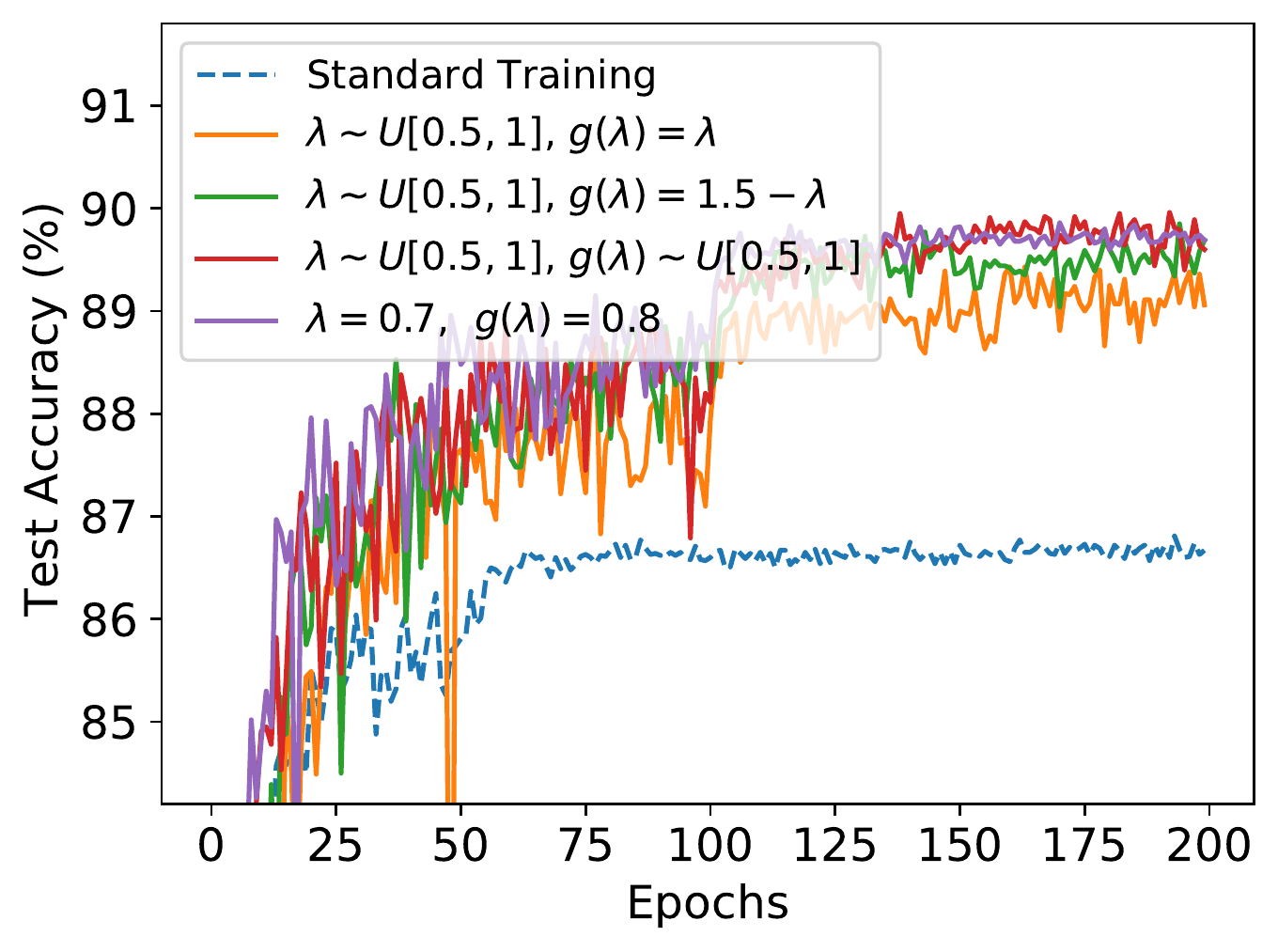}}
      \subfigure[VGG16]{\includegraphics[width=0.49\columnwidth]{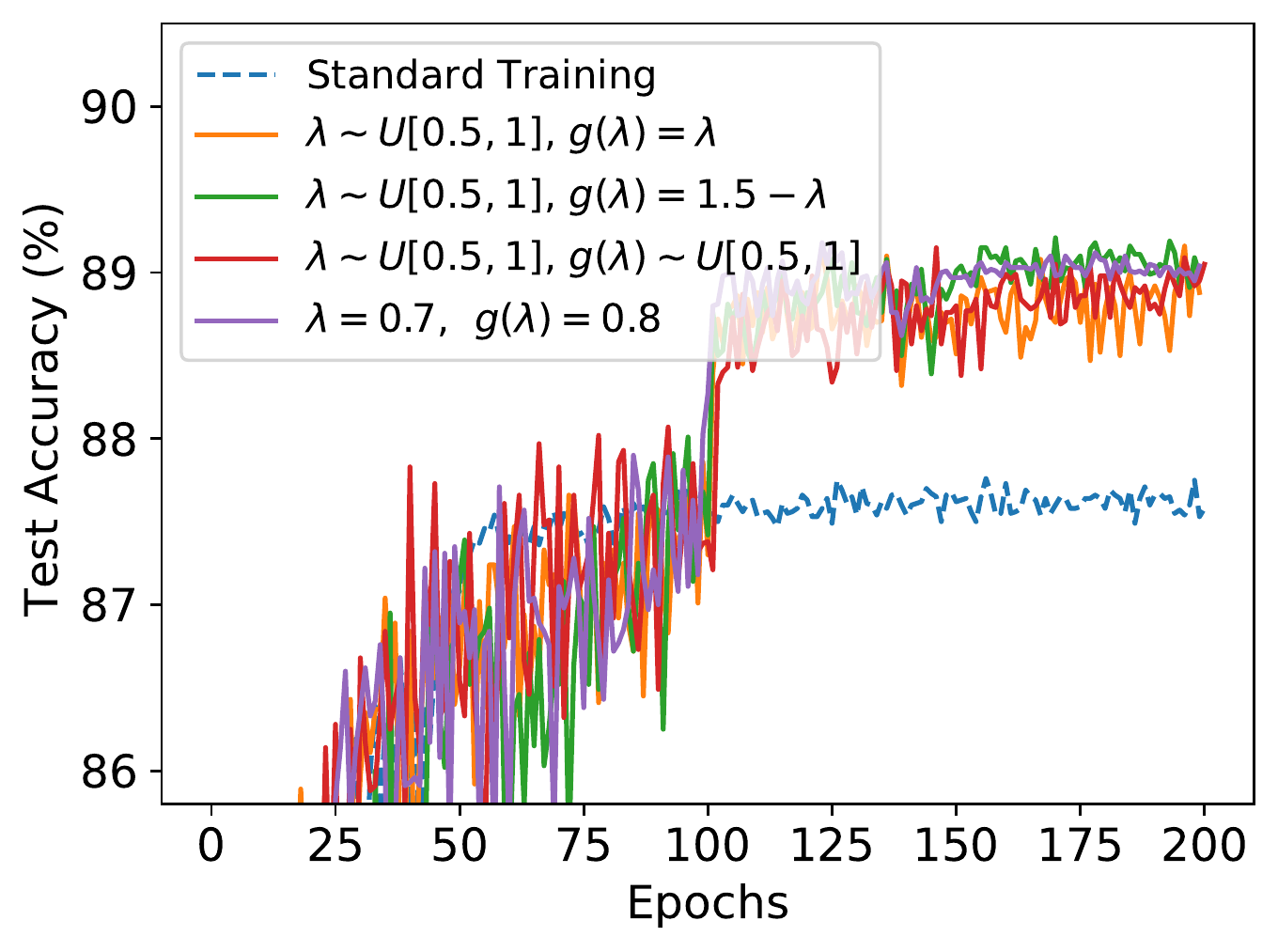}}
    \caption{Test accuracy achieved by Mixup training with different configurations of $\lambda$ and $g(\lambda)$. The results are evaluated by training ResNet18 and VGG16 on CIFAR-10 dataset without random crop \& flip data augmentation and weight decay regularization. We consider $5$ different configurations: (1) $\lambda=g(\lambda)=1$, i.e., standard training; (2) $\lambda = g(\lambda)\sim U[0.5, 1]$, i.e., standard Mixup; (3) $\lambda\sim U[0.5,1]$ and $g(\lambda)=1.5-\lambda$; (4) $\lambda\sim U[0.5,1]$ and $g(\lambda)\sim U[0.5,1]$; (5) $\lambda=0.7$ and $g(\lambda)=0.8$. It is clear that the performance gain of Mixup does not require setting $g(\lambda)=\lambda$. }
    \vspace{-2mm}\label{fig:linearity_negative}
\end{figure}

Although linearity is a nice inductive bias that tends to learn the models with low complexities, we are not clear about whether such an intuition from the algorithm design (i.e., performing the same linear interpolation for features and labels) can indeed explain the improvement in generalization.
To examine this, we conduct a proof-of-concept experiment on CIFAR-10 dataset. Instead of using the same linear interpolation in the feature and label space, we implement the interpolations using different $\lambda$'s for features and labels, i.e., we implement the  Mixup data augmentation on the features and labels as: $\lambda\xb_1+(1-\lambda)\xb_2$ and $g(\lambda)y_1 + [1-g(\lambda)]y_2$ for some nonlinear or even random function $g(\cdot):\RR^{[0.5, 1]}\rightarrow \RR^{[0.5,1]}$. Our results, shown in Figure \ref{fig:linearity_negative}, demonstrate that the substantial performance gain of Mixup training over standard training does not require $g(\lambda)=\lambda$. Other choices, such as fixed or independently random $\lambda$ and $g(\lambda)$, can lead to comparable or even better performance.

Therefore, it demands seeking a more fundamental understanding of Mixup that is beyond the linearization illustration. To address this issue, we draw inspiration from a recent work \citep{shen2022data}, which regards standard image data augmentation as a form of feature manipulation. This perspective offers a general framework to investigate the behavior of various data augmentation techniques, including Mixup in deep learning.  In particular, they consider a multi-view data model that consists of multiple feature vectors and noise vectors with different strengths and frequencies. More specifically, the feature vectors are categorized as the common ones (i.e., ``easy to learn'' features) and the rare ones (i.e., ``hard to learn'' features): the former refers to the feature appearing in a large fraction of data (thus contribute a lot to the gradient updates), and the latter refers to the features occurring in a small fraction of data (thus have limited contribution to the gradient). They further assume that the common features are the ones with rare orientations compared to the rare features and they can be balanced by applying data augmentations. For example, the common feature of a cow could be the left-facing cow, while the rare feature could be the right-facing cow, which can be generated by applying a horizontal flip to the common feature.

However, in many cases, the common and rare features may not be easily balanced by standard data augmentations. Let's still take the cow image as an example, the common and rare features could be brown cows and black cows, or front-view cows and side-view cows. Then the standard rotation or flip operations clearly cannot convert the common features to rare ones. We conjecture that Mixup may exhibit certain benefits in tackling this type of feature, as it has been shown to improve test accuracy when combined with standard data augmentations \citep{zhang2018mixup}. This motivates the problem setup considered in this study.

Particularly, we perform the theoretical study of the learning dynamics of Mixup based on a similar multi-view data model (see Definition \ref{def:data_distribution_new} for more details): each data point will either contain a common feature vector with a relatively high probability $1-\rho$, or a rare feature vector  with a relatively low probability $\rho$. The remaining components will be filled with random noise or feature noise. We then consider a two-layer convolutional neural network (CNN) model and study the learning behaviors of both standard training and Mixup training using gradient descent. The main contributions of this paper are highlighted as follows:

\begin{itemize}[leftmargin=*]
\item We identify that the linearity illustration may not be able to fully elucidate the exceptional performance of Mixup. In particular, we  show that using the same linear interpolations for both features and labels is not necessary, while some other choices, e.g., independently random linear interpolations, can also lead to substantial performance gains compared to standard training.

\item We prove a negative result (Theorem \ref{thm:std_training}) for standard training, demonstrating its inability to learn the rare features of the multi-view distribution. This failure leads to the domination of the rare feature data by its noise components during the test period, resulting in a $\Theta(\rho)$ test error. The reason for this lies in the tendency of the standard training algorithm to memorize the noise component of rare feature data to attain zero training error, while the rare feature itself, which appears in only a small fraction of the data, is not prominent enough to be effectively discovered by the algorithm.

\item More importantly, we establish a positive result (Theorem \ref{thm:Mixup_training}) for Mixup training by showcasing its ability to attain near-zero test errors on the multi-view distribution. Specifically, we demonstrate that Mixup can successfully mix the common and rare features so that the gradients along these two
features are correlated. As a result, the rare feature learning can be boosted by the fast learning
of common features, and ultimately reaches a sufficiently high level to overshadow the effects of noise on test data.

\item Our theory also suggests that the feature learning (especially the rare feature) benefits of Mixup are mostly gained in the early training phase. Then we develop the early-stopped Mixup, i.e., turning off the Mixup data augmentation after a certain number of iterations. Experimental results show that the test error achieved by early-stopped Mixup is comparable to or even better than that achieved by standard Mixup (i.e., using Mixup throughout the entire training). This not only corroborates our theoretical findings but also justifies the necessity to study the entire feature learning dynamics of Mixup rather than only the solution to the (equivalent) empirical risk of Mixup.

\end{itemize}

\paragraph{Notations.} We use $\poly(n)$ and $\polylog(n)$ to denote a polynomial function, with a sufficiently large (constant) degree, of $n$ or $\log(n)$ respectively. We use $o(1/\polylog(n))$ (and $\omega(\polylog(n))$) to denote some quantities that decrease (or grow) faster than $1/\log^c(n)$ (or $\log^c(n)$) for any constant $c$. We use $\tilde O$, $\tilde \Omega$, and $\tilde\Theta$ to hide some log factors in the standard Big-O, Big-Omega, and Big-Theta notations.



\section{Related Work}
\paragraph{Theoretical Analysis of Mixup.}
We would like to comment on some recent works that attempt to explain the benefits of Mixup from different angles. To name a few, \citet{thulasidasan2019mixup} showed that the models trained by Mixup are substantially better calibrated, i.e., the softmax logits are closer to the actual likelihood than that obtained by standard training. \citet{carratino2020mixup} studied the regularization effect of Mixup training and connected it to multiple known data-dependent regularization
schemes such as label smoothing. Following 
the same direction, \citet{parkunified} further developed a unified analysis for a class of Mixup methods, including the original one and CutMix \citep{yun2019cutmix}, and proposed a hybrid version of Mixup that achieves better test performance. \citet{chidambaram2021towards} studied the Mixup-optimal classifier and characterized its performance on original training data points. However, these works mostly focus on the solution to certain Mixup-version regularized empirical risk, while our experiments on early-stopped Mixup suggest that the entire learning dynamics could be more important.

Very recently, \citet{chidambaram2022provably} conducted feature learning-based analyses for Mixup and demonstrated its benefits. However, we would like to clarify some differences in our theoretical analysis. Firstly, in terms of the Mixup method, they considered only the mid-point Mixup, where $\lambda=g(\lambda)=0.5$, while we allow a more general choice of $\lambda\in(0.5, 1)$. Secondly, for the data model, they considered two features generated from a symmetric distribution for each class, along with feature noise, whereas we followed \citet{shen2022data} by considering a data model with two features of different frequencies (common and rare), feature noise, and random noise. Notably, the random noise component, which plays an important role in memorizing all training data points \citep{allen2020towards,shen2022data}, was ignored in \citet{chidambaram2022provably}. Finally,  their focus was on the competence between learning two symmetric features, while our focus was on the competence between rare feature learning and noise memorization. In conclusion, while \citet{chidambaram2022provably} and our work share a similar high-level spirit for understanding the benefits of Mixup, we approach this problem from different angles.

\paragraph{Data Augmentation.}
There are also many works studying the effect of standard data augmentation methods (i.e., performed within the data points) from different perspectives, such as regularization effect \citep{bishop1995training,dao2019kernel,pmlr-v119-wu20g}, algorithm bias \citep{hanin2021data},  margins \citep{rajput2019does}, model invariance \citep{chen2020group}, and feature learning \citep{shen2022data}. We view these works as orthogonal to our work as they mostly concern the data augmentation within the data points (e.g., random perturbation, random rotation, etc), which is different from the cross-data Mixup data augmentation.

\paragraph{Feature Learning in Deep Learning Theory.}
In the field of deep learning theory, there has emerged a series of works studying feature learning behavior during NN training. They focus on characterizing how different training approaches affect feature learning, such as ensembling \& knowledge distillation \citep{allen2020towards}, using adaptive gradients \citep{zou2021understanding}, mixture of expert \citep{chen2022towards}, and contrastive learning \citep{wen2021toward}. We point out that feature learning in Mixup is more complicated as the learning dynamics for different features are heavily coupled.

\section{Problem Setting.}\label{section:problemsetting}
As mentioned in the introduction section, we theoretically investigate the behaviors of standard training and Mixup training on a multi-view data model. In this section, we will first deliver a detailed set up of the multi-view data model and then introduce the two-layer CNN model as well as the gradient descent algorithms of standard training and Mixup training.

\subsection{Data Model}

In this work, we consider a binary classification problem on the data $(\xb,y)\in\RR^{dP}\times\{1,2\}$, where $\xb = (\xb^{(1)},\dots,\xb^{(P)})$ has $P$ patches and $y\in\{1,2\}$ denotes the data label. For ease of presentation, we define the data of label $y=1$ as the \textit{positive data} and the data of label $y=2$ as the \textit{negative data}. Moreover, the data will be randomly generated according to the following detailed process.

\begin{definition}\label{def:data_distribution_new}
Let $\cD$ denote the data distribution, from which a data point $(\xb,y)\in \RR^{d  P} \times \{1, 2\}$ is randomly generated as follows: 
\begin{enumerate}[leftmargin = *,nosep]
    \item Generate $y\in\{1, 2\}$ uniformly.
    \item Generate $\xb$ as a vector with $P$ patches $\xb = (\xb^{(1)},\ldots,\xb^{(2)})\in (\RR^d)^P$, where 
    \begin{itemize} [nosep, leftmargin=*]       
        \item \textbf{Feature Patch.}
        One patch, among all $P$ patches, will be randomly selected as the feature patch: with probability $1-\rho$ for some $\rho\in(0,1)$, this patch will contain a \textit{common feature}  ($\vb$ for positive data, $\ub$ for negative data); otherwise, this patch will contain a \textit{rare feature} ($\vb'$ for positive data, $\ub'$ for negative data). 
    
        \item \textbf{Feature Noise.} For all data, a feature vector from $\alpha\cdot\{\ub,\vb\}$ is randomly sampled and assigned to up to $b$ patches.
        \item \textbf{Noise patch.} The remaining patches (those haven't been assigned with a feature or feature noise) are random Gaussian noise  $\sim N(\boldsymbol{0}, \sigma_p^2\cdot \Hb)$, where $\Hb = \Ib - \frac{\ub\ub^\top}{\|\ub\|_2^2}-\frac{\vb\vb^\top}{\|\vb\|_2^2}-\frac{\vb'\vb'^\top}{\|\vb'\|_2^2}-\frac{\ub'\ub'^\top}{\|\ub'\|_2^2}$.
    \end{itemize}
\end{enumerate}
Without loss of generality, we assume all feature vectors are orthonormal, i.e., $\|\ab\|_2=1$ and $\la\ab,\bb\ra=0$ for all $\ab,\bb\in\{\vb,\ub,\vb',\ub'\}$ and $\ab\neq \bb$. 
Moreover, we set $d = \omega(n^6)$, $P,b = \polylog(n)$, $\rho = \Theta(n^{-3/4})$,  $\sigma_p = \Theta(d^{-1/2}n^{1/4})$, and $\alpha = \Theta(1/n)$\footnote{The choice of these parameters is not unique, here we only pick a feasible one for the ease of presentation.}.
\end{definition}


The multi-view model includes three types of critical vectors: common features, rare features, and noise vectors (the feature noise vectors can be categorized into common features since they are only different in terms of strength). 
All of them can be leveraged to fit the training data points and thus achieve a small training accuracy/loss. However, in order to achieve a nearly perfect test accuracy, one has to learn both common features and rare features as overfitting the random noise vectors of training data points will make no contribution or even be detrimental to the test performance, then the prediction will be heavily affected by the feature noise. Given the data model in Definition \ref{def:data_distribution_new}, we aim to show that Mixup is able to learn all informative features while standard training may only learn a part of them.

The feature-noise data model has been widely adopted to study many algorithmic aspects of deep learning, including adversarial training \citep{allen2020feature}, momentum \citep{jelassi2022towards},  ensemble and knowledge distillation \citep{allen2020towards}, benign overfitting \citep{cao2022benign}, and data augmentation \citep{shen2022data}.  Our data model mostly follows from the one considered in \citet{shen2022data}, which also includes the design of common features and rare features for studying the learning behaviors of data augmentation (that is performed within one single data point, e.g., random flip/rotation). However, instead of assuming that the rare features ($\vb'$ and $\ub'$) can be re-generated by applying data augmentation  on the common features ($\vb$ and $\ub$), we make nearly no assumption on their relationships. Therefore, learning the rare features in our model can be regarded as a harder problem,  and our theoretical analyses for Mixup are orthogonal to those in \citet{shen2022data}.


\subsection{Neural Network Function}

\paragraph{Two-layer CNN model.}
We consider a two-layer CNN model $F$ using quadratic activation function $\sigma(z) = z^2$. Note that we consider binary classification problem with $y\in\{1,2\}$, then given the input feature $\xb=(\xb^{(1)},\dots,\xb^{(p)})$, the $k$-th output of the network ($k\in\{1, 2\}$) is formulated as
\begin{align*}
F_k(\Wb;\xb)  =\sum_{p=1}^P\sum_{r=1}^m (\la\wb_{k,r},\xb^{(p)}\ra)^2.
\end{align*}
where  $\wb_{k,r}\in\RR^{d}$ denotes the neuron weight corresponding to the $k$-th output, $\Wb$ denotes the collection of all model weights, and $m$ denotes the NN width, which is set as $m=\polylog(n)$ throughout this paper\footnote{This choice of network width is to guarantee some nice properties hold with probability at least $1-1/\poly(n)$ at the initialization. We can also resort to setting $m$ as some large constant at the price of deriving a constant probability guarantee, e.g., $>0.9$.}.
Moreover, given the input $\xb$, we denote $\mathrm{Logit}_k(\Wb;\xb)$ by the logit of the $k$-th output of the NN model, which can be calculated via performing a softmax function on the NN outputs:
\begin{align*}
\textstyle{\mathrm{Logit}_k(\Wb;\xb) = e^{F_k(\Wb;\xb_i)}/\sum_{s\in\{1,2\}} e^{F_s(\Wb,\xb_i)}}.
\end{align*}
Using a polynomial activation function (or ReLU with polynomial smoothing) is not new in deep learning theory. The purpose is to better illustrate/distinguish the feature and noise learning dynamics during the neural network training  \citep{frei2022benign,cao2022benign,shen2022data,glasgow2022max}. Our analysis can also be extended to other polynomial functions $\sigma(x)=x^q$ for some $q> 1$.


\subsection{Training Algorithms}\label{sec:training_algorithms}
\paragraph{Initialization.} We assume that the initial weights of the neural network model are generated i.i.d. from the Gaussian initialization: $\wb_{k,r}^{(0)}\sim N(\boldsymbol{0}, \sigma_0^2\Ib)$, where $\sigma_0 = o(d^{-1/2})$.

\paragraph{Standard training.}
Given the training data points $\cS:=\{(\xb_i,y_i)\}_{i=1,\dots,n}$, we train the neural network model via applying standard full-batch gradient descent to optimize the following empirical risk function:
\begin{align*}
&L_\cS(\Wb) = \frac{1}{n}\sum_{i=1}^n \ell(\Wb;\xb_i,y_i),\quad\text{where}\quad 
\ell(\Wb;\xb_i,y_i) = -\log \frac{e^{F_{y_i}(\Wb,\xb_i)}}{\sum_{k\in\{1,2\}} e^{F_k(\Wb;\xb_i)}}.
\end{align*}
Starting from the initialization $\Wb^{(0)}$, the gradient descent of the standard training takes the following update step
\begin{align}\label{eq:update_std}
\Wb^{(t+1)}  = \Wb^{(t)} - \frac{\eta}{n}\sum_{i=1}^n \nabla_{\Wb} \ell(\Wb^{(t)};\xb_i,y_i),
\end{align}
where $\eta$ is the learning rate.
Then, the detailed calculation of the partial derivative $\nabla_{\wb_{k,r}} \ell(\Wb;\xb_i,y_i)$ is given by
\begin{align*}
\nabla_{\wb_{k,r}} \ell(\Wb;\xb_i,y_i) = -2\ell_{k,i}\cdot\sum_{p=1}^P\la\wb_{k,r},\xb_i^{(p)}\ra\cdot\xb_i^{(p)}.
\end{align*}
where $\ell_{k,i}=\ind_{k=y_i}-\mathrm{Logit}_k(\Wb^{(t)};\xb_i)$.

\textbf{Mixup Training.}
Given two training data points $(\xb_1,y_1)$ and $(\xb_2,y_2)$, Mixup trains a neural network based on 
the convex combinations of them: $(\lambda \xb_1 + (1-\lambda)\xb_2, \lambda y_1 + (1-\lambda) y_2)$ and $((1-\lambda) \xb_1 + \lambda\xb_2, (1-\lambda) y_1 + \lambda y_2)$, where we slightly abuse the notation by viewing the labels $y_1$ and $y_2$ as their one-hot encoding. Besides, Figure \ref{fig:linearity_negative} suggested that $\lambda$ does not need to be randomly sampled to achieve better performance than standard training, we will focus on a fixed constant $\lambda\in(0.5,1)$ in our theoretical analysis.
Finally, if considering all possible combinations of the training data pairs with a fixed $\lambda$, the (equivalent) training dataset of Mixup training is $
\cS_{\mathrm{Mixup}}: = \{\xb_{i,j}, y_{i,j}\}_{i, j\in[n]}$,
where we denote $\xb_{i,j}$ and $y_{i,j}$ by $\lambda\xb_i+(1-\lambda)\xb_j$ and $\lambda y_i+(1-\lambda)y_j$ respectively. Motivated by this, we can claim that the Mixup training actually aims to learn the model parameter by optimizing the following loss function:
\begin{align}\label{eq:trainingloss_Mixup}
L_{\cS}^{\mathrm{Mixup}}(\Wb) = \frac{1}{n^2}\sum_{i,j\in[n]} \ell(\Wb; \xb_{i,j}, y_{i,j}),
\end{align}
where 
\begin{align*}
\ell(\Wb;\xb_{i,j},y_{i,j}) = \lambda \ell(\Wb;\xb_{i,j}, y_i) + (1-\lambda)\ell(\Wb;\xb_{i,j}, y_j).
\end{align*}

In this paper, in order to better illustrate the key aspect of Mixup training as well as simplify the theoretical analysis, we resort to the gradient descent on the loss function \eqref{eq:trainingloss_Mixup}, which takes the following update step:
\begin{align*}
\Wb^{(t+1)} = \Wb^{(t)} - \frac{\eta}{n^2}\sum_{i=1}^n \sum_{j=1}^n\nabla_{\Wb} \ell(\Wb^{(t)};\xb_{i,j},y_{i,j}).
\end{align*}
Then, the detailed calculations of all partial derivatives are given as follows: for any Mixup data $(\xb_{i,j},y_{i,j})$, we have
\begin{align*}
\nabla_{\wb_{k,r}} \ell(\Wb;\xb_{i,j})
    &= 2\ell_{k,(i,j)}\cdot \sum_{p=1}^P\la\wb_{k,r}, \xb_{i,j}^{(p)}\ra\cdot\xb_{i,j}^{(p)},
\end{align*}
where $\ell_{k,i}$ is the loss derivative with respect to the network output $F_k(\Wb; \xb_{i,j}, y_{i,j})$:
\begin{align*}
\ell_{k,(i,j)}=\lambda\ind_{k=y_i}+(1-\lambda)\ind_{k=y_j}-\mathrm{Logit}_k(\Wb; \xb_{i,j}).
\end{align*}

    

\section{Main Theory}
In this section, we will theoretically characterize the generalization errors achieved by standard training and Mixup training on the multi-view model. In particular, the following Theorem states the negative result of standard training. 

\begin{theorem}\label{thm:std_training}
Suppose that the training data are generated according to Definition \ref{def:data_distribution_new}, let $\eta=1/\poly(n)$,  $T=\polylog(n)/\eta$, and $\{\Wb_{\mathrm{standard}}^{(t)}\}_{t=0,\dots,T}$ be the iterates of standard training, then with probability at least $1-1/\poly(n)$, it holds that for all $t\in[0,T]$, $\PP_{(\xb,y)\sim\cD}\big[\argmax_kF_k(\Wb_{\mathrm{standard}}^{(t)};\xb)\neq y\big]\ge \frac{\rho}{2.01}$.
\end{theorem}
Theorem \ref{thm:std_training} basically states that the two-layer CNN model obtained via standard training will lead to at least $\Theta(\rho)$ test error on the data model defined in Definition \ref{def:data_distribution_new}. In fact, as we will clarify in Section \ref{sec:proof_sketch_std_train}, this is due to the fact that the rare feature data will be fitted via their random noise components, while the rare features $\vb'$ and $\ub'$ will not be learned. Consequently, nearly a half of test rare feature data will be misled by the feature noise components, resulting in a $\Theta(\rho)$ test error.  

In comparison, Mixup training can help learn the rare features and thus achieve a smaller generalization error. We formally state this result in the following theorem.
\begin{theorem}\label{thm:Mixup_training}
Suppose the training data are generated according to Definition \ref{def:data_distribution_new}, let $\eta=\frac{1}{\poly(n)}$,  $T=\frac{\polylog(n)}{\eta}$, and $\{\Wb_{\mathrm{Mixup}}^{(t)}\}_{t=0,\dots,T}$ be the iterates of Mixup training, then with probability at least $1-\frac{1}{\poly(n)}$, it holds that for some $t\in[0,T]$, $\PP_{(\xb,y)\sim\cD}\big[\argmax_kF_k(\Wb_{\mathrm{Mixup}}^{(t)};\xb)\neq y\big]= o\big(\frac{1}{\poly(n)}\big)$.
\end{theorem}

Theorem \ref{thm:Mixup_training} shows that the two-layer CNN model obtained via Mixup training can achieve nearly zero test error, which is much better than that of standard training as $\rho=\Theta(n^{-3/4})\gg o(1/\poly(n))$ (see Definition \ref{def:data_distribution_new}). In particular, as we will show in Section \ref{sec:proof_sketch_mixup_train}, at the core of Mixup training is that it mixes common features and rare features together, thus the learning of these two types of features will be coupled. Consequently, the learning of rare features will be ``boosted'' by the learning of common features, reaching a sufficiently large level that dominates the effect of feature noise.

\section{Overview of the Analysis}

According to the data model in Definition \ref{def:data_distribution_new}, the critical step of the generalization analysis for standard training and Mixup training is to sharply characterize the magnitude of the feature learning, including both common features ($\vb$ and $\ub$) and rare features ($\vb'$, $\ub'$), as well as the noise learning, including all noise vectors $\bxi_i^{(p)}$'s (denoted by $\{\bxi\}$). Then, the key step to show the generalization gap between standard training and Mixup training is to identify their difference in terms of feature and noise learning. 

\subsection{Feature and Noise Learning of Standard Training}\label{sec:proof_sketch_std_train}
According to Definition \ref{def:data_distribution_new}, we define $\cS_0^{+}$ and $\cS_0^{-}$ as the set of training data that have strong positive and negative features respectively and $\cS_1^{+}$ and $\cS_1^{-}$  as the set of data that have weak positive and negative features respectively. In the following, the learning patterns of these vectors will be characterized by studying the inner products $\la\wb_{k,r}^{(t)},\ab\ra$, where $\ab\in\{\vb,\ub,\vb',\ub'\}\cup\{\bxi\}$. Intuitively, a larger inner product implies that the neural network has a stronger learning ability of $\ab$. Given the multi-view data model in Definition \ref{def:data_distribution_new} and the update rule \eqref{eq:update_std}, we have for any $\ab\in\{\vb,\ub,\vb',\ub'\}\cup\{\bxi\}$, 
\begin{align}\label{eq:update_features_main}
\la\wb_{k,r}^{(t+1)},\ab\ra & =
\la\wb_{k,r}^{(t)},\ab\ra +\frac{2\eta}{n}\cdot\sum_{i\in[n]} \ell_{k,i}^{(t)} \sum_{p=1}^P \la\wb_{k,r}^{(t)},\xb_i^{(p)}\ra\cdot \la\xb_i^{(p)},\ab\ra.
\end{align}
Then by the data model in Definition \ref{def:data_distribution_new}, we can see that for common feature vector $\ab\in\{\vb,\ub\}$, there will be $\Theta(n)$ training data points contributing  to the learning of $\ab$; while for rare feature vector $\ab\in\{\vb',\ub'\}$, only $\Theta(\rho n)$ data points contributing to the learning. Besides, since each noise vector $\ab\in\{\bxi\}$ in the training data point is randomly generated, its learning will largely rely on one single data, i.e., the data consisting of that noise vector. This difference clearly shows that the common features will be  preferably discovered and learned during the standard training. 

In the following analysis, we will decompose the entire standard training process into three phases, according to the learning of common features and noises. In particular, the \textbf{Phase 1} referred to the initial training iterations such that the neural network output, with respect to all input training data, is in the order of $O(1)$. In this phase, the loss derivatives $\ell_i^{(t)}$ will remain in the constant order and all critical vectors will be learned at a fast rate. Then 
The \textbf{Phase 2} is defined as the training period starting from the end of \textbf{Phase 1} to the iteration that the neural network output has reached $\tilde\Theta(1)$ for all training inputs. Finally, we refer to \textbf{Phase 3} as the training period starting from the end of \textbf{Phase 2} to convergence, i.e., the gradient converges to zero.


\paragraph{Standard Training, Phase 1.}
The following lemma characterizes the learning of all features and noise  in Phase 1.
\begin{lemma}\label{lemma:standard_learning_phase1_main}
There exists a iteration number $T_0 = \tilde\Theta(1/\eta)$ such that for any $t\le T_0$, it holds that
\begin{align}\label{eq:good_innerproduct}
\la\wb_{1,r}^{(t+1)},\vb\ra  &= \la\wb_{1,r}^{(t)},\vb\ra\cdot \big(1 + \Theta(\eta)\big),\quad
\la\wb_{2,r}^{(t+1)},\ub\ra  = \la\wb_{2,r}^{(t)},\ub\ra\cdot \big(1 + \Theta(\eta)\big).
\end{align}
Besides, for all remaining inner products, it holds that
\begin{align*}
\max_{r}|\la\wb_{k,r}^{(t+1)},\ab\ra| \le \max_{r}|\la\wb_{k,r}^{(t)},\ab\ra|\cdot \big[1 + o(\eta/\polylog(n))\big]
\end{align*}
where $t\le T_0$, $r\in[m]$, $k\in[2]$, $q\in[P]$, $\ab\in\{\ub,\vb,\ub',\vb'\}\cup\{\bxi\}$ are arbitrarily chosen as long as the inner products are different from those in \eqref{eq:good_innerproduct}.
\end{lemma}
Lemma \ref{lemma:standard_learning_phase1_main} shows the competence results of learning common features, rare features, and noise vectors in Phase 1. In particular, it can be observed that the learning of common features ($\vb$, $\ub$) enjoys a much faster rate, while other critical vectors, including rare features and noise vectors, will be staying at their initialization levels. 

\paragraph{Standard Training, Phase 2.}
During this phase, the loss derivative will remain in the constant order for the rare feature data, since either the rare feature learning (e.g, $\la\wb_{1,r}^{(t)}, \vb'\ra$) or the noise learning (e.g., $\la\wb_{1,r}^{(t)},\bxi_i^{(p)}\ra$) are still quite small. Recall that the common features have already been fitted during Phase 1, we will then focus on the competence between learning rare features and learning noise vectors in Phase 2. The following lemma characterizes the dynamics of standard training in Phase 2.

\begin{lemma}\label{lemma:standard_learning_phase2_main}
There exists a iteration number $T_1=\tilde O\big(\frac{n}{d\sigma^2 \eta}\big)$ such that for any $t\in[T_0, T_1]$, it holds that
\begin{align*}
\la\wb_{1,r}^{(t+1)},\vb'\ra &= \la\wb_{1,r}^{(t)},\vb'\ra\cdot \big[1 + \Theta(\rho\eta)\big],\quad
\la\wb_{2,r}^{(t+1)},\ub'\ra = \la\wb_{2,r}^{(t)},\ub'\ra\cdot \big[1 + \Theta(\rho\eta)\big].
\end{align*}
Besides, for any $i\in\cS_1^+\cup\cS_1^-$, any $q\in[P]$ and $k=y_s$,
\begin{align*}
\max_{r}|\la\wb_{k,r}^{(t+1)},\bxi_s^{(q)}\ra| =
\max_{r}|\la\wb_{k,r}^{(t)},\bxi_s^{(q)}\ra|\cdot \big[1 + \eta/n\cdot\tilde\Theta(d\sigma_p^2)\big]
\end{align*}
\end{lemma}
Lemma \ref{lemma:standard_learning_phase2_main} shows that for rare feature data points, standard training admits a faster noise learning speed compared to rare feature learning (note that $d\sigma_p^2\gg \rho$, according to Definition \ref{def:data_distribution_new}). This consequently leads to adequate learning of noise ($|\la\wb_{y_i,r}^{(T_1)}, \bxi_i^{(p)}\ra|=\tilde\Theta(1)$ for some $p\in[P]$) and nearly no learning of rare features ($|\la\wb_{k,r}^{(T_1)},\vb'\ra|,|\la\wb_{k,r}^{(T_1)},\ub'\ra|=\tilde O\big(\sigma_0\big)$).

\paragraph{Standard Training, Final Phase.}
The final phase is defined as the training period after the end of Phase 2 until convergence. In the following lemma, we will show that (1) the convergence can be guaranteed; and (2) the learning of features and noise vectors at  Phase 2 will be maintained.

\begin{lemma}\label{lemma:standard_learning_phase3_main}
Let $T_1$ be the iteration number defined in Lemma \ref{lemma:standard_learning_phase2_main}, then for any $t=\poly(n)>T_1$ and $k\in\{1,2\}$, 
\begin{align*}
\frac{1}{n}\sum_{\tau=T_1}^t\sum_{i=1}^n |\ell_{k,i}^{(\tau)}| = \tilde O(1/\eta).
\end{align*}
Moreover, we have $\sum_{r=1}^m(\la\wb_{1,r}^{(t)}, \vb\ra)^2, \sum_{r=1}^m(\la\wb_{2,r}^{(t)}, \ub\ra)^2 = \tilde\Theta(1)$ and $|\la\wb_{k,r}^{(t)},\vb'\ra|, |\la\wb_{k,r}^{(t)},\ub'\ra|=\tilde O(\sigma_0)$.
\end{lemma}

It can be clearly seen that the gradient descent can converge to the point with a small gradient (the averaged loss derivative will be roughly in the order of $\tilde O(1/(t\eta))$, which approaches zero when  $t$ is large). More importantly, the common feature data and rare feature data will be correctly classified by fitting different components: common feature data will be fitted by learning $\vb$ and $\ub$, while the rare feature data will be fitted by noise memorization (as standard training nearly makes no progress in learning. Consequently, when it comes to a fresh test rare feature data, the model prediction will be heavily affected by the feature noise component, thus leading to an incorrect prediction with a constant probability (the formal proof is deferred to Section \ref{sec:proof_main_std}).

\subsection{Feature and Noise Learning of Mixup Training}\label{sec:proof_sketch_mixup_train}

As mentioned in Section \ref{sec:training_algorithms}, any data pair sampled  from training dataset will be considered, which gives in total $n^2$ Mixup data.  
Note that we have two types of data in the origin training dataset: common feature data and rare feature data with two labels, denoted by $\cS_0^+$, $\cS_0^-$, $\cS_1^+$, and $\cS_1^-$ (see  Section \ref{sec:proof_sketch_std_train}),  we can also categorize the Mixup data points into multiple sets accordingly. Particularly, let $\cS_{*,**}^{\dagger, \dagger\dagger}$ be the set of mixed data $\xb_{i,j}=\lambda\xb_i+(1-\lambda)\xb_j$ with $\xb_i\in\cS_*^\dagger$ and $\xb_j\in\cS_{**}^{\dagger\dagger}$, we can accordingly categorize all Mixup data with the following $4$ classes:
\begin{itemize}[leftmargin=*,nosep]
    \item Mix between two common feature data points, including $\cS_{0,0}^{+, +}$, $\cS_{0,0}^{-, -}$, $\cS_{0,0}^{+, -}$, $\cS_{0,0}^{-, +}$, each of them is of size $\Theta(n^2)$.
    \item Mix between  common feature and rare feature data points with the same label, including  $\cS_{0,1}^{+, +}$, $\cS_{0,1}^{-, -}$, $\cS_{1,0}^{+,+}$, and $\cS_{1,0}^{-,-}$, each of them is of size $\Theta(\rho n^2)$.
    \item Mix between common feature  and rare feature data points with different labels, including  $\cS_{0,1}^{+, -}$, $\cS_{0,1}^{-, +}$, $\cS_{1, 0}^{+, -}$, and $\cS_{1, 0}^{-, +}$, each of them is of size $\Theta(\rho n^2)$.
    \item Mix between two rare feature data points, including $\cS_{1,1}^{+, +}$,$\cS_{1,1}^{-, -}$, $\cS_{1,1}^{+, -}$ and $\cS_{1, 1}^{-, +}$, each of them is of size $\Theta(\rho^2 n^2)$.
    
\end{itemize}
In contrast to standard training that nearly admits  separate learning dynamics for common and rare features, the second and third classes of Mixup training data points, actively mix the common  and rare features together. For instance, some data points in $\cS_{0,1}^{+,+}$ will contain a data patch of form $\lambda\vb + (1-\lambda)\vb'$. Then the learning of $\vb$ will benefit the learning of $\vb'$, since their gradient updates are positively correlated. In the following, we will provide a precise characterization on the learning dynamics of feature and noise vectors.


In particular, noting that we consider the full-batch gradient descent on the entire Mixup training dataset (see Section \ref{sec:training_algorithms}), the update formula of all critical vectors are provided as follows: for any $\ab\in\{\ub,\vb,\ub',\vb'\}\cup\{\bxi\}$, we have
\begin{align}\label{eq:update_allfeatures_Mixup}
\la\wb_{k,r}^{(t+1)}, \ab\ra &= \la\wb_{k,r}^{(t)}, \ab\ra - \eta\cdot\la\nabla_{\wb_{k,r}} L(\Wb^{(t)}),\ab\ra.
\end{align}
where we denote $L(\Wb^{(t)})$ as the short-hand notation of $L_\cS^{\mathrm{Mixup}}$ (defined in \eqref{eq:trainingloss_Mixup}) for simplifying the notation.
More specifically, we summarize the update of all critical vectors (e.g., common features, rare features, and data noise vectors) in the following Proposition.
\begin{proposition}\label{prop:update_Mixup_main}
For any critical vector $\ab\in\{\vb, \ub, \vb', \ub'\}\cup\{\bxi\}$, we have
\begin{align*}
-\la\nabla_{\wb_{k,r}} L(\Wb^{(t)}),\ab\ra = \hspace{-4mm}\sum_{\bb\in\{\vb,\ub,\vb',\ub'\}\cup\{\bxi\}}\gamma_k^{(t)}(\bb,\ab)\la\wb_{k,r}^{(t)},\bb\ra
\end{align*}
where $\gamma_k^{(t)}(\bb,\ab)$ is a scalar output function that depends on $\bb, \ab\in\{\vb, \ub, \vb', \ub'\}\cup\{\bxi\}$. More specifically, let 
\begin{align*}
\xb_{i,j}^{(p)} &= \theta_{i,j}^{(p)}(\vb) \cdot\vb + \theta_{i,j}^{(p)}(\ub)\cdot \ub +\theta_{i,j}^{(p)}(\vb') \cdot\vb'  + \theta_{i,j}^{(p)}(\ub')\cdot \ub'+ \sum_{s=1}^n\sum_{q\in[P]}\theta_{i,j}^{(p)}(\bxi_s^{(q)})\cdot \bxi_s^{(q)} 
\end{align*}
be a linear expansion of $\xb_{i,j}^{(p)}$ on the space spanned by $\{\vb, \ub, \vb', \ub'\}\cup\{\bxi\}$, we have
\begin{align*}
\gamma_k^{(t)}(\bb, \ab) = \frac{1}{n^2}\sum_{i,j\in[n]}\ell_{k,(i,j)}^{(t)} \sum_{p\in[P]} \theta_{i,j}^{(p)}(\bb)\cdot\la\xb_{i,j}^{(p)},\ab\ra.
\end{align*}

\end{proposition}
\vspace{-2mm}

From Proposition \ref{prop:update_Mixup_main}, it can be seen that the learning of common features, rare features,  and noise vectors are heavily coupled. 
Mathematically, the  coefficient $\gamma_k^{(t)}(\ab,\bb)$ precisely describes how the learning of $\ab$ affects the learning of $\bb$, where $\ab,\bb\in\{\vb,\ub,\vb',\ub'\}\cup\{\bxi\}$.
This effect can be either positive or negative, depending on the sign of $\gamma_k^{(t)}(\ab,\bb)$. Then, the next step is to sharply characterize the coefficients $\gamma_k^{(t)}(\bb, \ab)$. We will focus on early phase of Mixup training, where the loss derivatives can be regarded as the constant (i.e., approximately $0.5$, $-0.5$, $\lambda-0.5$, or $0.5-\lambda$). Particularly, we will consider the training stage such that $\max_{k\in[2],i,j\in[n]} |F_k(\Wb^{(t)};\xb_{i,j})|\le \zeta$, where $\zeta=o\big(\frac{1}{\polylog(n)}\big)$ is a user-defined parameter. Then based on $\zeta$, we summarize the results of some critical coefficients in the following lemma, while the results for all coefficients are presented in Lemma \ref{lemma:feature_learning_coefficients_v_mixup}-\ref{lemma:noise_learning_coefficients_incorrect_mixup}.

\begin{lemma}\label{lemma:critical_coefficients}
Assume $\max_{k\in[2],i,j\in[n]} |F_k(\Wb^{(t)};\xb_{i,j})|\le \zeta$ for some $\zeta\in[\omega(d\sigma_p^2/(Pn)),o(d^{-1/2}\sigma_p^{-1})]$, then,
\begin{align*}
&\gamma_1^{(t)}(\vb,\vb), \gamma_2^{(t)}(\ub,\ub) = \Theta(1),\ \gamma_{y_i}^{(t)}(\bxi_i^{(p)}, \bxi_i^{(p)})=\Theta\big(d\sigma_p^2/n\big),\notag\\
&\gamma_1^{(t)}(\vb,\vb'), \gamma_2^{(t)}(\ub,\ub') = \Theta(\rho/P),\ |\gamma_2^{(t)}(\ub,\vb')|, |\gamma_1^{(t)}(\vb,\ub' )|=O(\zeta\rho/P).
\end{align*}
\end{lemma}
\vspace{-2mm}
The coefficients presented in Lemma \ref{lemma:critical_coefficients} reveal some key differences between learning common features, rare features, and noise. Let's consider $\vb$ without loss of generality. First, similar to the standard training, the learning of common features is much faster than the learning of noises, since the leading terms of common feature learning (i.e., $\gamma_1^{(t)}(\vb,\vb)$) and noise learning (i.e., $(\gamma_{y_i}^{(t)}(\bxi_i^{(p)}, \bxi_i^{(p)})$) satisfy:
$\gamma_1^{(t)}(\vb,\vb)\gg\gamma_{y_i}^{(t)}(\bxi_i^{(p)}, \bxi_i^{(p)})$. Second, different from standard training where the rare features are nearly unexplored, Mixup training has the ability to boost the learning of rare features via common feature learning, which is characterized by $\gamma_1^{(t)}(\vb, \vb')\cdot \la\wb_{1,r}^{(t)},\vb\ra$ or $\gamma_2^{(t)}(\ub, \vb')\cdot \la\wb_{2,r}^{(t)},\ub\ra$. Finally, we also show that such a boosting effect is positive: the  boosting of $\vb'$ to the correct neurons (i.e., $\{\wb_{1,r}^{(t)}\}_{r\in[m]}$) is stronger than that to the incorrect neurons (i.e., $\{\wb_{2,r}^{(t)}\}_{r\in[m]}$), since $\gamma_1^{(t)}(\vb,\vb')\gg|\gamma_2^{(t)}(\ub,\vb')|$ (recall we pick $\zeta = o\big(\frac{1}{\polylog(n)}\big)$). This implies that the rare features will be effectively discovered by  Mixup training, and finally, the neural network will have non-negligible components along the directions of $\vb'$ and $\ub'$. We formally stated this in the following lemma. 

\begin{lemma}\label{lemma:outcome_Mixup_main}
Let $\zeta$ be the same as that in Lemma \ref{lemma:critical_coefficients} and $T$ be the smallest iteration number such that $\max_{k\in[2], i,j\in[n]} |F_k(\Wb^{(T)}; \xb_{i,j})|\ge \zeta/2$, then $T=\tilde O(1/\eta)$ and  with probability at least $1-1/\poly(n)$, it holds that
\begin{align*}
&\max_{r}|\la\wb_{1,r}^{(T)},\vb\ra|, \max_{r}|\la\wb_{2,r}^{(T)},\ub\ra| = \tilde\Omega(\zeta^{1/2}), \ \max_{r}|\la\wb_{1,r}^{(T)},\vb'\ra|, \max_{r}|\la\wb_{2,r}^{(T)},\ub'\ra| =\Omega(\rho \zeta^{1/2})\notag\\
&\max_{r}|\la\wb_{2,r}^{(T)},\vb\ra|, \max_{r}|\la\wb_{1,r}^{(T)},\ub\ra| = \tilde O(\zeta^{3/2}),\ \max_{r}|\la\wb_{2,r}^{(T)},\vb'\ra|, \max_{r}|\la\wb_{1,r}^{(T)},\ub'\ra| =\tilde O(\zeta^{3/2}).
\end{align*}
\end{lemma}
\vspace{-2mm}
We can then make a comparison between Lemma \ref{lemma:standard_learning_phase3_main} and Lemma \ref{lemma:outcome_Mixup_main} to illustrate the similarities and differences between standard training and Mixup training in feature learning. In particular, it is clear that both standard and Mixup training can successfully learn the common features, i.e., the inner products $\la\wb_{1,r}^{(T)},\vb\ra$ and $\la\wb_{2,r}^{(T)},\ub\ra$ are the dominating ones among all critical inner products. While more importantly, the Mixup training can lead to much better rare feature learning compared to standard training: the standard training gives $|\la\wb_{1,r}^{(t)},\vb\ra|,|\la\wb_{2,r}^{(t)},\ub\ra| = \tilde O(\sigma_0)$ for all iterations; in contrast, the Mixup training gives $|\la\wb_{1,r}^{(T)},\vb'\ra|, |\la\wb_{2,r}^{(T)},\ub'\ra|=\Omega(\rho\zeta^{1/2})$, which are much larger. Consequently, the strength of rare feature learning in Mixup training will dominate the effect of feature noise, thus achieving a nearly zero test error (the formal proof is deferred to Section \ref{sec:proof_mixup}).


\begin{figure}[!t]
\vskip -0.1in
     \centering
     \subfigure[Common Feature Learning]{\includegraphics[width=0.48\columnwidth]{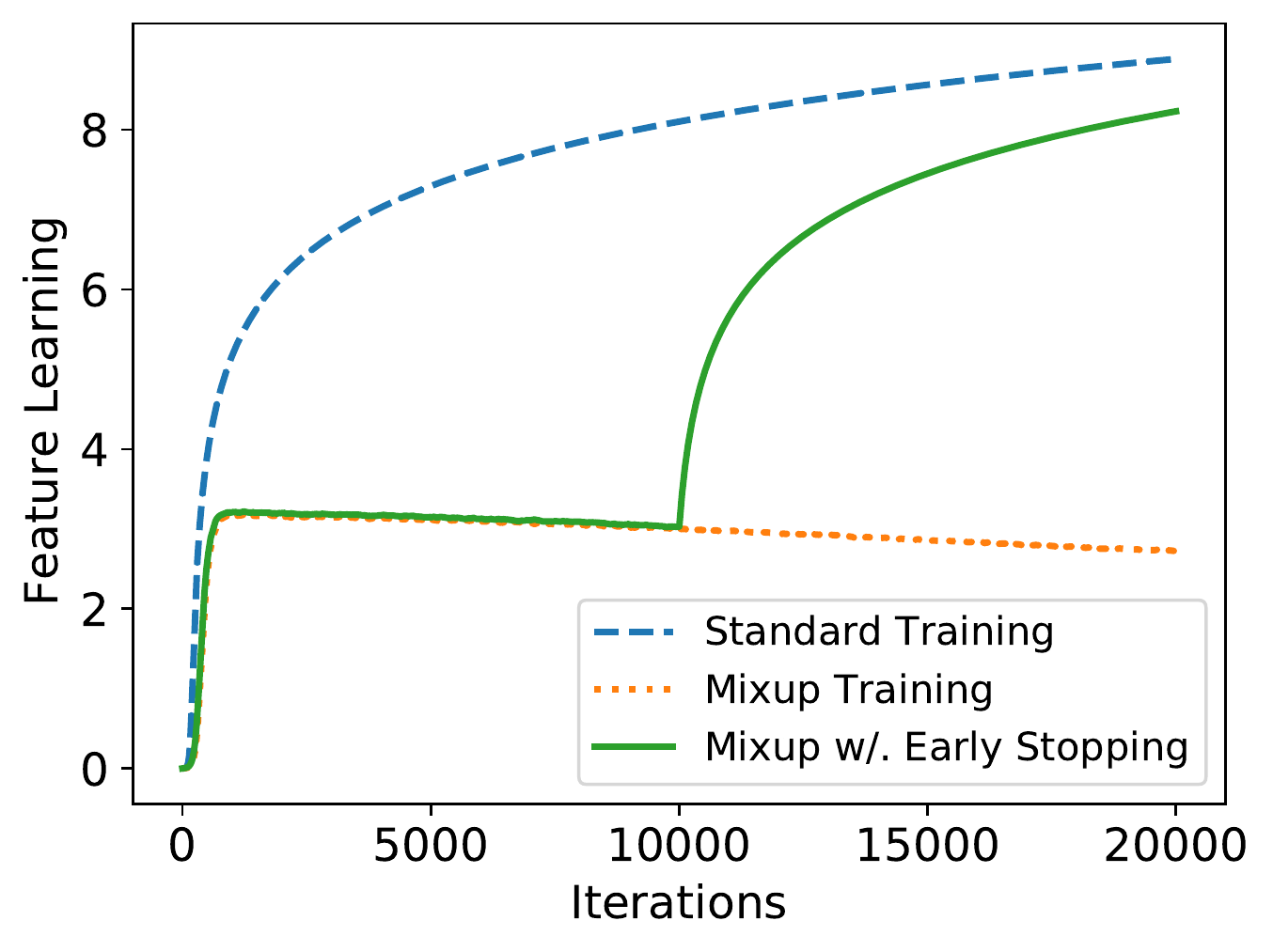}}
      \subfigure[Rare Feature Learning]{\includegraphics[width=0.505\columnwidth]{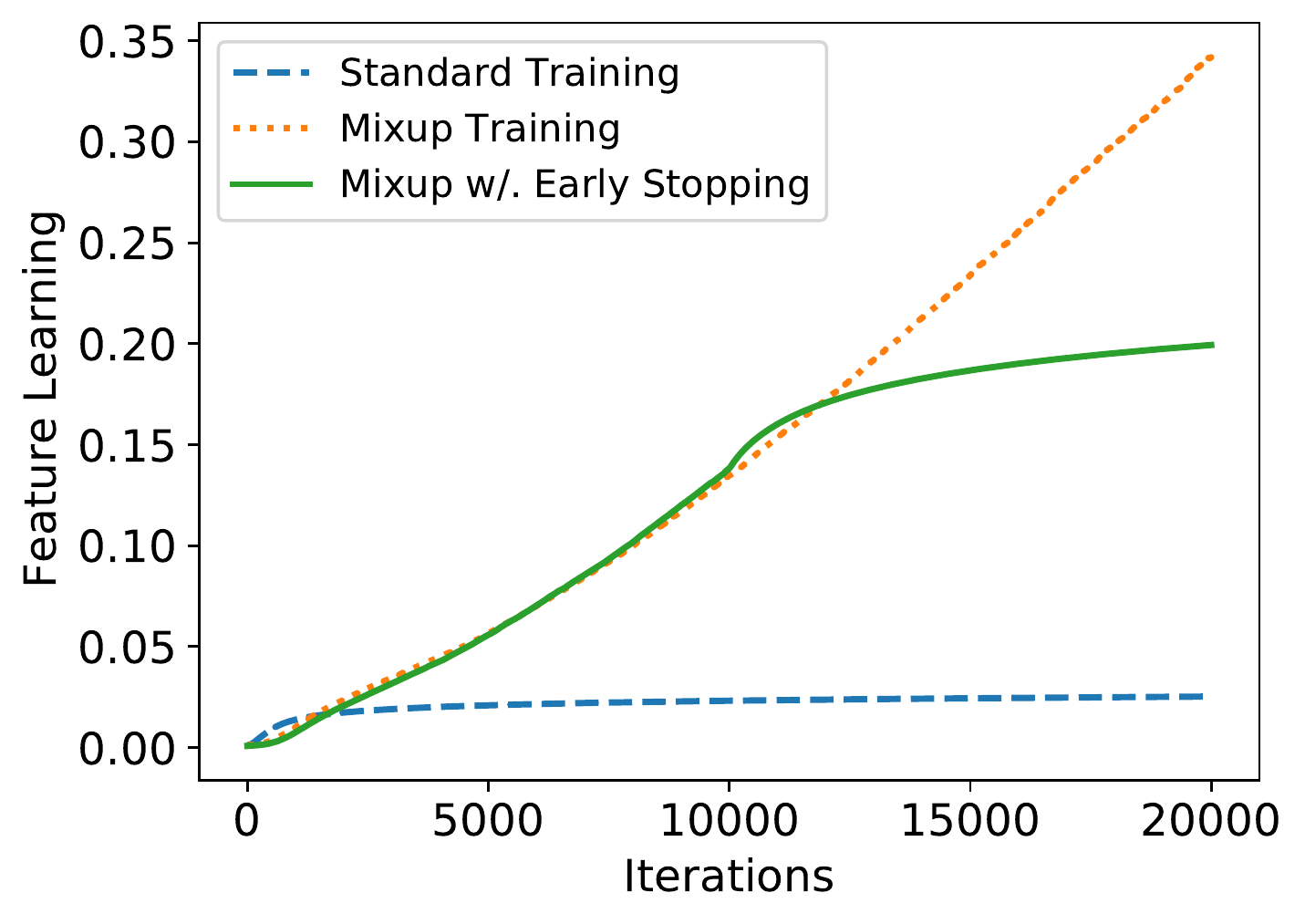}}
      \vskip -0.1in
    \caption{Common feature learning and rare feature learning on synthetic data, all experiments are conducted using full-batch gradient descent. Here we consider three training methods: standard training, Mixup training, and Mixup training with early stopping (at the $10000$-th iteration).}
    \label{fig:synthetic_feat_learn}
\end{figure}

\subsection{Implications to the Early Stopping of Mixup} 
In addition to demonstrating the ability of Mixup in learning rare features,
Lemma \ref{lemma:outcome_Mixup_main} also reveals that the benefits of Mixup training mostly come from its early training phase. Therefore, this motivates us to study the early-stopped Mixup training, i.e., the Mixup data augmentation will be turned off after a number of iterations. Then clearly, after turning off the Mixup data augmentation, the learned features will never be forgotten since the gradient update in this period will be always positively correlated (by \eqref{eq:update_features_main}). This immediately leads to the following fact.
\begin{fact}\label{fact:nodecrease}
Let $T$ be the same as that in Lemma \ref{lemma:outcome_Mixup_main}, then if early stopping Mixup training at the iteration $T$, we have for any $t>T$, it holds that $
\max_{r}|\la\wb_{1,r}^{(t)},\vb'\ra|, \max_{r}|\la\wb_{2,r}^{(t)},\ub'\ra| =\Omega(\rho\zeta^{1/2})$.
\end{fact}
\vspace{-2mm}

This further implies that applying proper early stopping in Mixup training will not affect the rare feature learning. Besides, turning off Mixup will  enhance the learning of common features (since its learning speed will no longer be affected by the mix with rare features and noises), which could potentially lead to even better generalization performance. In the next section, we will empirically justify the effectiveness of applying early stopping in Mixup training.

\section{Experiments}

\paragraph{Synthetic Data.} We first perform numerical experiments on synthetic data to verify our theoretical results. In particular, the synthetic data is generated according to Definition \ref{def:data_distribution_new}.  In particular, we set dimension $d=2000$, training sample size $n=300$, the ratio of rare feature data $\rho=0.1$, noise strength $\sigma_p=0.15$, feature noise strength $\alpha=0.05$, number of total patches $P=5$, and number feature noise patches $b=2$. For the two-layer CNN model and the training algorithm, we  set network width $m=10$, and conduct full-batch gradient descent with learning rate $\eta=0.05$ and total iteration number $T=20000$. We characterize the learning of common features and rare features via calculating $\sum_{r=1}^m(\la\wb_{1,r},\vb\ra)^2$ and $\sum_{r=1}^m(\la\wb_{1,r},\vb'\ra)^2$ (we only consider $\vb$ and $\vb'$ as the dynamics for $\ub$ and $\ub'$ are similar). The results are reported in Figure \ref{fig:synthetic_feat_learn}.
It is clear that both standard training, Mixup training, and Mixup with early stopping can exhibit sufficiently common feature learning, while the rare feature learning of standard training is much lower than those of Mixup and Mixup with early stopping. This verifies Lemmas \ref{lemma:standard_learning_phase3_main} and \ref{lemma:outcome_Mixup_main}. Besides, we can also see that turning off Mixup after a number of iterations will lead to no decrease in rare feature learning and an increase in common feature learning. This verifies Fact \ref{fact:nodecrease} and demonstrates the benefits of early stopping.\\
\begin{figure}[!t]
\vskip -0.1in
     \centering
     \subfigure[Training Loss]{\includegraphics[width=0.49\columnwidth]{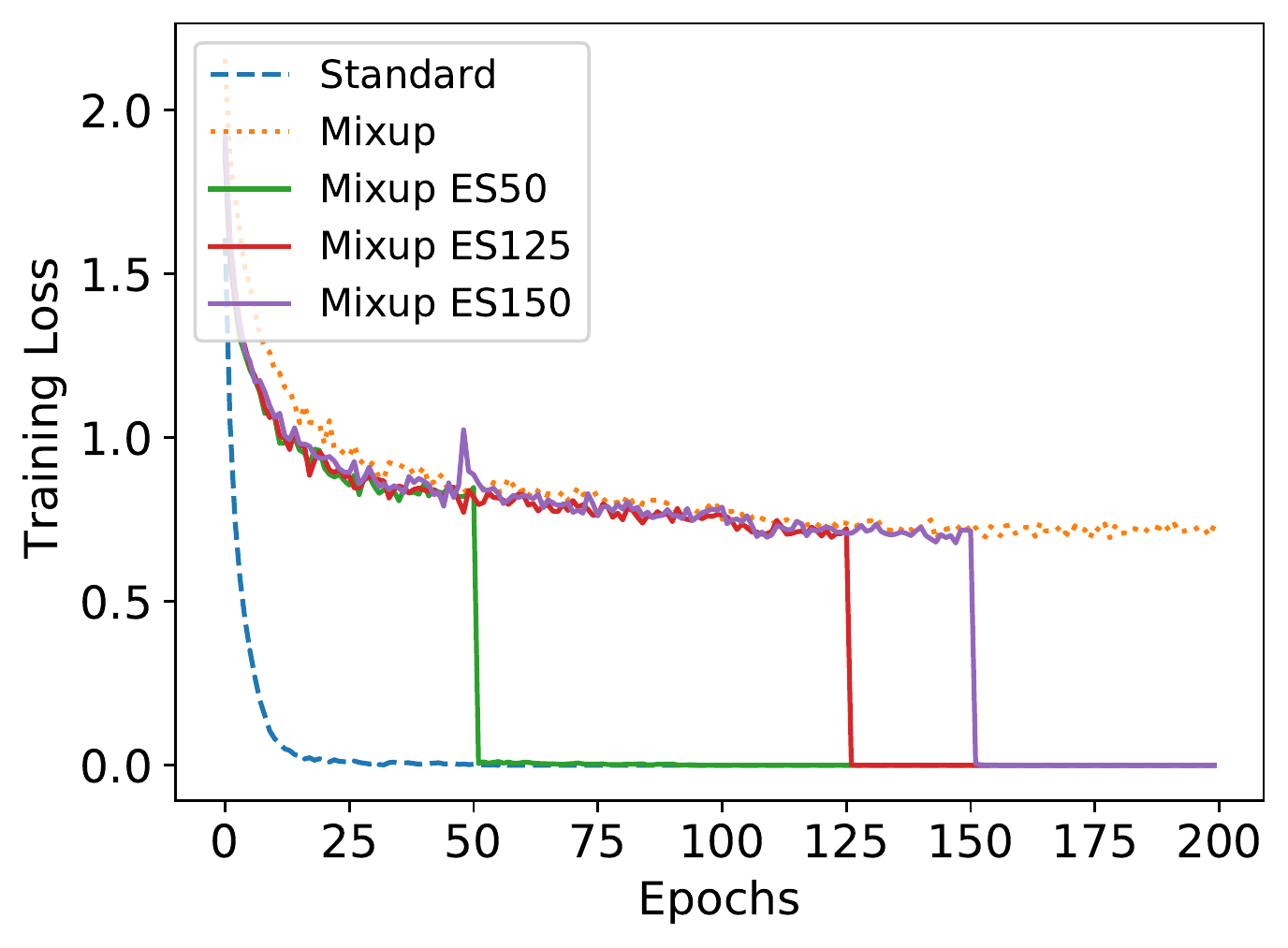}}
      \subfigure[Test Accuracy]{\includegraphics[width=0.49\columnwidth]{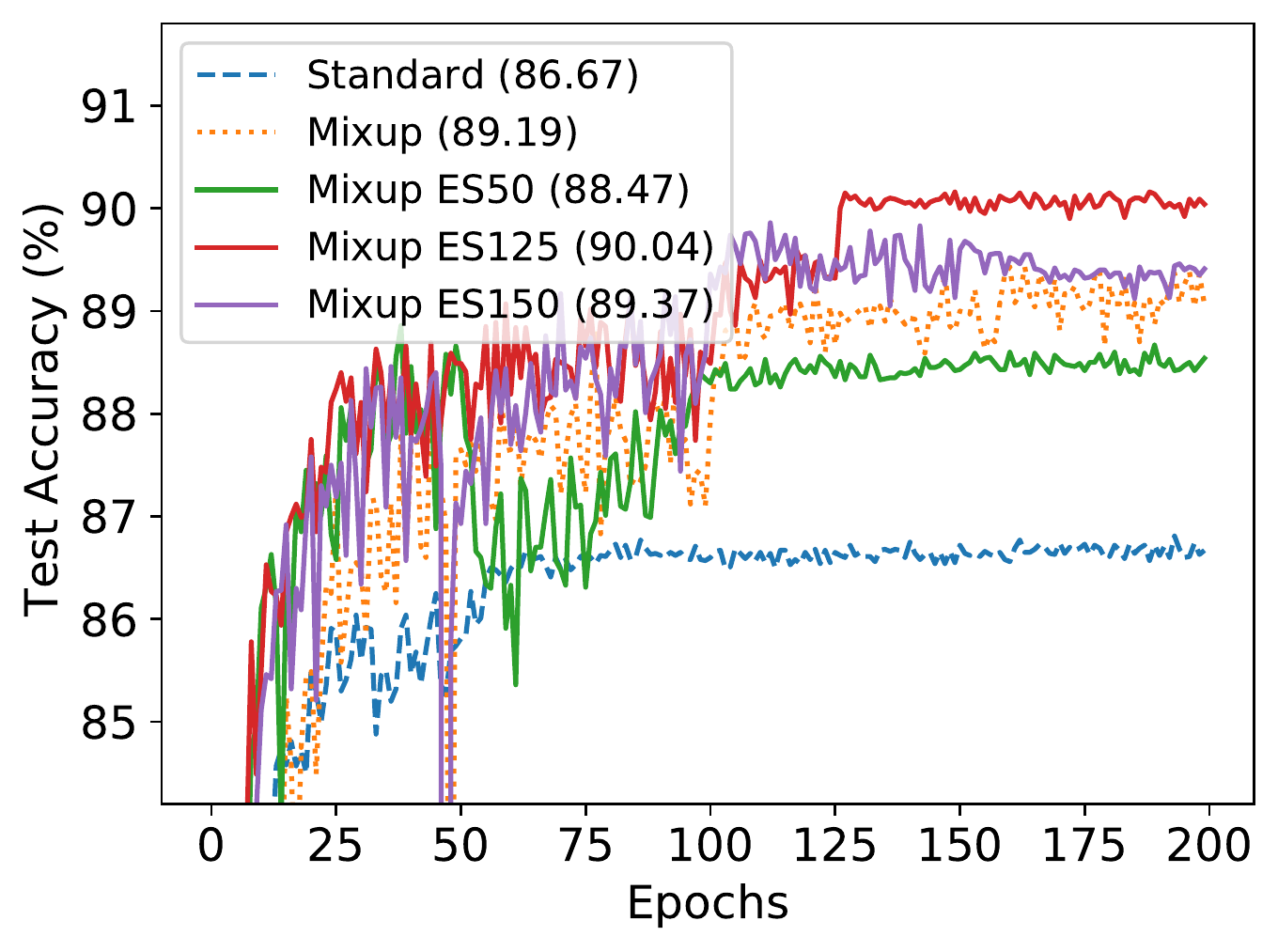}}
      \vskip -0.1in
    \caption{Training loss (the cross-entropy loss on the mixup data/clean data) and test accuracy achieved by Mixup with different early stopping iterations: 0 (standard), 50, 125, 150, 200 (Mixup), numbers in the legend denote the average accuracy of the last $10$ iterates. The results are evaluated by training ResNet18 on CIFAR-10 dataset without random crop \& flip data augmentation and weight decay regularization.}
\label{fig:early_stop_noaug}
\end{figure}
\paragraph{CIFAR-10 Data.} We further perform the Mixup training on CIFAR-10 dataset to evaluate the performance of early stopping, where we use SGD with momentum $0.9$ and learning rate $0.1$, followed by $\times0.1$ decaying at the $100$-th and $150$-th iterations. We first train the ResNet18 model \citep{he2015delving} via Mixup without other data augmentations and regularizations. We consider applying early stopping at the $0$-th (standard training), $50$-th, $125$-th, $150$-th, and $200$-th (Mixup training) iterations and report the training loss and test accuracy in Figure \ref{fig:early_stop_noaug}. First, it can be observed that the cross-entropy loss on the training data quickly drops to nearly zero after the stopping of Mixup, showing that the neural network has correctly predicted the labels of training data points with high confidence. Besides, the test accuracy results show that 
such a high-confidence fitting on training data will not affect the test performance, while proper early stopping can even gain further improvements, e.g., Mixup with early stopping at the $125$-th iteration achieves substantially higher test accuracy than that of Mixup training. This demonstrates the effectiveness of early-stopped Mixup and backs up our theoretical finding that the benefits of Mixup mainly stem from the early training phase. 

We further perform Mixup training for different neural network models and add the random crop/flip data augmentation and weight decay regularization (set as $10^{-4}$). In particular, we consider two (relatively) high-capacity models: ResNet18 and ResNet34; and two low-capacity models: LeNet and VGG16. For ResNet18 and ResNet34, we set the learning rate as $0.1$; for LeNet and VGG16, we set the learning rate
as $0.02$ and $0.1$ respectively. Then we can clearly see that applying proper early stopping in Mixup will not downgrade the test performance but can even lead to higher test accuracy. In particular, Mixup with early stopping at the $50$-th, $125$-th, and $150$-th iterations can still achieve a substantial performance improvement compared to standard training for LeNet, VGG16, and ResNet18. Moreover, 
we can also observe that Mixup with early stopping at the $150$-th iteration performs better than the standard Mixup for all $4$  models, especially for LeNet and VGG16, two relatively simpler models. This justifies our theoretical findings and demonstrates the benefit of early stopping in Mixup.

\begin{figure*}[!t]
\vskip -0.1in
     \centering
     \subfigure[LeNet]{\includegraphics[width=0.24\textwidth]{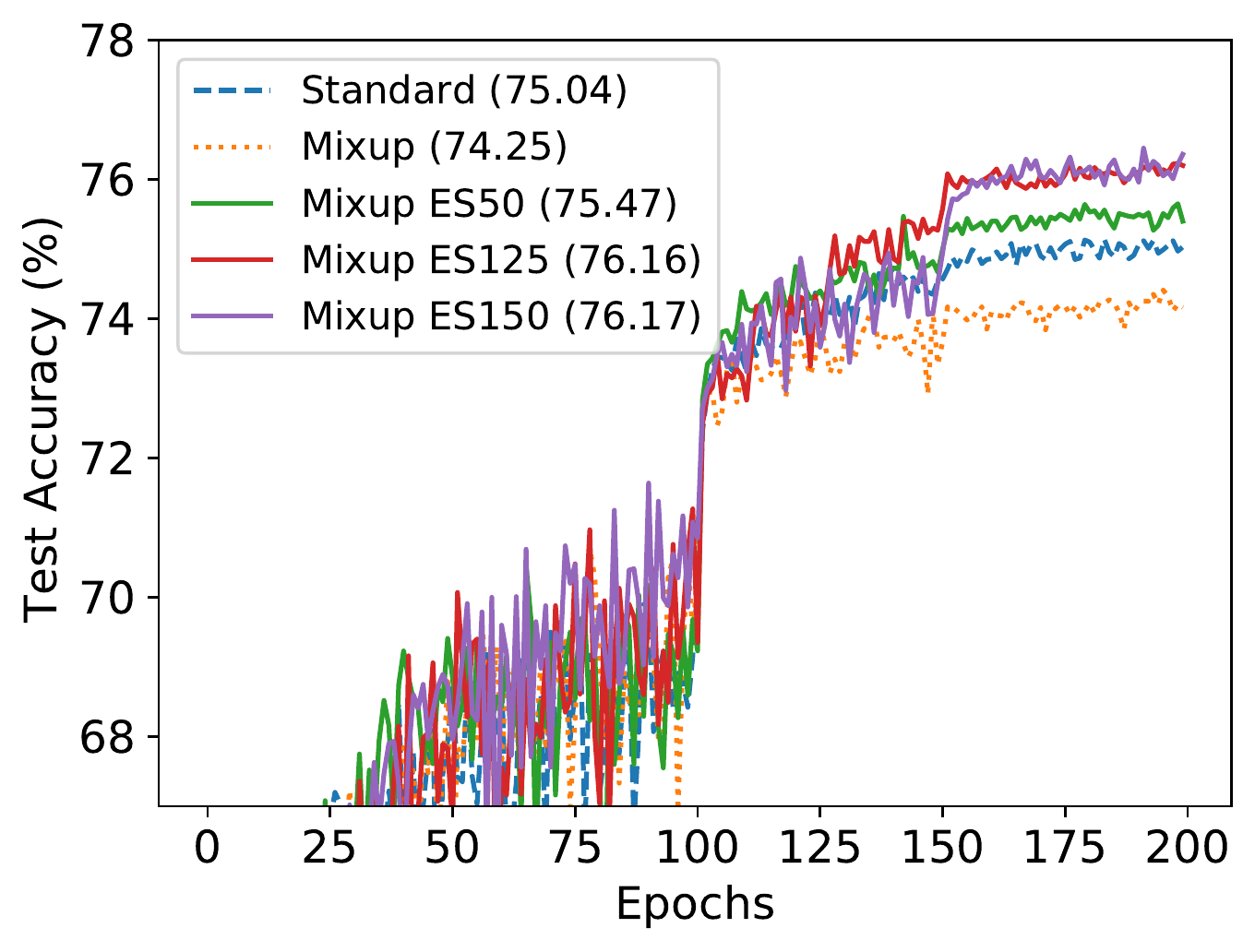}}
      \subfigure[VGG16]{\includegraphics[width=0.24\textwidth]{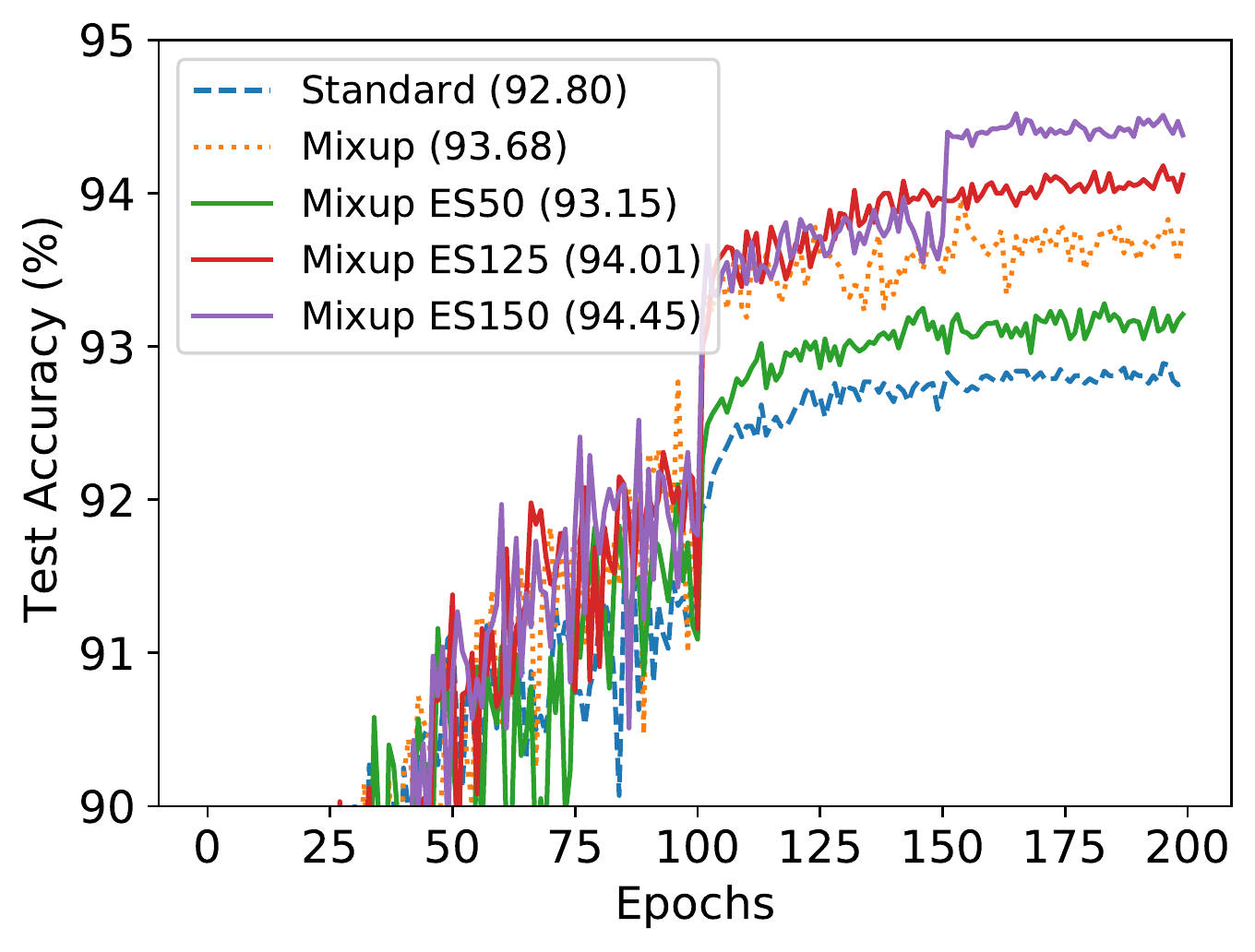}}
      \subfigure[ResNet18]{\includegraphics[width=0.24\textwidth]{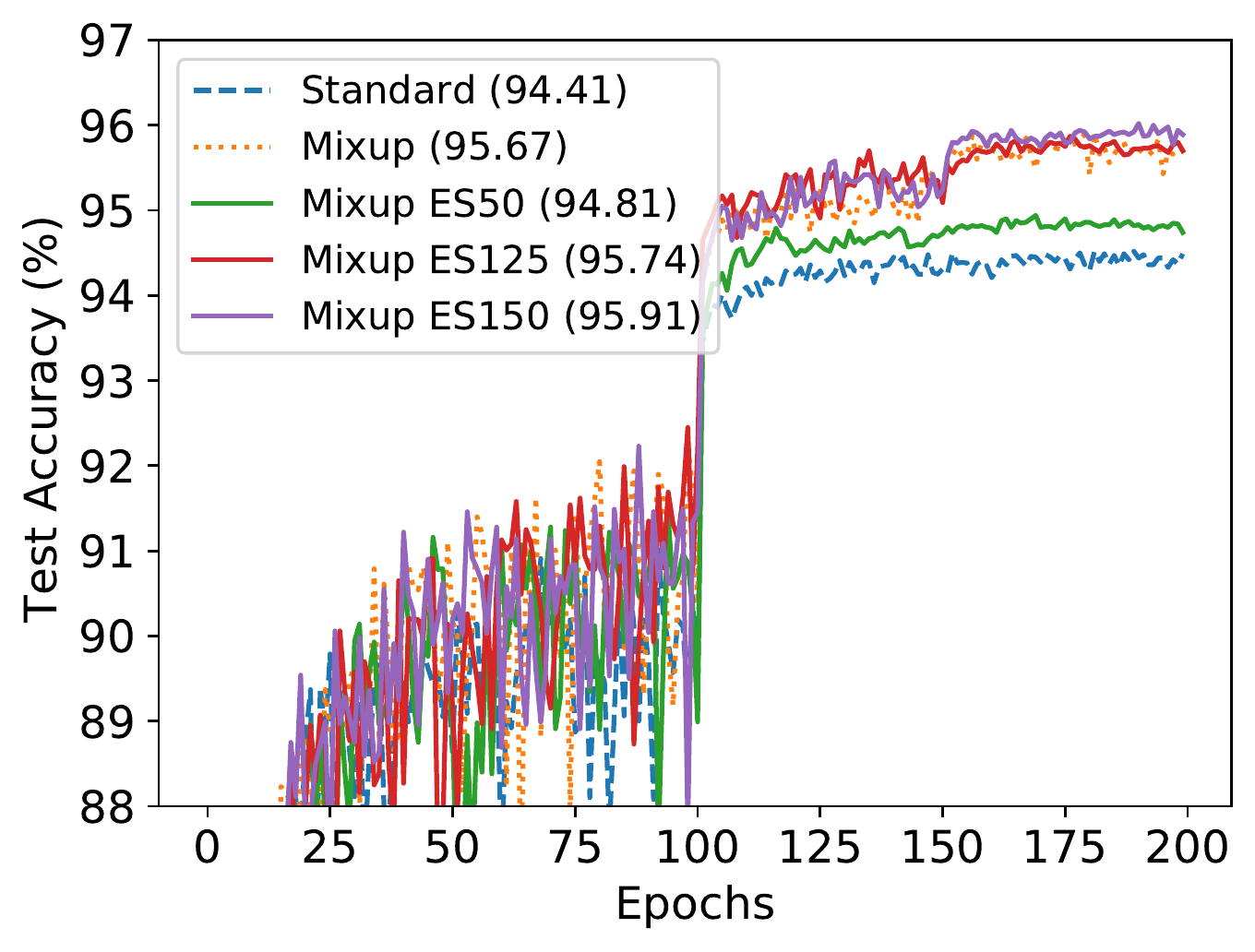}}
      \subfigure[ResNet34]{\includegraphics[width=0.24\textwidth]{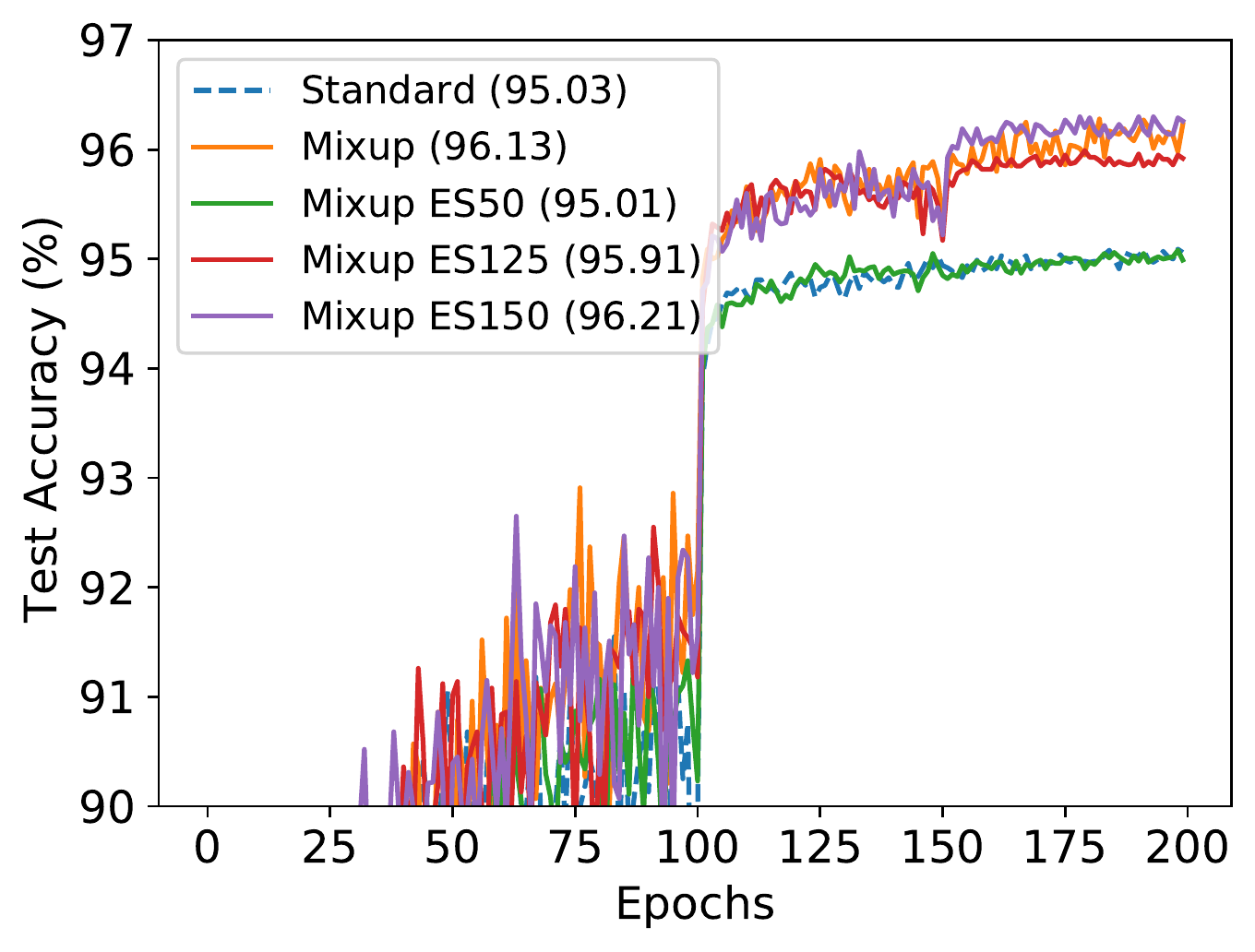}}
      \vskip -0.1in
    \caption{Test errors achieved by Mixup training with different early stopping iterations: 0 (standard), 50, 125, 150, 200 (Mixup), numbers in the legend denote the average accuracy of the last $10$ iterates. The results are evaluated by training LeNet, VGG16, ResNet18, and ResNet34 on CIFAR-10 dataset  with random crop \& flip data augmentation and weight decay regularization. Experimental results suggest that applying proper early stopping in Mixup will not downgrade the test performance but can even lead to higher test accuracy, especially for simpler models such as LeNet and VGG16.}
    \label{fig:early_stop_aug}
\end{figure*}


\section{Conclusion}
In this work, we attempted to develop a comprehensive understanding of the benefits of Mixup training.  We first identified that the benefits cannot be fully explained by the   linearity inductive bias of Mixup. Then we theoretically studied the dynamics of Mixup training from a feature learning. We showed that Mixup is more beneficial in learning rare features compared to standard training. Moreover, our analysis revealed that the benefits of Mixup in feature learning mostly stem from early training stages, based on which we developed the early-stopped Mixup. Our experimental results demonstrated that the early-stopped Mixup can achieve a comparable or even better performance than the standard one, which supports our theoretical findings.

\appendix

\section{Detailed Proof for Standard Training}

\subsection{Critical Quantities at the Initialization}
Before moving on to the detailed characterization of the dynamics of standard training and Mixup training, we first characterize a set of critical quantities at the initialization. Recall (1) the data model in Definition \ref{def:data_distribution_new} that the feature vectors have unit norm and the noise vectors are randomly generated  from $N(\boldsymbol{0},\sigma_p^2\Ib)$; and (2) the initial model parameter $\wb_{k,r}^{(0)}$ is randomly generated from $N(\boldsymbol{0},\sigma_0^2\Ib)$, we first give the following lemma that characterizes some critical quantities that will be repeatedly used in the later analysis.
\begin{lemma}\label{lemma:initialization}
With probability at least $1-1/\poly(n)$, it holds that for all $i\in[n]$, $k\in[2]$, $r\in[m]$, $\ab\in\{\vb,\ub,\vb',\ub'\}$,
\begin{align*}
&|\la\wb_{k,r}^{(0)},\ab\ra| = \tilde O(\sigma_0), \ \sum_{r\in[m]}\big(\la\wb_{k,r}^{(0)},\ab\ra\big)^2 = \tilde\Theta(\sigma_0^2).
\end{align*}
Additionally, for any noise patch $\bxi\in\{\bxi\}$,
\begin{align*}
|\la\wb_{k,r}^{(0)},\bxi\ra|=\tilde O(d^{1/2}\sigma_p\sigma_0),\ \sum_{r\in[m]}\big(\la\wb_{k,r}^{(0)},\bxi\ra\big)^2 = \tilde \Theta(d\sigma_p^2\sigma_0^2).
\end{align*}
\end{lemma}
\begin{proof}
Note that $\wb_{k,r}^{(0)}$ is randomly generated from $N(\boldsymbol{0},\sigma_0^2\Ib)$. Then using the fact that $m=\polylog(n)$, $\|\ab\|_2^2=1$, and $\|\bxi\|_2^2=\Theta(d\sigma_p^2)$ with probability at least $1-1/\poly(n)$, applying standard concentration arguments can lead to the desired results. 

\end{proof}

\subsection{Feature and Noise Learning of Standard Training}

We first restate the feature and noise learning of standard training as follows: for features, we have
\begin{align}\label{eq:update_features}
\la\wb_{k,r}^{(t+1)},\vb\ra & = \la\wb_{k,r}^{(t)},\ub\ra + \frac{2\eta}{n}\cdot\sum_{i\in[n]} \ell_{k,i}^{(t)} \sum_{p\in \cP_i(\vb)} \la\wb_{k,r}^{(t)},\vb\ra\cdot \alpha_{i,p}^2\|\vb\|_2^2\notag\\
\la\wb_{k,r}^{(t+1)},\ub\ra & = \la\wb_{k,r}^{(t)},\ub\ra + \frac{2\eta}{n}\cdot\sum_{i\in[n]} \ell_{k,i}^{(t)} \sum_{p\in \cP_i(\ub)} \la\wb_{k,r}^{(t)},\ub\ra\cdot \alpha_{i,p}^2\|\ub\|_2^2,\notag\\
\la\wb_{k,r}^{(t+1)},\vb'\ra & = \la\wb_{k,r}^{(t)},\vb\ra + \frac{2\eta}{n}\cdot\sum_{i\in\cS_1^{+}} \ell_{k,i}^{(t)} \sum_{p\in \cP_i(\vb')} \la\wb_{k,r}^{(t)},\vb'\ra\cdot \|\vb'\|_2^2,\notag\\
\la\wb_{k,r}^{(t+1)},\ub'\ra & = \la\wb_{k,r}^{(t)},\ub'\ra + \frac{2\eta}{n}\cdot\sum_{i\in\cS_1^{-}} \ell_{k,i}^{(t)} \sum_{p\in \cP_i(\ub')} \la\wb_{k,r}^{(t)},\ub'\ra\cdot \|\ub'\|_2^2,
\end{align}
where $\cP_i(\ab)$ denotes the set of patches in $\xb_i$ containing the feature $\ab$ and $\alpha_{i,p}^2=1$ if $\xb_i^{(p)}$ is a feature patch and $\alpha_{i,p}^2=\alpha^2$ if $\xb_i^{(p)}$ is the feature noise. 
Additionally, note that the update of rare features only depends on the data in $\cS_1^+$ and $\cS_1^-$ since the data $(\xb_i, y_i)$ in $\cS_0^+$ and $\cS_0^-$ satisfies $\cP_i(\vb')=\emptyset$ and $\cP_i(\ub')=\emptyset$. Similarly, we can also obtain the following result regarding noise learning
\begin{align*}
&\la\wb_{k,r}^{(t+1)},\bxi_s^{(q)}\ra = \la\wb_{k,r}^{(t)},\bxi_s^{(q)}\ra + \frac{2\eta}{n}\cdot\sum_{i=1}^n\ell_{k,i}^{(t)} \sum_{p=1}^P \la\wb_{k,r}^{(t)},\xb_i^{(p)}\ra\cdot \la\xb_i^{(p)}, \bxi_s^{(q)}\ra.
\end{align*}
Moreover, note that if $\xb_i^{(p)}\neq \bxi_s^{(q)}$ (i.e., $i\neq s$ or $p\neq q$), then $|\la\xb_i^{(p)},\bxi_s^{(p)}\ra|$ is in the order of $\tilde O(d^{1/2}\sigma_p^2)$.
Therefore, we further have
\begin{align}\label{eq:update_noise_simplified}
\la\wb_{k,r}^{(t+1)},\bxi_s^{(q)}\ra 
& = \la\wb_{k,r}^{(t)},\bxi_s^{(q)}\ra \cdot\bigg[ 1 + \frac{2\eta}{n}\cdot \ell_{k,s}^{(t)}\cdot\|\bxi_s^{(q)}\|_2^2\bigg]  \pm \frac{2\eta}{n}\cdot\sum_{i\neq s || p\neq q} |\ell_{k,i}^{(t)}|\cdot|\la\wb_{k,r}^{(t)},\bxi_i^{(q)}\ra|\cdot \tilde O\big(d^{1/2}\sigma_p^2\big).
\end{align}

\paragraph{Phase 1, Fitting Common Feature Data.}

The following lemma characterizes the learning of all feature and noise vectors in Phase 1.

\begin{lemma}[Phase 1, Standard Training]\label{lemma:learning_phase1}
Let $T_0$ be the iteration number such that the neural network output satisfies $|F_k(\Wb^{(t)};\xb_i)|\le O(1)$ for all $t\le T_0$ and $i\in[n]$, then for any $t\le T_0$, it holds that
\begin{align*}
\la\wb_{1,r}^{(t+1)},\vb\ra  = \la\wb_{1,r}^{(t)},\vb\ra\cdot \big(1 + \Theta(\eta)\big),\quad
\la\wb_{2,r}^{(t+1)},\ub\ra  = \la\wb_{2,r}^{(t)},\ub\ra\cdot \big(1 + \Theta(\eta)\big).
\end{align*}
Besides, we also have for any $t\le T_0$, $r\in[m]$, $k\in[2]$, $q\in[P]$, and $s\in[n]$,
\begin{align*}
&|\la\wb_{2,r}^{(t)},\vb\ra| = \tilde O(\sigma_0),\quad |\la\wb_{1,r}^{(t)},\ub\ra| = \tilde O(\sigma_0),\\
&|\la\wb_{k,r}^{(t)},\vb'\ra| = \tilde O(\sigma_0),\quad |\la\wb_{k,r}^{(t)},\ub'\ra| = \tilde O(\sigma_0), \quad|\la\wb_{k,r}^{(t)}, \bxi_s^{(q)}\ra| = \tilde O\big(d^{1/2}\sigma_p\sigma_0\big).
\end{align*}
\end{lemma}
\begin{proof}
First, note that in the first stage, the neural network outputs are in the order of $O(1)$, implying that the loss derivatives satisfy $|\ell_{k,i}^{(t)}| = \Theta(1)$. More specifically, we can get that $\ell_{k,i}^{(t)} = \Theta(1)$ if $k=y_i$ and $\ell_{k,i}^{(t)} = -\Theta(1)$ otherwise. Then by \eqref{eq:update_features}, we have
\begin{align*}
\la\wb_{1,r}^{(t+1)}, \vb\ra = \la\wb_{1,r}^{(t)}, \vb\ra\cdot\bigg[1 + \frac{2\eta}{n}\cdot\sum_{i\in\cS_0^+} \ell_{1,i}^{(t)} \sum_{p\in \cP_i(\vb)} \alpha_{i,p}^2\|\vb\|_2^2 + \frac{2\eta}{n}\cdot\sum_{i\in[n]\backslash\cS_0^+} \ell_{1,i}^{(t)} \sum_{p\in \cP_i(\vb)}  \alpha_{i,p}^2\|\vb\|_2^2\bigg].
\end{align*}
Note that by Definition \ref{def:data_distribution_new}, for any data $i\in[n]$ let $\cP_i'(\vb)$ and $\cP_i'(\ub)$ be the set of patches corresponding to the feature noise vectors $\vb$ and $\ub$ respectively, we have $|\cP_i'(\vb)|\le b$ and $\sum_{p\in\cP'_i(\vb)}\alpha_{i,p}^2\le b\alpha^2=o\big(1/\polylog(n)\big)$. Additionally, note that $\ell_{1,i}^{(t)}=\Theta(1)$ for $i\in\cS_0^+$ and $\cP_i(\vb) = \cP_i'(\vb)$ for all $i\in[n]\backslash \cS_0^+$, we have
\begin{align}\label{eq:update_v_phase1_std}
\la\wb_{1,r}^{(t+1)}, \vb\ra = \la\wb_{1,r}^{(t)}, \vb\ra\cdot\bigg[1 + \frac{2\eta}{n}\cdot |\cS_0^+|\cdot C_v^{(t)} \pm o\big(\eta/\polylog(n)\big)\bigg] = \la\wb_{1,r}^{(t)}, \vb\ra\cdot \big[1+\Theta(\eta)\big], 
\end{align}
where $C_v^{(t)}=|\cS_0^+|^{-1}\cdot\sum_{i\in\cS_0^+}\ell_{1,i}^{(t)}$ remains in the constant level for all $t\le T_0$.
Similarly, we can also get that 
\begin{align}\label{eq:update_u_phase1_std}
\la\wb_{2,r}^{(t+1)}, \ub\ra = \la\wb_{2,r}^{(t)}, \ub\ra\cdot\bigg[1 + \frac{2\eta}{n}\cdot |\cS_0^-|\cdot C_u^{(t)} \pm o\big(\eta/\polylog(n)\big)\bigg] =  \la\wb_{2,r}^{(t)}, \ub\ra\cdot \big[1+\Theta(\eta)\big],
\end{align}
where $C_u^{(t)}=|\cS_0^-|^{-1}\cdot\sum_{i\in\cS_0^-}\ell_{2,i}^{(t)}$ remains in the constant level for all $t\le T_0$.
Moreover, in terms of the learning of wrong features, we have
\begin{align}\label{eq:update_wrongstrongfeat}
\la\wb_{2,r}^{(t+1)},\vb\ra &= \la\wb_{2,r}^{(t)}, \vb\ra\cdot\bigg[1 + \frac{2\eta}{n}\cdot\sum_{i\in\cS_0^+} \ell_{2,i}^{(t)} \sum_{p\in \cP_i(\vb)} \alpha_{i,p}^2\|\vb\|_2^2 + \frac{2\eta}{n}\cdot\sum_{i\in[n]\backslash\cS_0^+} \ell_{2,i}^{(t)} \sum_{p\in \cP_i(\vb)}  \alpha_{i,p}^2\|\vb\|_2^2\bigg]\notag\\
& = \la\wb_{2,r}^{(t)}, \vb\ra\cdot\bigg[1 - \frac{2\eta}{n}\cdot |\cS_0^+|\cdot\Theta(1)\pm o(\eta)\bigg]\notag\\
& = \la\wb_{2,r}^{(t)}, \vb\ra\cdot\big[1 - \Theta(\eta)\big].
\end{align}
Then by Lemma \ref{lemma:initialization}, this further implies that for all $t$ in the first stage, we have
\begin{align}\label{eq:bound_wrongstrongfeat}
|\la\wb_{2,r}^{(t)},\vb\ra|\le |\la\wb_{2,r}^{(t-1)},\vb\ra|\le\dots\le |\la\wb_{2,r}^{(0)},\vb\ra|=\tilde O(\sigma_0).
\end{align}

Now we can move on to the learning of rare features and noise vectors. Particularly, for rare features, we have
\begin{align*}
\la\wb_{1,r}^{(t+1)}, \vb'\ra &= \la\wb_{1,r}^{(t)}, \vb'\ra\cdot \bigg[1 + \frac{2\eta}{n}\cdot \sum_{i\in\cS_1^+}\ell_{k,i}^{(t)}\cdot \sum_{p\in\cP_i(\vb')}\|\vb'\|_2^2\bigg] \notag\\
& = \la\wb_{1,r}^{(t)}, \vb'\ra\cdot \bigg[1 + \Theta\bigg(\frac{\eta|\cS_1^+|}{n}\bigg)\bigg]\notag\\
& = \la\wb_{1,r}^{(t)}, \vb'\ra\cdot \big[1 + \Theta(\rho \eta)\big],
\end{align*}
where the second equality is due to $|\cP_i(\vb')|=\Theta(1)$ and the last equality is due to $|\cS_1^+| = \Theta(\rho n)$ with probability at least $1-1/\poly(n)$. Therefore, by Lemma \ref{lemma:initialization}, we can then obtain
\begin{align*}
|\la\wb_{1,r}^{(t)}, \vb'\ra|\le \big[1+\Theta(\rho\eta)\big]^{t}\cdot |\la\wb_{1,r}^{(t)}, \vb'\ra|\le \tilde O(\sigma_0)\cdot e^{\Theta(T_0\eta)}=\tilde O(\sigma_0),
\end{align*}
where we use the fact that $T_0 = \tilde O(1/\eta)$.
Similarly, it also follows that
\begin{align*}
\la\wb_{1,r}^{(t+1)}, \ub'\ra = \la\wb_{1,r}^{(t)}, \ub'\ra\cdot [1 + \Theta(\rho \eta)] = \tilde O(\sigma_0).
\end{align*}
Moreover, using the fact that $\ell_{2,i}^{(t)}= -\Theta(1)$ for $i\in\cS_1^+$ and $\ell_{1,i}^{(t)}= -\Theta(1)$ for $i\in\cS_2^+$, we can follow the same proof in  \eqref{eq:update_wrongstrongfeat} and \eqref{eq:bound_wrongstrongfeat} and get
\begin{align*}
|\la\wb_{2,r}^{(t)}, \vb'\ra| \le |\la\wb_{2,r}^{(0)}, \vb'\ra|=\tilde O(\sigma_0), \ |\la\wb_{1,r}^{(t)}, \ub'\ra| \le |\la\wb_{1,r}^{(0)}, \ub'\ra| =\tilde O(\sigma_0).
\end{align*}
where the results for $|\la\wb_{1,r}^{(t)}, \vb'\ra|$ and $|\la\wb_{1,r}^{(t)}, \ub'\ra|$ are by Lemma \ref{lemma:initialization}.

Finally, regarding the learning of the noise vector $\bxi_s^{(q)}$, if $j = y_s$, we have the following by \eqref{eq:update_noise_simplified},
\begin{align*}
\max_{s,r}|\la\wb_{k,r}^{(t+1)},\bxi_s^{(q)}\ra| \le \max_{s,r}|\la\wb_{k,r}^{(t)},\bxi_s^{(q)}\ra |\cdot\bigg[1 + \frac{\eta}{n}\cdot \tilde \Theta(d\sigma_p^2) + \frac{\eta}{n}\cdot \tilde O\big(nP d^{1/2}\sigma_p^2\big)\bigg].
\end{align*}
Note that we have $nP=o(d^{1/2})$, then the above equation further leads to
\begin{align*}
\max_{s,r}|\la\wb_{k,r}^{(t+1)},\bxi_s^{(q)}\ra| =\max_{s,r}|\la\wb_{k,r}^{(t)},\bxi_s^{(q)}\ra| \cdot\bigg[1 + \frac{\eta}{n}\cdot \tilde \Theta(d\sigma_p^2) \bigg].
\end{align*}
Besides, we can also get if $k\neq y_s$,
\begin{align*}
\max_{s,r}|\la\wb_{k,r}^{(t+1)},\bxi_s^{(q)}\ra| \le \max_{s,r}|\la\wb_{k,r}^{(t)},\bxi_s^{(q)}\ra| \cdot\bigg[1 - \frac{\eta}{n}\cdot \tilde \Theta(d\sigma_p^2) \bigg].
\end{align*}
Then for any $t\le T_0 = \tilde O(1/\eta)$ and any $k$, we have
\begin{align*}
\max_{s,r}|\la\wb_{k,r}^{(t)},\bxi_s^{(q)}| &\le \max_{s,r}|\la\wb_{k,r}^{(0)},\bxi_s^{(q)}\ra| \cdot \bigg[1 + \frac{\eta}{n}\cdot \tilde \Theta(d\sigma_p^2) \bigg]^t\notag\\
&\le \max_{s,r}|\la\wb_{k,r}^{(0)},\bxi_s^{(q)}\ra|\cdot \bigg[1 + \frac{\eta}{n}\cdot \tilde \Theta(d\sigma_p^2) \bigg]^{T_0}\notag\\
&\le \max_{s,r}|\la\wb_{k,r}^{(0)},\bxi_s^{(q)}\ra|\cdot \exp\big\{\tilde\Theta(\eta T_0 d \sigma_p^2/n)\big\}\notag\\
& \le \max_{s,r}|\la\wb_{k,r}^{(0)},\bxi_s^{(q)}\ra|\cdot \Theta(1)\notag\\
& = \tilde \Theta(d^{1/2}\sigma_0\sigma_p).
\end{align*}
This completes the proof.
\end{proof}

\begin{lemma}\label{lemma:results_phase1}
At the end of Phase 1 with maximum iteration number $T_0 = \tilde O(1/\eta)$, we have
\begin{align*}
\sum_{r=1}^m (\la\wb_{1,r}^{(T_0)},\vb\ra)^2 = \tilde\Theta(1), \ \sum_{r=1}^m (\la\wb_{2,r}^{(T_0)},\ub\ra)^2 = \tilde\Theta(1);
\end{align*}
besides, it holds that
\begin{align*}
|\la\wb_{2,r}^{(T_0)}, \vb\ra|, |\la\wb_{1,r}^{(T_0)}, \ub\ra|, |\la\wb_{k,r}^{(T_0)}, \ub'\ra|, |\la\wb_{k,r}^{(T_0)}, \vb'\ra|=\tilde O(\sigma_0); \quad|\la\wb_{k,r}^{(T_0)}, \bxi\ra|=\tilde O(d^{1/2}\sigma_p\sigma_0)
\end{align*}
for all $k\in[2]$, $r\in[m]$ and $\bxi\in\{\bxi\}$.
\end{lemma}
\begin{proof}
We first characterize the difference between $C_v^{(t)}$ and $C_u^{(t)}$ in \eqref{eq:update_v_phase1_std} and \eqref{eq:update_u_phase1_std}. Particularly, we consider the iterations that $\max_{i\in[n], k\in[2]}|F_k(\Wb^{(t)};\xb_i)|\le \zeta$ for some $\zeta = \Theta\big(1/\log(1/\sigma_0)\big)=\Theta(1/\polylog(n))$, then we can immediately get that it holds that $|\ell_{1,i}^{(t)}-0.5|\le O(\zeta)$ for all $i\in\cS_0^+$ and $|\ell_{2,i}^{(t)}-0.5|\le O(\zeta)$ for all $i\in\cS_0^-$. Therefore, we can further get
\begin{align*}
 C_v^{(t)} = \frac{1}{|\cS_0^+|}\cdot \sum_{i\in\cS_0^+}\ell_{1,i}^{(t)} = 0.5 \pm O(\zeta),\quad C_u^{(t)} = \frac{1}{|\cS_0^-|}\cdot \sum_{i\in\cS_0^-}\ell_{2,i}^{(t)} = 0.5 \pm O(\zeta).
\end{align*}
Further note that the positive and negative data are independently generated from the data distribution, which implies that with probability at least $1-1/\poly(n)$, it holds that $||\cS_0^+| - (1-\rho)n/2|\le \tilde O(n^{1/2})$ and $||\cS_0^-| - (1-\rho)n/2|\le \tilde O(n^{1/2})$. Therefore, applying the fact that $\zeta = \Theta(1/\polylog(n))$,  we can obtain the following by \eqref{eq:update_v_phase1_std} and \eqref{eq:update_u_phase1_std} 
\begin{align}\label{eq:update_strong_feature_phase1_tmp1}
\sum_{r=1}^m(\la\wb_{1,r}^{(t+1)},\vb\ra)^2 &= \sum_{r=1}^m(\la\wb_{1,r}^{(t)},\vb\ra)^2\cdot \bigg[1+(1-\rho)\eta \pm O(\zeta\eta))\bigg]\notag\\
\sum_{r=1}^m(\la\wb_{2,r}^{(t+1)},\ub\ra)^2 &= \sum_{r=1}^m(\la\wb_{2,r}^{(t)},\ub\ra)^2\cdot \bigg[1+(1-\rho)\eta \pm O(\zeta\eta))\bigg].
\end{align}
Then let $T_0'$ be the largest iteration number such that $\max_{k,i}|F_k(\Wb^{(t)};\xb_i)|\le \zeta$, which clearly satisfies $T_0'<T_0$ ($T_0$ is defined in Lemma \ref{lemma:learning_phase1}), applying Lemma \ref{lemma:learning_phase1} and considering the data $i$ with largest neural network output (w.o.l.g assuming it's positive data),
\begin{align*}
\sum_{r=1}^m(\la\wb_{1,r}^{(T_0'+1)},\vb\ra)^2 \ge  c\cdot F_1(\Wb^{(T_0'+1)};\xb_i) \ge c\cdot\zeta
\end{align*}
for some absolute constant $c$. By \eqref{eq:update_strong_feature_phase1_tmp1}, we can immediately obtain that $T_0'=\Theta(\log(\zeta/(m\sigma_0^2))/\eta)$, where we apply the initialization results in Lemma \ref{lemma:initialization}. Besides, we can also obtain that
\begin{align*}
\frac{\sum_{r=1}^m(\la\wb_{2,r}^{(t+1)},\ub\ra)^2}{\sum_{r=1}^m(\la\wb_{1,r}^{(t+1)},\vb\ra)^2 } &\ge  \frac{\sum_{r=1}^m(\la\wb_{2,r}^{(0)},\ub\ra)^2}{\sum_{r=1}^m(\la\wb_{1,r}^{(0)},\vb\ra)^2 }\cdot \bigg(\frac{1+(1-\rho)\eta - O(\zeta \eta))}{1+(1-\rho)\eta + O(\zeta\eta)}\bigg)^{T_0'}\notag\\
& = \tilde\Theta(1)\cdot \big(1 - O(\eta\zeta T_0) \big).
\end{align*}
Then note that $\zeta = \Theta\big(1/\log(1/\sigma_0)\big)$, we can get $\zeta T_0\eta = \Theta\big(\zeta\log(\zeta) + \zeta \log(1/(m\sigma_0^2))\big)=o(1)$, which implies that $\sum_{r=1}^m(\la\wb_{2,r}^{(T_0'+1)},\ub\ra)^2\ge \Theta(\zeta)$.  Finally, by Lemma \ref{lemma:learning_phase1}, we know that $\sum_{r=1}^m(\la\wb_{1,r}^{(t)},\vb\ra)^2$
and $\sum_{r=1}^m(\la\wb_{2,r}^{(t+1)},\ub\ra)^2$ will keep increasing for all $t\le T_0$. Then based on the definition of $T_0$ and the fact that $\zeta = \tilde\Theta(1)$, we can conclude that 
\begin{align*}
\sum_{r=1}^m (\la\wb_{1,r}^{(T_0)},\vb\ra)^2 = \tilde\Theta(1), \ \sum_{r=1}^m (\la\wb_{2,r}^{(T_0)},\ub\ra)^2 = \tilde\Theta(1).
\end{align*}
The remaining arguments in this lemma directly follow from Lemma \ref{lemma:learning_phase1}, thus we omit their proof here.
\end{proof}

\paragraph{Phase 2. Fitting Rare Feature Data.} After \textbf{Phase 1}, the neural network output will become larger so that the loss derivatives (i.e, $\ell_{k,i}^{(t)}$) or the output logits may no longer be viewed as a quantity in the constant order.
Particularly, as shown in Lemma \ref{lemma:results_phase1}, when $t>T_0$, the feature learning, i.e., $\la\wb_{1,r}^{(t)},\vb\ra$ and $\la\wb_{2,r}^{(t)},\ub\ra$ will reach the constant order, implying that $|\ell_{k,i}^{(t)}|$ will be closer to $1$ or $0$ for all common feature data. Additionally, the loss derivative will remain in the constant order for the rare feature data, since either the rare feature learning (e.g, $\la\wb_{1,r}^{(t)}, \vb'\ra$) or the noise learning (e.g., $\la\wb_{1,r}^{(t)},\bxi_i^{(p)}\ra$) will be in the order of $o\big(1/\polylog(n)\big)$, so that the corresponding neural network outputs are also in the order of $o\big(1/\polylog(n)\big)$. Therefore, we define \textbf{Phase 2} by the period that (1) is after \textbf{Phase 1} and (2) the neural network outputs for the rare feature data are still in the order of $O\big(1/\polylog(n)\big)$ (or equivalently, the loss derivatives of rare feature data are in the constant order.)

Then, similar to the analysis in Phase 1, we will also characterize the learning of feature and noise separately. Regarding the learning of common feature, by \eqref{eq:update_features}, we have
\begin{align}\label{eq:update_feature_v_phase2}
\la\wb_{k,r}^{(t+1)},\vb\ra & = \la\wb_{k,r}^{(t)},\vb\ra + \frac{2\eta}{n}\cdot\sum_{i\in[n]} \ell_{k,i}^{(t)} \sum_{p\in \cP_i(\vb)} \la\wb_{k,r}^{(t)},\vb\ra\cdot \alpha_{i,p}^2\notag\\
& = \la\wb_{k,r}^{(t)},\vb\ra\cdot\bigg[1 + \frac{2\eta}{n}\cdot\bigg(\sum_{i\in\cS_0^+}\ell_{k,i}^{(t)}\sum_{p\in\cP_i(\vb)}\alpha_{i,p}^2 +\sum_{i\in\cS_0^-}\ell_{k,i}^{(t)}\sum_{p\in\cP_i(\vb)}\alpha_{i,p}^2+\sum_{\cS_1^+\cup\cS_1^-}\ell_{k,i}^{(t)}\sum_{p\in\cP_i(\vb)}\alpha_{i,p}^2 \bigg)\bigg].
\end{align}
Similarly, we can also get that
\begin{align}\label{eq:update_feature_u_phase2}
\la\wb_{k,r}^{(t+1)},\ub\ra 
& = \la\wb_{k,r}^{(t)},\ub\ra\cdot\bigg[1 + \frac{2\eta}{n}\cdot\bigg(\sum_{i\in\cS_0^+}\ell_{k,i}^{(t)}\sum_{p\in\cP_i(\ub)}\alpha_{i,p}^2 +\sum_{i\in\cS_0^-}\ell_{k,i}^{(t)}\sum_{p\in\cP_i(\ub)}\alpha_{i,p}^2+\sum_{\cS_1^+\cup\cS_1^-}\ell_{k,i}^{(t)}\sum_{p\in\cP_i(\ub)}\alpha_{i,p}^2 \bigg)\bigg].
\end{align}
Moreover, according to the data distribution in Definition \ref{def:data_distribution_new}, we have
\begin{itemize}
    \item For any $i\in\cS_0^+$, it holds that $\sum_{p\in\cP_i(\vb)}\alpha_{i,p}^2 = \Theta(1)$ and $\sum_{p\in\cP_i(\ub)}\alpha_{i,p}^2 = b\alpha^2 = o\big(1/\polylog(n)\big)$.
    \item For any $i\in\cS_0^-$, it holds that $\sum_{p\in\cP_i(\ub)}\alpha_{i,p}^2 = \Theta(1)$ and $\sum_{p\in\cP_i(\vb)}\alpha_{i,p}^2 = b\alpha^2 = o\big(1/\polylog(n)\big)$.
    \item For any $i\in\cS_1^+\cup\cS_1^-$, it holds that $\sum_{p\in\cP_i(\ub)}\alpha_{i,p}^2 = b\alpha^2 = o\big(1/\polylog(n)\big)$ and $\sum_{p\in\cP_i(\vb)}\alpha_{i,p}^2 = b\alpha^2 = o\big(1/\polylog(n)\big)$
\end{itemize}
Therefore, we have the following results regarding the relation between $\la\wb_{k,r}^{(t)},\vb\ra$ and $\la\wb_{k,r}^{(t)},\ub\ra$.

\begin{lemma}\label{lemma:update_feature_relation_phase2}
Let $T_1'=O\big(1/(\eta\rho b\alpha^2)\big)$ be a  quantity that is greater than $T_0$, then for any $t\in [T_0, T_1]$, there exists an absolute constant $C$ such that
\begin{align*}
\frac{|\la \wb_{1,r}^{(t)}, \vb \ra|}{|\la \wb_{1,r}^{(t)}, \ub\ra|}\ge C\cdot \frac{|\la \wb_{1,r}^{(T_0)}, \vb \ra|}{|\la \wb_{1,r}^{(T_0)}, \ub\ra|}=\tilde\Omega\bigg(\frac{1}{\sigma_0}\bigg),\quad\mbox{and}\quad
\frac{|\la \wb_{2,r}^{(t)}, \ub \ra|}{|\la \wb_{2,r}^{(t)}, \vb\ra|}\ge C\cdot \frac{|\la \wb_{2,r}^{(T_0)}, \ub \ra|}{|\la \wb_{2,r}^{(T_0)}, \vb\ra|}=\tilde\Omega\bigg(\frac{1}{\sigma_0}\bigg).
\end{align*}
\end{lemma}
\begin{proof}
Based on the update rules in \eqref{eq:update_feature_v_phase2} and \eqref{eq:update_feature_u_phase2}, we have
\begin{align*}
\la\wb_{1,r}^{(t+1)},\vb\ra  &= \la\wb_{1,r}^{(t)},\vb\ra\cdot\bigg[1 + \frac{\eta}{n}\cdot\bigg(\Theta(1)\cdot\sum_{i\in\cS_0^+}\ell_{1,i}^{(t)} +o\big(1/\polylog(n)\big)\cdot\sum_{i\in\cS_0^-}\ell_{1,i}^{(t)}\pm O(\rho n b\alpha^2) \bigg)\bigg];\notag\\
\la\wb_{1,r}^{(t+1)},\ub\ra  &= \la\wb_{1,r}^{(t)},\ub\ra\cdot\bigg[1 + \frac{\eta}{n}\cdot\bigg(\Theta(1)\cdot\sum_{i\in\cS_0^-}\ell_{1,i}^{(t)} +o\big(1/\polylog(n)\big)\cdot\sum_{i\in\cS_0^+}\ell_{1,i}^{(t)}\pm O(\rho n b\alpha^2) \bigg)\bigg].
\end{align*}
where we use the fact that $|\ell_{k,i}^{(t)}|\le 1$. 
This further implies that
\begin{align*}
\frac{|\la\wb_{1,r}^{(t+1)},\vb\ra|}{|\la\wb_{1,r}^{(t+1)},\ub\ra|} = \frac{|\la\wb_{1,r}^{(t)},\vb\ra|}{|\la\wb_{1,r}^{(t)},\ub\ra|}\cdot \underbrace{\frac{1 +\frac{\eta}{n}\cdot\bigg(\Theta(1)\cdot\sum_{i\in\cS_0^+}\ell_{1,i}^{(t)} +o\big(1/\polylog(n)\big)\cdot\sum_{i\in\cS_0^-}\ell_{1,i}^{(t)}\pm O(\rho n b\alpha^2) \bigg) }{1 + \frac{\eta}{n}\cdot\bigg(\Theta(1)\cdot\sum_{i\in\cS_0^-}\ell_{1,i}^{(t)} +o\big(1/\polylog(n)\big)\cdot\sum_{i\in\cS_0^+}\ell_{1,i}^{(t)}\pm O(\rho n b\alpha^2) \bigg)}}_{\star}.
\end{align*}
Note that we have $\ell_{1,i}^{(t)}>0$ for $i\in\cS_0^+$ and $\ell_{1,i}^{(t)}>0$ for $i\in\cS_0^-$. Then it can be readily verified that
\begin{align*}
\Theta(1)\cdot\sum_{i\in\cS_0^+}\ell_{1,i}^{(t)} +o\big(1/\polylog(n)\big)\cdot\sum_{i\in\cS_0^-}\ell_{1,i}^{(t)}\ge \Theta(1)\cdot\sum_{i\in\cS_0^-}\ell_{1,i}^{(t)} +o\big(1/\polylog(n)\big)\cdot\sum_{i\in\cS_0^+}\ell_{1,i}^{(t)}.
\end{align*}
Then we can get that
\begin{align*}
(\star)\ge 1 - \frac{O(\rho \eta b \alpha^2)}{1 + \frac{\eta}{n}\cdot\bigg(\Theta(1)\cdot\sum_{i\in\cS_0^-}\ell_{1,i}^{(t)} +o\big(1/\polylog(n)\big)\cdot\sum_{i\in\cS_0^+}\ell_{1,i}^{(t)}\pm O(\rho n b\alpha^2) \bigg)}\ge 1 - O(\rho \eta b \alpha^2). 
\end{align*}
Therefore we have for all $t\in[T_0, T_1']$,
\begin{align*}
\frac{|\la \wb_{1,r}^{(t)}, \vb \ra|}{|\la \wb_{1,r}^{(t)}, \ub\ra|}\ge \frac{|\la \wb_{1,r}^{(T_0)}, \vb \ra|}{|\la \wb_{1,r}^{(T_0)}, \ub\ra|}\cdot \big[1- O(\rho \eta b \alpha^2) \big]^{T_1' - T_0}\ge \frac{|\la \wb_{1,r}^{(T_0)}, \vb \ra|}{|\la \wb_{1,r}^{(T_0)}, \ub\ra|}\cdot \big[1- O(\rho \eta b \alpha^2) \big]^{O\big(\frac{1}{\rho \eta b \alpha^2}\big)}.
\end{align*}
Then applying the fact that $\big[1- O(\rho \eta b \alpha^2) \big]^{O\big(\frac{1}{\rho \eta b \alpha^2}\big)} \ge C$ holds
for some absolute constant $C$, we are able to complete the proof for bounding $\frac{|\la \wb_{2,r}^{(t)}, \ub \ra|}{|\la \wb_{2,r}^{(t)}, \vb\ra|}$. The results on $\frac{|\la \wb_{2,r}^{(t)}, \ub \ra|}{|\la \wb_{2,r}^{(t)}, \vb\ra|}$ can be obtained similarly. 
\end{proof}

In the next step, we will show that the learning of common features $\vb$ and $\ub$ will not be too large, i.e., exceeding the $\polylog(n)$ order. 
\begin{lemma}\label{lemma:upperbound_feature_phase2}
Let $T_1'$ be the same quantity defined in Lemma \ref{lemma:update_feature_relation_phase2}, we have for all $t\in[T_0, T_1']$, it holds that
\begin{align*}
|\la\wb_{1,r}^{(t+1)},\vb\ra|, |\la\wb_{2,r}^{(t+1)},\ub\ra| \le O\big(\polylog(n)\big).
\end{align*}
\end{lemma}
\begin{proof}[Proof of Lemma \ref{lemma:upperbound_feature_phase2}]
Based on the update rules in \eqref{eq:update_feature_v_phase2} and \eqref{eq:update_feature_u_phase2}, we have
\begin{align*}
\la\wb_{1,r}^{(t+1)},\vb\ra  &= \la\wb_{1,r}^{(t)},\vb\ra\cdot\bigg[1 + \frac{\eta}{n}\cdot\bigg(\Theta(1)\cdot\sum_{i\in\cS_0^+}\ell_{1,i}^{(t)} +o\big(1/\polylog(n)\big)\cdot\sum_{i\in\cS_0^-}\ell_{1,i}^{(t)}\pm O(\rho n b\alpha^2) \bigg)\bigg].
\end{align*}
Using the fact that $\ell_{1,i}^{(t)}<0$ for all $i\in\cS_0^{-}$, we further have
\begin{align*}
(\la\wb_{1,r}^{(t+1)},\vb\ra)^2  &\le (\la\wb_{1,r}^{(t)},\vb\ra)^2\cdot\bigg[1 + \frac{\eta}{n}\cdot\bigg(\Theta(1)\cdot\sum_{i\in\cS_0^+}\ell_{1,i}^{(t)} + O(\rho n b\alpha^2) \bigg)\bigg]^2\notag\\
& = (\la\wb_{1,r}^{(t)},\vb\ra)^2\cdot\bigg[1 + \frac{\eta}{n}\cdot\bigg(\Theta(1)\cdot\sum_{i\in\cS_0^+}\ell_{1,i}^{(t)} + O(\rho n b\alpha^2) \bigg)\bigg],
\end{align*}
where the second equality holds since $(1+o(1))^2= 1+o(1)$. Further take a summation over $r\in[m]$ leads to
\begin{align}\label{eq:feature_learning_v_allneurons_phase2}
\sum_{r=1}^m (\la\wb_{1,r}^{(t+1)},\vb\ra)^2 \le \bigg[\sum_{r=1}^m (\la\wb_{1,r}^{(t)},\vb\ra)^2\bigg]\cdot \bigg[1 + \frac{\eta}{n}\cdot\bigg(\Theta(1)\cdot\sum_{i\in\cS_0^+}\ell_{1,i}^{(t)} + O(\rho n b\alpha^2) \bigg)\bigg].
\end{align}
Similarly, we can also get that
\begin{align}\label{eq:feature_learning_u_allneurons_phase2}
\sum_{r=1}^m (\la\wb_{2,r}^{(t+1)},\ub\ra)^2 \le \bigg[\sum_{r=1}^m (\la\wb_{1,r}^{(t)},\ub\ra)^2\bigg]\cdot \bigg[1 + \frac{\eta}{n}\cdot\bigg(\Theta(1)\cdot\sum_{i\in\cS_0^-}\ell_{1,i}^{(t)} + O(\rho n b\alpha^2) \bigg)\bigg].
\end{align}
Regarding the loss derivative $\ell_{1,i}^{(t)}$, we can get that for any $i\in\cS_0^+$,
\begin{align}\label{eq:upperbound_lossderivate}
\ell_{1,i}^{(t)} &= 1 - \mathrm{Logit}_1(\Wb^{(t)}; \xb_i)=\frac{\exp\big[F_2(\Wb^{(t)};\xb_i)-F_1(\Wb^{(t)};\xb_i)\big]}{1 + \exp\big[F_2(\Wb^{(t)};\xb_i)-F_1(\Wb^{(t)};\xb_i)\big]}\le \exp\big[F_2(\Wb^{(t)};\xb_i)-F_1(\Wb^{(t)};\xb_i)\big]
\end{align}
Before moving to the analysis on the feature, we first show that the model weight corresponding to the wrong label will not learn the noise of the data, i.e., $|\la\wb_{2,r}^{(t)},\bxi_s^{(q)}\ra|$
will be very small for all $q\in[P]$ and $s\in\cS_0^+$. Particularly, we have the following by \eqref{eq:update_noise_simplified}
\begin{align*}
\max_{r,s}|\la\wb_{2,r}^{(t+1)},\bxi_s^{(q)}\ra |&\le \max_{r,s}|\la\wb_{2,r}^{(t)},\bxi_s^{(q)}\ra| \cdot\bigg[ 1 + \frac{\eta}{n}\cdot \ell_{2,s}^{(t)}\cdot\tilde\Theta(d\sigma_p^2)+ \frac{\eta}{n}\cdot\sum_{i\neq s || p\neq q} |\ell_{2,s}^{(t)}|\cdot \tilde O\big(d^{1/2}\sigma_p^2\big)\bigg]\notag\\
&\le \max_{r,s}|\la\wb_{2,r}^{(t)},\bxi_s^{(q)}\ra| \cdot\bigg[ 1 +  \frac{\eta}{n}\cdot\tilde O\big(nPd^{1/2}\sigma_p^2\big)\bigg],
\end{align*}
where the second inequality is due to $|\ell_{k,i}^{(t)}|\le 1$ and $\ell_{2,s}^{(t)}<0$ for $s\in\cS_0^+$. Therefore, we can get that for all $t\in\big[T_0, T_1'\big]$, where $T_1'\le \tilde O\big(1/(\eta Pd^{1/2}\sigma_p^2)\big)$, that
\begin{align}\label{eq:bound_noise_strong_phase2}
\max_{r,s}|\la\wb_{2,r}^{(t)},\bxi_s^{(q)}\ra|&\le \max_{r,s}|\la\wb_{2,r}^{(T_0)},\bxi_s^{(q)}\ra|\cdot \bigg[ 1 +  \tilde O\big(\eta Pd^{1/2}\sigma_p^2\big)\bigg]^{\tilde O\big(\frac{1}{\eta Pd^{1/2}\sigma_p^2}\big)}\notag\\
&\le C\cdot \max_{r,s}|\la\wb_{2,r}^{(T_0)},\bxi_s^{(q)}\ra| =\tilde O (d^{1/2}\sigma_p\sigma_0),
\end{align}
where the last equality is by Lemma \ref{lemma:learning_phase1}. Therefore, we can get the following bound on $F_2(\Wb^{(t)};\xb_i)-F_1(\Wb^{(t)};\xb_i)$ for any $i\in \cS_0^+$,
\begin{align*}
F_2(\Wb^{(t)};\xb_i)-F_1(\Wb^{(t)};\xb_i)& = \sum_{r=1}^m\sum_{p=1}^P(\la\wb_{2,r}^{(t)},\xb_i^{(p)}\ra)^2-\sum_{r=1}^m\sum_{p=1}^P(\la\wb_{1,r}^{(t)},\xb_i^{(p)}\ra)^2\notag\\
&\le \sum_{r=1}^m\sum_{p\in\cP_i(\vb)}(\la\wb_{2,r}^{(t)},\vb\ra)^2 + \alpha^2\sum_{r=1}^m\sum_{p\in\cP_i(\ub)}(\la\wb_{2,r}^{(t)},\ub\ra)^2 \notag\\
&\qquad+ \sum_{r=1}^m\sum_{p\in\cP_i(\bxi)}(\la\wb_{2,r}^{(t)},\bxi_i^{(p)}\ra)^2 - \sum_{r=1}^m(\la\wb_{1,r}^{(t)},\vb\ra)^2.
\end{align*}
Then by Lemma \ref{lemma:update_feature_relation_phase2} and \eqref{eq:bound_noise_strong_phase2}, we can further get that
\begin{align*}
F_2(\Wb^{(t)};\xb_i)-F_1(\Wb^{(t)};\xb_i)\le  O(b\alpha^2)\cdot \sum_{r=1}^m(\la\wb_{2,r}^{(t)},\ub\ra)^2-\sum_{r=1}^m(\la\wb_{1,r}^{(t)},\vb\ra) + \tilde O\big(mPd\sigma_p^2\sigma_0^2\big),
\end{align*}
where we use the fact that $|\cP_i(\ub)|\le b$ and $(\la\wb_{2,r}^{(t)},\vb\ra/\la\wb_{2,r}^{(t)},\ub\ra)^2=o(\alpha^2)$ by Lemma \ref{lemma:update_feature_relation_phase2}. 
This further implies the following according to \eqref{eq:upperbound_lossderivate}: for all $i\in\cS_0^+$,
\begin{align*}
\ell_{1,i}^{(t)}&\le \exp\big[F_2(\Wb^{(t)};\xb_i)-F_1(\Wb^{(t)};\xb_i)\big]\notag\\
&\le 2 \exp\bigg[O(b\alpha^2)\cdot \sum_{r=1}^m(\la\wb_{2,r}^{(t)},\ub\ra)^2-\sum_{r=1}^m(\la\wb_{1,r}^{(t)},\vb\ra)\bigg],
\end{align*}
where we use the fact that $mPd\sigma_p^2\sigma_0^2=o(1)$. Similarly, we can also get that for all $i\in\cS_0^-$,
\begin{align*}
\ell_{2,i}^{(t)}\le 2 \exp\bigg[O(b\alpha^2)\cdot \sum_{r=1}^m(\la\wb_{1,r}^{(t)},\vb\ra)^2-\sum_{r=1}^m(\la\wb_{2,r}^{(t)},\ub\ra)\bigg].
\end{align*}
Consequently, let $a_t:=\sum_{r=1}^m(\la\wb_{1,r}^{(t+1)},\vb\ra)^2$ and $b_t:=\sum_{r=1}^m(\la\wb_{2,r}^{(t+1)},\ub\ra)^2$, 
further applying \eqref{eq:feature_learning_v_allneurons_phase2} and \eqref{eq:feature_learning_u_allneurons_phase2} gives
\begin{align*}
a_{t+1} &\le a_t \cdot \Big[1 + \Theta(\eta)\cdot \exp\big[O(b\alpha^2)\cdot b_t -a_t\big] +O(\eta \rho  b \alpha^2) \Big]\notag\\
b_{t+1} &\le b_t \cdot \Big[1 + \Theta(\eta)\cdot \exp\big[O(b\alpha^2)\cdot a_t -b_t\big] +O(\eta \rho  b \alpha^2)\Big].
\end{align*}
Then we will first prove a weaker argument on $a_t$ and $b_t$: for all $t\le T_1'$ it holds that $a_t, b_t = o\big(1/(b\alpha^2)\big)$. In particular, we will apply standard induction techniques. First, it is easy to verify that this condition holds for $t=T_0$ according to Lemma \ref{lemma:results_phase1}. Then assuming this condition holds for all $\tau\le t$, we have $\exp\big[O(b\alpha^2)\cdot b_t\big], \exp\big[O(b\alpha^2)\cdot a_t\big]=\Theta(1)$ and thus
\begin{align}
a_{\tau+1} &\le a_\tau \cdot \Big[1 + \Theta(\eta)\cdot \exp(-a_\tau) +O(\eta \rho  b \alpha^2) \Big],\notag\\
b_{\tau+1} &\le b_\tau \cdot \Big[1 + \Theta(\eta)\cdot \exp( -b_\tau) +O(\eta \rho  b \alpha^2)\Big],
\end{align}
for all $\tau\in[T_0, t]$. 
Then by Lemma \ref{lemma:technical_lemma1}, we can immediately get that 
\begin{align*}
a_{t+1}\le O\bigg( \log\bigg(\frac{1}{\rho b \alpha^2}\cdot e^{t\eta \rho b \alpha^2}\bigg)\bigg).
\end{align*}
Then recall that $T_1' = O\big(1/(\eta\rho b \alpha^2)\big)$ and $t\le T_1'$, we can further get $a_{t+1}, b_{t+1} = O\big(\log(\frac{1}{\rho b \alpha^2})\big)$, which verify the hypothesis that $a_{t+1}, b_{t+1}\le o(1/(b\alpha^2))$. Moreover, recall the definitions of $a_t$ and $b_t$: $a_t:=\sum_{r=1}^m(\la\wb_{1,r}^{(t+1)},\vb\ra)^2$ and $b_t:=\sum_{r=1}^m(\la\wb_{2,r}^{(t+1)},\ub\ra)^2$, we can further get that for all $r\in[m]$,
\begin{align*}
|\la\wb_{1,r}^{(t+1)},\vb\ra| = \tilde O\bigg(\log^{1/2}\bigg(\frac{1}{\rho b \alpha^2}\bigg)\bigg)=O\big(\polylog(n)\big),
\end{align*}
and $|\la\wb_{2,r}^{(t+1)},\ub\ra| = O\big(\polylog(n)\big)$. This completes the proof.

\end{proof}

\begin{lemma}\label{lemma:technical_lemma1}
Let $\{a_t\}_{t=0,\dots,}$ be a sequence with $a_0\in[0, 1]$ that satisfies
\begin{align*}
a_{t+1}  = a_t \cdot [1 + c_1\cdot e^{-a_t} + c_2],
\end{align*}
where $c_1$ and $c_2$ are two constants satisfying $c_1, c_2\in[0, 1]$ and $c_2\le c_1$. Then it holds that
\begin{align*}
a_t \le O\Big(\log(c_1/c_2)\cdot e^{2c_2 t}\Big).
\end{align*}
\end{lemma}
\begin{proof}[Proof of Lemma \ref{lemma:technical_lemma1}]
Note that $c_2\le c_1$, we will then consider two cases: (1) $c_1 e^{-a_t}\ge c_2$ and (2) $c_1 e^{-a_t}< c_2$. Then case (2) will occur after case (1) since $a_t$ is strictly increasing. Regarding case (1), it is easy to see that $a_t \le \log(c_1/c_2)$ by the condition that $c_1 e^{-a_t}\ge c_2$. For case (2),
let $t_0$ be the first iteration $t$ that $c_1 e^{-a_t}<c_2$, we can get that $a_{t_0} = O\big(\log(c_1/c_2)\big)$ and then for all $t>t_0$
\begin{align*}
a_{t+1} \le a_t\cdot[1+2c_2], 
\end{align*}
which implies that
\begin{align*}
a_t \le a_{t_0}\cdot [1+2c_2]^{t-T_0}\le O\Big(\log(c_1/c_2)\cdot e^{2c_2 t}\Big).
\end{align*}
Combining the results for case (1) and case (2), we can complete the proof.

\end{proof}

Then we will focus on the rare feature data. Note that in the early stage of the second phase, their corresponding loss derivatives $\ell_{k,i}^{(t)}$'s are still in the constant order. The following Lemma summarizes the learning of rare features and noises for the rare feature data.
\begin{lemma}\label{lemma:learning_noise_weakdata_phase2}
Let $T_1= O\big(\frac{n\log(1/(\sigma_0d^{1/2}\sigma_p))}{d\sigma^2\eta}\big)$ be a quantity that satisfies $T_0<T_1<T_1'$. Then for any $t\in[T_0, T_1]$, it holds that
\begin{align*}
\la\wb_{1,r}^{(t+1)},\vb'\ra &= \la\wb_{1,r}^{(t)},\vb'\ra\cdot \big[1 + \Theta(\rho\eta)\big],\  \la\wb_{2,r}^{(t+1)},\vb'\ra = \la\wb_{2,r}^{(t)},\vb'\ra\cdot \big[1 - \Theta(\rho\eta)\big]\notag\\
\la\wb_{1,r}^{(t+1)},\ub'\ra &= \la\wb_{1,r}^{(t)},\ub'\ra\cdot \big[1 - \Theta(\rho\eta)\big],\  \la\wb_{2,r}^{(t+1)},\ub'\ra = \la\wb_{2,r}^{(t)},\ub'\ra\cdot \big[1 + \Theta(\rho\eta)\big]
\end{align*}
Besides, for any $i\in\cS_1^+\cup\cS_1^-$ and $k=y_s$, we have
\begin{align*}
\max_{r,p}|\la\wb_{k,r}^{(t+1)},\bxi_s^{(q)}\ra| = \max_{r,p}|\la\wb_{k,r}^{(t)},\bxi_s^{(q)}\ra|\cdot \bigg[1 + \frac{\eta}{n}\cdot\tilde\Theta(d\sigma_p^2)\bigg];
\end{align*}
for $k\neq y_s$,
\begin{align*}
\max_{r,p}|\la\wb_{k,r}^{(t+1)},\bxi_s^{(q)}\ra| = \tilde O(d^{-1/2}n).
\end{align*}
\end{lemma}
\begin{proof}
The proof is similar to that of Lemma \ref{lemma:learning_phase1}, except  the proof for the dynamics of $\la\wb_{k,r}^{(t+1)},\bxi_s^{(q)}\ra$.  
First, by standard concentration argument, we can get with probability $1-1/\poly(n)$, for all $\bxi_i^{(p)}\in\{\bxi\}$, it holds that
\begin{align*}
d\sigma_p^2 -\tilde O\big(d^{1/2}\sigma_p^2\big)\le \|\bxi_i^{(p)}\|_2^2 \le d\sigma_p^2 + \tilde O\big(d^{1/2}\sigma_p^2\big).
\end{align*}
Then by \eqref{eq:update_noise_simplified}, we can get
\begin{align*}
\la\wb_{k,r}^{(t+1)},\bxi_s^{(q)}\ra 
& = \la\wb_{k,r}^{(t)},\bxi_s^{(q)}\ra \cdot\bigg[ 1 + \frac{2\eta}{n}\cdot \ell_{k,s}^{(t)}\cdot\|\bxi_s^{(q)}\|_2^2\bigg]  \pm \frac{2\eta}{n}\cdot\sum_{i\neq s || p\neq q} |\ell_{k,i}^{(t)}|\cdot|\la\wb_{k,r}^{(t)},\bxi_i^{(q)}\ra|\cdot \tilde O\big(d^{1/2}\sigma_p^2\big)\notag\\
&=\la\wb_{k,r}^{(t)},\bxi_s^{(q)}\ra\cdot \bigg[ 1 + \frac{2\eta}{n}\cdot \ell_{k,s}^{(t)}\cdot d\sigma_p^2\bigg]\pm \frac{2\eta}{n}\cdot\sum_{\bxi_i^{(q)}\in\{\bxi\}} |\ell_{k,i}^{(t)}|\cdot|\la\wb_{k,r}^{(t)},\bxi_i^{(q)}\ra|\cdot \tilde O\big(d^{1/2}\sigma_p^2\big).
\end{align*}
Then let $\zeta=O\big(1/\polylog(n)\big)$ be some user-defined constant, then let $T'$ be the smallest iteration number such that $\max_{i}|\ell_{k,i}^{(t)}|\in[0.5-\zeta,0.5+\zeta]$. Then we can get for any $t\le T'$ and any $i,r$, 
\begin{align}\label{eq:update_noise_proof}
\la\wb_{k,r}^{(t+1)},\bxi_s^{(q)}\ra 
&=\la\wb_{k,r}^{(t)},\bxi_s^{(q)}\ra\cdot \bigg[ 1 + \frac{2\eta}{n}\cdot (0.5\pm \zeta)\cdot d\sigma_p^2\bigg]\pm \frac{2\eta}{n}\cdot\sum_{\bxi_i^{(q)}\in\{\bxi\}} |\ell_{k,i}^{(t)}|\cdot|\la\wb_{k,r}^{(t)},\bxi_i^{(q)}\ra|\cdot \tilde O\big(d^{1/2}\sigma_p^2\big).
\end{align}
Then we will prove the main arguments via mathematical induction, including the following hypothesis: 
\begin{itemize}
\item For all $i\in\cS_1^+\cup\cS_1^-$, it holds that $\max_{r,q}|\la\wb_{k,r}^{(t)},\bxi_i^{(p)}\ra|\ge \frac{1}{n^{0.1}}\cdot \max_{r,i',q}|\la\wb_{k,r}^{(t)},\bxi_{i'}^{(q)}\ra|$
\item $\max_{r,q}|\la\wb_{k,r}^{(t+1)},\bxi_s^{(q)}\ra| 
=\max_{r,q}|\la\wb_{k,r}^{(t)},\bxi_s^{(q)}\ra|\cdot \bigg[ 1 + \frac{2\eta}{n}\cdot (0.5\pm 2\zeta)\cdot d\sigma_p^2\bigg]$.
\end{itemize}
Then it is clear that the first argument holds for $t=T_0$ as with probability at least $1-1/\poly(n)$ we have $\max_{r,q}|\la\wb_{k,r}^{(T_0)},\bxi_i^{(p)}\ra| = \tilde\Theta(\sigma_0d^{1/2}\sigma_p)$ and $\max_{r,i',q}|\la\wb_{k,r}^{(t)},\bxi_{i'}^{(q)}\ra|=\tilde\Theta(\sigma_0d^{1/2}\sigma_p)$, which implies that $|\la\wb_{k,r}^{(t)},\bxi_i^{(p)}\ra|\ge \frac{1}{\polylog(n)}\cdot \max_{r,i',q}|\la\wb_{k,r}^{(t)},\bxi_{i'}^{(q)}\ra|$.

Besides, given the first argument, we have
\begin{align*}
\sum_{\bxi_i^{(q)}\in\{\bxi\}} |\ell_{k,i}^{(t)}|\cdot|\la\wb_{k,r}^{(t)},\bxi_i^{(q)}\ra|\cdot \tilde O\big(d^{1/2}\sigma_p^2\big)&\le \tilde O(nPd^{1/2}\sigma_p^2)\cdot \max_{r,i',q}|\la\wb_{k,r}^{(t)},\bxi_{i'}^{(q)}\ra|\notag\\
&\le O(n^{1.1}Pd^{1/2}\sigma_p^2)\cdot \max_{r,q}|\la\wb_{k,r}^{(t)},\bxi_i^{(p)}\ra|\notag\\
&\le \zeta \cdot d\sigma_p^2\cdot\max_{r,q}|\la\wb_{k,r}^{(t)},\bxi_i^{(p)}\ra|,
\end{align*}
where we use the fact that $d^{-1/2}n^{1.1}P = o\big(1/\polylog(n)\big)=o(\zeta)$. Then by \eqref{eq:update_noise_proof}, we can directly obtain the second argument.

Now we will verify the hypotheses by induction. First, similar to the previous derivation, the first argument at the $t$-th iteration can directly imply the second argument at the $t+1$-th iteration. Then it remains to verify the first argument. In fact, given the second argument, we have for any $i$ and $i'$ and $\tau\le t$,
\begin{align*}
\frac{\max_{r,q}|\la\wb_{k,r}^{(\tau+1)},\bxi_i^{(p)}\ra|}{\max_{r,q}|\la\wb_{k,r}^{(\tau+1)},\bxi_{i'}^{(p)}\ra|}&\le \frac{1+\frac{2\eta}{n}\cdot (0.5 + 2\zeta)\cdot d\sigma_p^2}{1+\frac{2\eta}{n}\cdot (0.5 - 2\zeta)\cdot d\sigma_p^2} \cdot \frac{\max_{r,q}|\la\wb_{k,r}^{(\tau)},\bxi_i^{(p)}\ra|}{\max_{r,q}|\la\wb_{k,r}^{(\tau)},\bxi_{i'}^{(p)}\ra|}\notag\\
&\le \bigg(1+\frac{\eta\zeta d\sigma_p^2}{n}\bigg)^{T_1}\cdot \frac{\max_{r,q}|\la\wb_{k,r}^{(T_1)},\bxi_i^{(p)}\ra|}{\max_{r,q}|\la\wb_{k,r}^{(T_1)},\bxi_{i'}^{(p)}\ra|}.
\end{align*}
Therefore, using the fact that $T_1 = O\big(\frac{n\log(1/(\sigma_0d^{1/2}\sigma_p))}{d\sigma^2\eta}\big)$, setting $\zeta = 1/\log^2(1/(\sigma_0d^{1/2}\sigma_p)$, we can directly get that
\begin{align*}
\frac{\max_{r,q}|\la\wb_{k,r}^{(t+1)},\bxi_i^{(p)}\ra|}{\max_{r,q}|\la\wb_{k,r}^{(t+1)},\bxi_{i'}^{(p)}\ra|}
\le \bigg(1+\frac{\eta\zeta d\sigma_p^2}{n}\bigg)^{T_1}\cdot O(\polylog(n))= o(n^{0.1}).
\end{align*}
Note that the above holds for all $i$ and $i'$, taking $i' = \arg\max_i|\la\wb_{k,r}^{(t+1)},\bxi_{i'}^{(p)}\ra|$ directly completes the verification of the first argument.

The proof for $\max_{r,q}|\la\wb_{k,r}^{(t)},\bxi_s^{(q)}\ra|$ with $k\neq y_s$, we have the following by \eqref{eq:update_noise_simplified},
\begin{align*}
\max_{r,q}|\la\wb_{k,r}^{(t+1)},\bxi_s^{(q)}\ra| 
& \le \max_{r,q}|\la\wb_{k,r}^{(t)},\bxi_s^{(q)}\ra |+ \frac{2\eta}{n}\cdot\sum_{i\neq s || p\neq q} |\ell_{k,i}^{(t)}|\cdot|\la\wb_{k,r}^{(t)},\bxi_i^{(q)}\ra|\cdot \tilde O\big(d^{1/2}\sigma_p^2\big)\\
&\le \max_{r,q}|\la\wb_{k,r}^{(T_0)},\bxi_s^{(q)}\ra | + T_1 P\eta \cdot \tilde O(d^{1/2}\sigma_p^2)\notag\\
& = \tilde O(d^{-1/2}n),
\end{align*}
where we use the fact that for all $t\le T_1$, it holds that $\max_{i,r,q}|\la\wb_{k,r}^{(t+1)},\bxi_s^{(q)}\ra|=\tilde O(1)$.


\end{proof}

\begin{lemma}[End of Phase 2]\label{lemma:results_endofphase2}
Let $T_1$ be the same quantity defined in Lemma \ref{lemma:learning_noise_weakdata_phase2}, we have for all $i\in\cS_0^+\cup\cS_0^-$,
\begin{align*}
|\la\wb_{1,r}^{(T_1)}, \vb\ra|, |\la\wb_{2,r}^{(T_1)}, \ub\ra| = \tilde\Theta(1),\quad |\la\wb_{2,r}^{(T_1)}, \vb\ra|, |\la\wb_{1,r}^{(T_1)}, \ub\ra| = \tilde O(\sigma_0),\quad |\la\wb_{k,r}^{(T_1)}, \bxi_i^{(p)}\ra|=\tilde O\big(d^{-1/2}n\big);
\end{align*}
for all $i\in\cS_1^+\cup\cS_1^-$,
\begin{align*}
|\la\wb_{1,r}^{(T_1)}, \vb'\ra|, |\la\wb_{2,r}^{(T_1)}, \ub'\ra| = \tilde O(\sigma_0),\quad |\la\wb_{2,r}^{(T_1)}, \vb'\ra|, |\la\wb_{1,r}^{(T_1)}, \ub'\ra| = \tilde O(\sigma_0),\quad |\la\wb_{y_i,r}^{(T_1)}, \bxi_i^{(p)}\ra|=\tilde\Theta(1).
\end{align*}

\end{lemma}
\begin{proof}
The proof of this lemma is simply a combination of Lemmas \ref{lemma:learning_noise_weakdata_phase2} and \ref{lemma:learning_noise_weakdata_phase2}, where we only need to verify the bound for $|\la\wb_{k,r}^{(T_1)}, \vb'\ra|$ and $|\la\wb_{k,r}^{(T_1)}, \ub'\ra|$. This can be done as follows:
\begin{align*}
|\la\wb_{k,r}^{(T_1)}, \vb'\ra|\le \big[1+\Theta(\rho\eta)\big]^{T_1}\cdot |\la\wb_{k,r}^{(T_0)}, \vb'\ra| \le \exp\big[\tilde O(n\rho /(d\sigma^2))\big]\cdot|\la\wb_{k,r}^{(T_0)}, \vb'\ra| = \tilde O(\sigma_0), 
\end{align*}
where we use the fact that $\rho n = o(d\sigma_p^2)$. The proof for $|\la\wb_{k,r}^{(T_1)}, \vb'\ra|$ will be similar and thus is ommited here.

\end{proof}
\paragraph{Phase 3. Training until convergence.} 
In this phase, we will show that the feature learning and noise learning in \textbf{Phase 2} will be maintained. Particularly, we first make the following hypothesis and then verify them via mathematical induction. 
\begin{hypothesis}\label{hypothesis}
For all $t = \poly(n)$ that is greater than $T_1$, it holds that
\begin{enumerate}[label=(\alph*)]
    \item \label{item:strongfeat_correct} We have $\sum_{r=1}^m(\la\wb_{1,r}^{(t)}, \vb\ra)^2 = \tilde\Theta(1)$ and $\sum_{r=1}^m\big(\la\wb_{2,r}^{(t)},\ub\ra\big)^2 = \tilde\Theta(1)$.
    \item \label{item:strongfeat_incorrect} We have $|\la\wb_{1,r}^{(t)}, \ub\ra|=O\big(|\la\wb_{1,r}^{(T_1)}, \ub\ra|\big)=o\big(\frac{1}{\polylog(n)}\big)$ and $|\la\wb_{2,r}^{(t)}, \vb\ra|=O\big(|\la\wb_{2,r}^{(T_1)}, \vb\ra|\big)=o\big(\frac{1}{\polylog(n)}\big)$.
    \item \label{item:weakfeat} We have $|\la\wb_{k,r}^{(t)}, \vb'\ra|=O\big(|\la\wb_{k,r}^{(T_1)}, \vb'\ra|\big)=o\big(\frac{1}{\polylog(n)}\big)$ and $|\la\wb_{k,r}^{(t)}, \ub'\ra|=O\big(|\la\wb_{k,r}^{(T_1)}, \ub'\ra|\big)=o\big(\frac{1}{\polylog(n)}\big)$.
    \item \label{item:strongfeat_noise} For all $i\in\cS_0^+\cup\cS_0^-$, we have
    $|\la\wb_{k,r}^{(t)},\bxi_i^{(p)}\ra|=O\big(|\la\wb_{k,r}^{(T_1)},\bxi_i^{(p)}\ra|\big)=o\big(\frac{1}{\polylog(n)}\big)$.
    \item \label{item:weakfeat_noise_correct} For all $i\in\cS_1^+$, we have
    $\sum_{r=1}^m\sum_{p\in\cP_i(\bxi)}(\la\wb_{1,r}^{(t)}, \bxi_i^{(p)}\ra)^2 = \tilde\Theta(1)$; for all $i\in\cS_1^-$, we have $\sum_{r=1}^m\sum_{p\in\cP_i(\bxi)}(\la\wb_{2,r}^{(t)},\bxi_i^{(p)}\ra)^2 = \tilde\Theta(1)$.
    \item \label{item:weakfeat_noise_incorrect} For all $i\in\cS_1^+$, we have
    $|\la\wb_{2,r}^{(t)}, \bxi_i^{(p)}\ra| =o\big(\frac{1}{\polylog(n)}\big)$; for all $i\in\cS_1^-$, we have $|\la\wb_{1,r}^{(t)}, \bxi_i^{(p)}\ra| =o\big(\frac{1}{\polylog(n)}\big)$.
\end{enumerate}
\end{hypothesis}

The hypothesis will be verified via induction. First, it is clear that all hypothesis are satisfied at $t=T_1$ according to Lemma \ref{lemma:results_endofphase2}. Then, the following lemma is useful in the entire proof.
\begin{lemma}\label{lemma:bound_sum_lossderivative}
Assuming all hypothesis in Hypothesis \ref{hypothesis} hold for $\tau\in[0, t]$, then we have for all $k\in[2]$,
\begin{align*}
\sum_{\tau=T_1}^{t}\sum_{i\in\cS_0^+\cup\cS_0^-} |\ell_{k,i}^{(\tau)}| = \tilde O\bigg(\frac{n}{\eta}\bigg),\quad\mbox{and}\quad \sum_{\tau=T_1}^{t}\sum_{i\in\cS_1^+\cup\cS_1^-}|\ell_{k,i}^{(\tau)}|\le \tilde O\bigg(\frac{\rho n^2}{d\sigma_p^2 \eta}\bigg),
\end{align*}
moreover, for any $i\in\cS_0^+\cup\cS_0^-$, we have
\begin{align*}
\sum_{t=T_1}^t|\ell_{k,i}^{(\tau)}|=\tilde O(1/\eta).
\end{align*}

\end{lemma}
\begin{proof}[Proof of Lemma \ref{lemma:bound_sum_lossderivative}]
By \eqref{eq:update_features}, we have 
\begin{align}\label{eq:total_feature_learning_phase3}
\sum_{r=1}^m(\wb_{1,r}^{(\tau+1)},\vb)^2 &= \sum_{r=1}^m(\wb_{1,r}^{(\tau)},\vb)^2\cdot \bigg[1+\frac{2\eta}{n}\cdot \sum_{i\in[n]}\ell_{1,i}^{(\tau)}\sum_{p\in\cP_i(\vb)}\alpha_{i,p}^2\bigg]^2\notag\\
&\ge \sum_{r=1}^m(\wb_{1,r}^{(\tau)},\vb)^2\cdot \bigg[1+\Theta\bigg(\frac{\eta}{n}\bigg)\cdot \sum_{i\in\cS_0^+}|\ell_{1,i}^{(\tau)}| - \Theta\bigg(\frac{b\eta\alpha^2}{n}\bigg)\cdot \sum_{i\not\in\cS_0^+}|\ell_{1,i}^{(\tau)}|\bigg];\notag\\
\sum_{r=1}^m(\wb_{2,r}^{(\tau+1)},\ub)^2 &= \sum_{r=1}^m(\wb_{2,r}^{(\tau)},\vb)^2\cdot \bigg[1+\frac{2\eta}{n}\cdot \sum_{i\in[n]}\ell_{2,i}^{(\tau)}\sum_{p\in\cP_i(\vb)}\la\wb_{2,r}^{(\tau)},\vb\ra\cdot\alpha_{i,p}^2\bigg]^2\notag\\
&\ge \sum_{r=1}^m(\wb_{2,r}^{(\tau)},\vb)^2\cdot \bigg[1+\Theta\bigg(\frac{\eta}{n}\bigg)\cdot \sum_{i\in\cS_0^-}|\ell_{2,i}^{(\tau)}| - \Theta\bigg(\frac{b\eta\alpha^2}{n}\bigg)\cdot \sum_{i\not\in\cS_0^-}|\ell_{2,i}^{(\tau)}|\bigg].
\end{align}
where we use the fact that $|\ell_{1,i}|=|\ell_{2,i}|$
Summing them up and further taking a summation over $\tau\in[T_1, t-1]$, applying Hypothesis \ref{hypothesis}\ref{item:strongfeat_correct} gives
\begin{align}\label{eq:bound_sum_lossderivatives1}
\Theta\bigg(\frac{\eta}{n}\bigg)\cdot\sum_{\tau=T_1}^{t-1}\sum_{i\in\cS_0^+\cup\cS_0^-}|\ell_{1,i}^{(\tau)}| - \Theta\bigg(\frac{b\eta\alpha^2}{n}\bigg)\cdot \sum_{\tau=T_1}^{t-1}\sum_{i=1}^n|\ell_{1,i}^{(\tau)}|\le \tilde O(1).
\end{align}
where we use the fact that $|\ell_{1,i}^{(t)}|=|\ell_{2,i}^{(t)}|$ and $\sum_{r=1}^m(\la\wb_{1,r}^{(t)}, \vb\ra)^2, \sum_{r=1}^m\big(\la\wb_{2,r}^{(t)},\ub\ra\big)^2 = \tilde\Theta(1)$.
Besides, by \eqref{eq:update_noise_simplified} and Hypotheses \ref{item:weakfeat_noise_correct} and \ref{item:weakfeat_noise_incorrect}, we know that the correct noise learning for different weak feature data will be different by at most 
$O(\polylog(n))$ factors, therefore, we can  get that
\begin{align}\label{eq:total_noise_learning_phase3}
\sum_{i\in\cS_1^+}\sum_{p\in[P]}(\la\wb_{1,r}^{(\tau+1)},\bxi_i^{(p)}\ra)^2 
&\ge  \sum_{i\in\cS_1^+}\sum_{p\in[P]}(\la\wb_{k,r}^{(\tau)},\bxi_i^{(p)}\ra)^2 \cdot\bigg[ 1 + \frac{2\eta}{n}\cdot \ell_{1,i}^{(\tau)}\cdot\|\bxi_i^{(p)}\|_2^2 - \frac{2\eta}{n}\cdot\sum_{i=1}^n |\ell_{1,i}^{(\tau)}|\cdot \tilde O\big(Pd^{1/2}\sigma_p^2\big)\bigg]^2\notag\\
&\ge \sum_{i\in\cS_1^+}\sum_{p\in[P]}(\la\wb_{1,r}^{(\tau)},\bxi_i^{(p)}\ra)^2 \cdot\bigg[ 1 + \tilde\Theta\bigg(\frac{\eta d\sigma_p^2}{n}\bigg)\cdot |\ell_{1,i}^{(\tau)}| - \tilde\Theta\bigg(\frac{\eta d^{3/2}\sigma_p^4 P}{n}\bigg)\cdot\sum_{i=1}^n |\ell_{1,i}^{(\tau)}|\bigg],
\end{align}
and similarly,
\begin{align*}
\sum_{i\in\cS_1^-}\sum_{p\in[P]}(\la\wb_{2,r}^{(\tau+1)},\bxi_i^{(p)}\ra)^2 
&\ge \sum_{i\in\cS_1^-}\sum_{p\in[P]}(\la\wb_{2,r}^{(\tau)},\bxi_i^{(p)}\ra)^2 \cdot\bigg[ 1 + \tilde\Theta\bigg(\frac{\eta d\sigma_p^2}{n}\bigg)\cdot |\ell_{2,i}^{(\tau)}| - \tilde\Theta\bigg(\frac{\eta d^{3/2}\sigma_p^4 P}{n}\bigg)\cdot\sum_{i=1}^n |\ell_{2,i}^{(\tau)}|\bigg].
\end{align*}
Therefore, taking a summation over $r\in[m]$ and $\tau\in[T_1, t-1]$, and using the Hypothesis  \ref{hypothesis}\ref{item:weakfeat_noise_correct}, we have
\begin{align}\label{eq:bound_sum_lossderivatives2}
\Theta\bigg(\frac{\eta d\sigma_p^2}{n}\bigg)\cdot\sum_{\tau=T_1}^{t-1}\sum_{i\in\cS_1^+\cup\cS_1^-}|\ell_{1,i}^{(\tau)}| - \Theta\bigg(\frac{\eta d^{3/2}\sigma_p^4P}{n}\bigg)\cdot \sum_{\tau=T_1}^{t-1}\sum_{i=1}^n|\ell_{1,i}^{(\tau)}|\le \tilde O(|\cS_1^+\cup\cS_1^-|)=\tilde O(\rho n).
\end{align}
Combining \eqref{eq:bound_sum_lossderivatives1} and \eqref{eq:bound_sum_lossderivatives2} and using the fact that $d\sigma_p^2=\omega(1)$ and $b\alpha^2=\omega(d^{3/2}\sigma_p^4P)$, we can get that
\begin{align*}
\Theta\bigg(\frac{\eta}{n}\bigg)\cdot\sum_{\tau=T_1}^{t-1}\sum_{i=1}^n|\ell_{1,i}^{(\tau)}| -\Theta\bigg(\frac{b\eta\alpha^2}{n}\bigg)\cdot\sum_{\tau=T_1}^{t-1}\sum_{i=1}^n|\ell_{1,i}^{(\tau)}|\le \tilde O(\rho n).
\end{align*}
Note that $b\alpha^2=o(1)$, the above inequality immediately implies that
\begin{align*}
\sum_{\tau=T_1}^{t-1}\sum_{i=1}^n|\ell_{1,i}^{(\tau)}|\le \tilde O\bigg(\frac{\rho n^2}{\eta}\bigg).
\end{align*}
We will further use this argument to sharpen our result. First, \eqref{eq:bound_sum_lossderivatives2} directly leads to
\begin{align*}
\Theta\bigg(\frac{\eta d\sigma_p^2}{n}\bigg)\cdot\sum_{\tau=T_1}^{t-1}\sum_{i\in\cS_1^+\cup\cS_1^-}|\ell_{1,i}^{(\tau)}|\le \tilde O(\rho n) + \tilde O\big(d^{3/2}\sigma_p^4 P \rho n\big) = \tilde O(\rho n),
\end{align*}
which implies that 
\begin{align*}
\sum_{\tau=T_1}^{t-1}\sum_{i\in\cS_1^+\cup\cS_1^-}|\ell_{1,i}^{(\tau)}|\le \tilde O\bigg(\frac{\rho n^2}{d\sigma_p^2 \eta}\bigg).
\end{align*}
Plugging the above inequality into \eqref{eq:bound_sum_lossderivatives2} and using the fact that $b\alpha^2=o(1)$ gives
\begin{align*}
\Theta\bigg(\frac{\eta}{n}\bigg)\cdot\sum_{\tau=T_1}^{t-1}\sum_{i\in\cS_0^+\cup\cS_0^-}|\ell_{1,i}^{(\tau)}| \le \tilde O(1) + \Theta\bigg(\frac{b\eta\alpha^2}{n}\bigg)\cdot \sum_{\tau=T_1}^{t-1}\sum_{i\in\cS_1^+\cup\cS_1^-}|\ell_{1,i}^{(\tau)}|\le \tilde O(1) + \tilde O\bigg(\frac{\rho n b\alpha^2}{d\sigma_p^2}\bigg)=\tilde O(1).
\end{align*}
where the last inequality is due to $\rho n = o(d\sigma_p^2)$. 
Further note that $|\ell_{1,i}^{(t)}|\le 1$ and $\eta =o(1)$, we have
\begin{align*}
&\sum_{\tau=T_1}^{t}\sum_{i\in\cS_1^+\cup\cS_1^-}|\ell_{1,i}^{(\tau)}|\le \tilde O\bigg(\frac{\rho n^2}{d\sigma_p^2 \eta}\bigg) + O(\rho n) = \tilde O\bigg(\frac{\rho n^2}{d\sigma_p^2 \eta}\bigg);\notag\\
&\sum_{\tau=T_1}^{t}\sum_{i\in\cS_0^+\cup\cS_0^-}|\ell_{1,i}^{(\tau)}| \le\tilde O\bigg(\frac{n}{\eta}\bigg)+O(n) =  \tilde O\bigg(\frac{n}{\eta}\bigg).
\end{align*}
Moreover, by Hypothesis \ref{hypothesis} for all $\tau\in[T_1, t]$, we also have for all $i\in\cS_0^+$,
\begin{align*}
|\ell_{1,i}^{(\tau)}| = \frac{\exp\big[F_2(\Wb^{(\tau)};\xb_i)-F_1(\Wb^{(\tau)};\xb_i)\big]}{1 + \exp\big[F_2(\Wb^{(\tau)};\xb_i)-F_1(\Wb^{(\tau)};\xb_i)\big]}.
\end{align*}
Moreover, we have
\begin{align*}
F_2(\Wb^{(\tau)};\xb_i)-F_1(\Wb^{(\tau)};\xb_i) &= \sum_{r=1}^m\sum_{p\in[P]}\la\wb_{2,r}^{(\tau)},\xb_i^{(p)}\ra - \sum_{r=1}^m\sum_{p\in[P]}\la\wb_{1,r}^{(\tau)},\xb_i^{(p)}\ra\notag\\
& = -\sum_{r=1}^{m}\sum_{p\in\cP_i(\vb)}(\la\wb_{k,r}^{(\tau)},\vb\ra)^2 \pm o\bigg(\frac{1}{\polylog(n)}\bigg)\le 0.
\end{align*}
This implies that for any $i,j\in\cS_0^+$ with $|\cP_i(\vb)|=|\cP_j(\vb)|$, we have
\begin{align*}
\frac{|\ell_{1,i}^{(\tau)}|}{|\ell_{1,j}^{(\tau)}|} = \Theta\bigg(\frac{\exp\big[F_2(\Wb^{(\tau)};\xb_i)-F_1(\Wb^{(\tau)};\xb_i)\big]}{\exp\big[F_2(\Wb^{(\tau)};\xb_j)-F_1(\Wb^{(\tau)};\xb_j)\big]}\bigg)= \Theta\big(\exp[o\big(1/\polylog(n)\big)]\big)=\Theta(1).
\end{align*}
Further note that, by Definition \ref{def:data_distribution_new}, the number of feature patches are uniformly sampled from $[1, \Theta(1)]$, implying that with probability at least $1-1/\poly(n)$, for any $i\in\cS_0^+$,
\begin{align*}
\#\big\{j:j\in\cS_0^+, |\cP_j(\vb)|=|\cP_i(\vb)|\big\} = \Theta(n)
\end{align*}
Therefore, let $\cS'$  be the above set of data points, we have for any $s\in\cS_0^+$ or $s\in\cS_1^+$,
\begin{align*}
\sum_{\tau=T_1}^{t}|\ell_{1,s}^{(\tau)}|&= \Theta\big(|\cS'|^{-1}\big)\sum_{\tau=T_1}^{t}\sum_{i\in\cS'}|\ell_{1,i}^{(\tau)}|\le \Theta\big(|\cS'|^{-1}\big)\sum_{\tau=T_1}^{t}\sum_{i\in\cS_0^+\cup\cS_0^-}|\ell_{1,i}^{(\tau)}|\le \tilde O\bigg(\frac{1}{ \eta}\bigg).
\end{align*}
where the last inequality is due to $|\cS'|=\Theta(n)$. This completes the proof.
\end{proof}

We will then verify Hypothesis \ref{hypothesis}\ref{item:weakfeat}, which is summarized in the following lemma.
\begin{lemma}\label{lemma:weakfeat_phase3}
Let Hypothesis \ref{hypothesis} holds for all $\tau\le t$, then we have
$|\la\wb_{k,r}^{(t+1)}, \vb'\ra|=O\big(|\la\wb_{k,r}^{(T_1)},\vb'\ra|\big)$ and $|\la\wb_{k,r}^{(t+1)}, \ub'\ra|=O\big(|\la\wb_{k,r}^{(T_1)},\ub'\ra|\big)$.
\end{lemma}
\begin{proof}[Proof of Lemma \ref{lemma:weakfeat_phase3}]
Recall the update of rare features in \eqref{eq:update_features}, we have
\begin{align*}
\la\wb_{k,r}^{(\tau+1)},\vb'\ra & = \la\wb_{k,r}^{(\tau)},\vb'\ra + \frac{2\eta}{n}\cdot\sum_{i\in\cS_1^{+}} \ell_{k,i}^{(\tau)} \sum_{p\in \cP_i(\vb')} \la\wb_{k,r}^{(\tau)},\vb'\ra\notag\\
\la\wb_{k,r}^{(\tau+1)},\ub'\ra & = \la\wb_{k,r}^{(\tau)},\ub'\ra + \frac{2\eta}{n}\cdot\sum_{i\in\cS_1^{-}} \ell_{k,i}^{(\tau)} \sum_{p\in \cP_i(\ub')} \la\wb_{k,r}^{(\tau)},\ub'\ra.
\end{align*}
Then according to the Hypothesis \ref{hypothesis}\ref{item:weakfeat} for all $\tau\in[T_1, t]$, we have
\begin{align*}
|\la\wb_{k,r}^{(t+1)},\vb'\ra| &\le |\la\wb_{k,r}^{(T_1)},\vb'\ra| + \frac{2\eta}{n}\cdot\sum_{\tau=T_1}^{t}\sum_{i\in\cS_1^{+}} |\ell_{k,i}^{(\tau)}| \sum_{p\in \cP_i(\vb')} |\la\wb_{k,r}^{(\tau)},\vb'\ra|\notag\\
& \le O\big(|\la\wb_{k,r}^{(T_1)},\vb'\ra|\big) + \frac{\eta}{n}\cdot \sum_{\tau=T_1}^{t}\sum_{i\in\cS_1^{+}} |\ell_{k,i}^{(\tau)}|\cdot O\big(|\la\wb_{k,r}^{(T_1)},\vb'\ra|\big), 
\end{align*}
where the last inequality is due to the fact that $|\cP_i(\vb')|=\Theta(1)$. By Lemma \ref{lemma:bound_sum_lossderivative}, it is clear that $\sum_{\tau=T_1}^{t}\sum_{i\in\cS_1^{+}} |\ell_{k,i}^{(\tau)}| = \tilde O\big(\frac{\rho n^2}{d\sigma_p^2 \eta}\big)$. Therefore,
\begin{align*}
|\la\wb_{k,r}^{(t+1)},\vb'\ra| \le O\big(|\la\wb_{k,r}^{(T_1)},\vb'\ra|\big) + \tilde O\bigg(\frac{\rho n}{d\sigma^2}\bigg)\cdot O\big(|\la\wb_{k,r}^{(T_1)},\vb'\ra|\big) = O\big(|\la\wb_{k,r}^{(T_1)},\vb'\ra|\big).
\end{align*}
The proof for $|\la\wb_{k,r}^{(t+1)},\ub'\ra|$ is similar so we omit it here. 
\end{proof}

Using the similar proof technique, we are able to verify Hypothesis \ref{hypothesis}\ref{item:strongfeat_incorrect}, \ref{hypothesis}\ref{item:strongfeat_noise}, and \ref{hypothesis}\ref{item:weakfeat_noise_incorrect}, which are summarized in the following lemmas.

\begin{lemma}\label{lemma:strongfeat_incorrect_phase3}
Let Hypothesis \ref{hypothesis} holds for all $\tau\le t$, then we have
$|\la\wb_{2,r}^{(t+1)}, \vb\ra|=O\big(|\la\wb_{2,r}^{(T_1)}, \vb\ra|\big)$ and $|\la\wb_{1,r}^{(t+1)}, \ub\ra|=O\big(|\la\wb_{1,r}^{(T_1)}, \ub\ra|\big)$.
\end{lemma}
\begin{proof}[Proof of Lemma \ref{lemma:strongfeat_incorrect_phase3}]
Since the proofs for $|\la\wb_{2,r}^{(t+1)}, \vb\ra|$ and $|\la\wb_{1,r}^{(t+1)}, \ub\ra|$ are basically identical, we will only provide the proof regarding $|\la\wb_{2,r}^{(t+1)}, \vb\ra|$. By \eqref{eq:update_features} and data distribution in Definition \ref{def:data_distribution_new}, we have
\begin{align*}
\la\wb_{2,r}^{(\tau+1)},\vb\ra &= \la\wb_{2,r}^{(\tau)}, \vb\ra\cdot\bigg[1 + \frac{2\eta}{n}\cdot\sum_{i\in\cS_0^+} \ell_{2,i}^{(\tau)} \sum_{p\in \cP_i(\vb)} \alpha_{i,p}^2\|\vb\|_2^2 + \frac{2\eta}{n}\cdot\sum_{i\in[n]\backslash\cS_0^+} \ell_{2,i}^{(\tau)} \sum_{p\in \cP_i(\vb)}  \alpha_{i,p}^2\|\vb\|_2^2\bigg]\notag\\
&\le \la\wb_{2,r}^{(\tau)}, \vb\ra + \frac{2\eta b\alpha^2}{n}\cdot \la\wb_{2,r}^{(\tau)}, \vb\ra\cdot \sum_{i=1}^n|\ell_{2,i}^{(\tau)}|.
\end{align*}
Taking an absolute value on both sides and then applying Hypothesis \ref{hypothesis}\ref{item:strongfeat_incorrect}, we have
\begin{align*}
|\la\wb_{2,r}^{(t+1)},\vb\ra|& \le |\la\wb_{2,r}^{(T_1)},\vb\ra| + \frac{2\eta b\alpha^2}{n}\cdot\sum_{\tau=T_1}^{t}\sum_{i=1}^n|\ell_{2,i}^{(\tau)}|\la\wb_{2,r}^{(\tau)},\vb\ra|\notag\\
& \le |\la\wb_{2,r}^{(T_1)},\vb\ra| + O\big(|\la\wb_{2,r}^{(T_1)},\vb\ra|\big)\cdot O\bigg(\frac{\eta b\alpha^2}{n}\bigg)\cdot \tilde O\bigg(\frac{\rho n^2}{d\sigma_p^2\eta}+\frac{n}{\eta}\bigg)\notag\\
& =O\big(|\la\wb_{2,r}^{(T_1)},\vb\ra|\big),
\end{align*}
where the second inequality is by Lemma \ref{lemma:bound_sum_lossderivative} and the last inequality is due to the fact that $\rho n=o( d\sigma_p^2)$ and $b\alpha^2=o(1)$.
This completes the proof.

\end{proof}

\begin{lemma}\label{lemma:strongfeat_noise}
Let Hypothesis \ref{hypothesis} holds for all $\tau\le t$, then we have
$|\la\wb_{k,r}^{(t+1)}, \bxi_s^{(q)}\ra|=o\big(1/\polylog(n)\big)$ for all $s\in\cS_0^+\cup\cS_0^-$, $r\in[m]$, $k\in[2]$, and $q\in[P]$.
\end{lemma}
\begin{proof}[Proof of Lemma \ref{lemma:strongfeat_noise}]

By \eqref{eq:update_noise_simplified}, we have
\begin{align*}
|\la \wb_{k,r}^{(t+1)},\bxi_s^{(q)}\ra| &\le |\la\wb_{k,r}^{(t)}, \bxi_s^{(q)}\ra| + |\la\wb_{k,r}^{(t)}, \bxi_s^{(q)}\ra|\cdot \tilde O\bigg(\frac{\eta d\sigma_p^2}{n}\bigg)\cdot |\ell_{k,s}^{(t)}| + \tilde O\bigg(\frac{d^{1/2}\sigma_p^2\eta}{n}\bigg)\cdot\sum_{p\in[P]}\sum_{i=1}^n |\ell_{k,i}^{(t)}|\cdot |\la\wb_{k,r}^{(t)},\bxi_i^{(p)}\ra|\notag\\
&\le |\la\wb_{k,r}^{(T_1)}, \bxi_s^{(q)}\ra| + \sum_{\tau=T_1}^t|\la\wb_{k,r}^{(\tau)}, \bxi_s^{(q)}\ra|\cdot \tilde O\bigg(\frac{\eta d\sigma_p^2}{n}\bigg)\cdot |\ell_{k,s}^{(t)}| \notag\\
&\qquad+ \tilde O\bigg(\frac{d^{1/2}\sigma_p^2\eta}{n}\bigg)\cdot\sum_{\tau=T_1}^t\sum_{p\in[P]}\sum_{i=1}^n |\ell_{k,i}^{(\tau)}|\cdot |\la\wb_{k,r}^{(\tau)},\bxi_i^{(p)}\ra|.
\end{align*}
Then by Hypotheses \ref{hypothesis}, we can further get
\begin{align}\label{eq:update_strongfeat_noise_phase3}
|\la \wb_{k,r}^{(t+1)},\bxi_s^{(q)}\ra| \le |\la \wb_{k,r}^{(T_1)},\bxi_s^{(q)}\ra| + o\bigg(\frac{1}{\polylog(n)}\bigg)\cdot \tilde O\bigg(\frac{\eta d\sigma_p^2}{n}\bigg)\cdot \sum_{\tau=T_1}^t|\ell_{k,s}^{(t)}| + \tilde O\bigg(\frac{Pd^{1/2}\sigma_p^2\eta}{n}\bigg)\cdot\sum_{\tau=T_1}^t\sum_{i=1}^n|\ell_{k,i}^{(\tau)}|.
\end{align}
Note that $s\in\cS_0^+\cup\cS_0^-$, 
then by Lemma \ref{lemma:bound_sum_lossderivative}, we have
\begin{align*}
\sum_{t=T_1}^t|\ell_{k,s}^{(\tau)}| = \tilde O\bigg(\frac{1}{\eta}\bigg),\quad \sum_{\tau=T_1}^t\sum_{i=1}^n |\ell_{k,i}^{(\tau)}|\le \tilde O\bigg(\frac{n}{\eta}\bigg).
\end{align*}
Therefore, plugging the above inequalities into \eqref{eq:update_strongfeat_noise_phase3} gives
\begin{align*}
|\la \wb_{k,r}^{(t+1)},\bxi_s^{(q)}&\ra|\le |\la\wb_{k,r}^{(T_1)}, \bxi_s^{(q)}\ra| + o\bigg(\frac{1}{\polylog(n)}\bigg)\cdot \tilde O\bigg(\frac{\eta d\sigma_p^2}{n}\bigg)\cdot \tilde O\bigg(\frac{1}{\eta}\bigg) + \tilde O\bigg(\frac{Pd^{1/2}\sigma_p^2\eta}{n}\bigg)\cdot\tilde O\bigg(\frac{n}{\eta}\bigg)\notag\\
& =o\bigg(\frac{1}{\polylog(n)}\bigg) + o\bigg(\frac{d\sigma_p^2}{n}\bigg) + \tilde O\big(d^{1/2}\sigma_p^2\big)\notag\\
& = o\bigg(\frac{1}{\polylog(n)}\bigg),
\end{align*}
where we use the fact that $d\sigma_p^2=o(n)$ and $d^{1/2}\sigma_p^2=o(1/\polylog(n))$. This completes the proof.
\end{proof}
\begin{lemma}\label{lemma:weakfeat_noise_incorrect}
Let Hypothesis \ref{hypothesis} holds for all $\tau\le t$, then we have
$|\la\wb_{k,r}^{(t+1)}, \bxi_s^{(q)}\ra|=O\big(|\la \wb_{2,r}^{(T_1)},\bxi_s^{(q)}\ra|\big)$ for all $s\in\cS_1^+\cup\cS_1^-$, $r\in[m]$, $k\neq y_s$, and $q\in[P]$.
\end{lemma}
\begin{proof}[Proof of Lemma \ref{lemma:weakfeat_noise_incorrect}]
Similar to the previous proof, we will only prove the argument for $s\in\cS_1^+$, the proof for $s\in\cS_1^-$ can be performed using exactly the same analysis. By \eqref{eq:update_noise_simplified}, we have for $s\in\cS_1^+$
\begin{align*}
|\la \wb_{2,r}^{(t+1)},\bxi_s^{(q)}\ra| &\le |\la\wb_{2,r}^{(t)}, \bxi_s^{(q)}\ra| -|\la\wb_{2,r}^{(t)}, \bxi_s^{(q)}\ra|\cdot \tilde O\bigg(\frac{\eta d\sigma_p^2}{n}\bigg)\cdot |\ell_{2,s}^{(t)}| + \tilde O\bigg(\frac{d^{1/2}\sigma_p^2\eta}{n}\bigg)\cdot\sum_{p\in[P]}\sum_{i=1}^n |\ell_{2,i}^{(t)}|\cdot|\la\wb_{2,r}^{(t)},\bxi_i^{(P)}\ra|\notag\\
&\le |\la\wb_{2,r}^{(T_1)}, \bxi_s^{(q)}\ra| + \sum_{\tau=T_1}^t \tilde O\bigg(\frac{Pd^{1/2}\sigma_p^2\eta}{n}\bigg)\cdot\sum_{i=1}^n |\ell_{k,i}^{(\tau)}|\notag\\
& = |\la \wb_{2,r}^{(T_1)},\bxi_s^{(q)}\ra| +  \tilde O\bigg(\frac{Pd^{1/2}\sigma_p^2\eta}{n}\bigg)\cdot\sum_{\tau=T_1}^t\sum_{i=1}^n |\ell_{2,i}^{(\tau)}|\notag\\
& = o\bigg(\frac{1}{\polylog(n)}\bigg),
\end{align*}
where the last inequality is by Lemma \ref{lemma:bound_sum_lossderivative}.
This completes the proof.

\end{proof}

Finally, we will verify the common features learning (Hypothesis \ref{hypothesis}\ref{item:strongfeat_correct}) and noise learning for rare feature data (Hypothesis \ref{hypothesis}\ref{item:weakfeat_noise_correct}).

\begin{lemma}\label{lemma:strongfeat_correct_phase3}
Let Hypothesis \ref{hypothesis} holds for all $\tau \le t$, then we have $\sum_{r=1}^m(\la\wb_{1,r}^{(t+1)},\vb\ra)^2=\tilde\Theta(1)$ and $\sum_{r=1}^m(\la\wb_{2,r}^{(t+1)},\ub\ra)^2=\tilde\Theta(1)$. 
\end{lemma}
\begin{proof}[Proof of Lemma \ref{lemma:strongfeat_correct_phase3}]
We first prove the upper bound: $\sum_{r=1}^m(\la\wb_{1,r}^{(t+1)},\vb\ra)^2\le\tilde\Theta(1)$. Particularly, by \eqref{eq:update_feature_v_phase2}, \eqref{eq:update_feature_u_phase2} and Definition \ref{def:data_distribution_new}, we have
\begin{align*}
\la\wb_{1,r}^{(t+1)}, \vb\ra &\le \la\wb_{1,r}^{(t)},\vb\ra \cdot\bigg[1 + \Theta\bigg(\frac{\eta}{n}\bigg)\cdot\sum_{i\in\cS_0^+}|\ell_{1,i}^{(t)}| + \Theta\bigg(\frac{ b\alpha^2\eta}{n} \big)\cdot\sum_{i=1}^n |\ell_{1,i}^{(t)}|  \bigg].
\end{align*}
Therefore, we can get that
\begin{align}\label{eq:update_feature_learning_phase3_proof}
\sum_{r=1}^m(\la\wb_{1,r}^{(t+1)},\vb\ra)^2 &\le \sum_{r=1}^m(\la\wb_{1,r}^{(t)},\vb\ra)^2  \cdot\bigg[1 + \Theta\bigg(\frac{\eta}{n}\bigg)\cdot\sum_{i\in\cS_0^+}|\ell_{1,i}^{(t)}| + \Theta\bigg(\frac{ b\alpha^2\eta}{n} \bigg)\cdot\sum_{i=1}^n |\ell_{1,i}^{(t)}|  \bigg],
\end{align}
where we use the fact that $|\ell_{1,i}^{(t)}|=|\ell_{2,i}^{(t)}|$. By Hypothesis \ref{hypothesis}, we have for all $\tau\le t$ and $i$,
\begin{align*}
\ell_{1,i}^{(\tau)} = \frac{\exp\big[F_2(\Wb^{(\tau)};\xb_i)-F_1(\Wb^{(\tau)};\xb_i)}{1+ \exp\big[F_2(\Wb^{(\tau)};\xb_i)-F_1(\Wb^{(\tau)};\xb_i)\big]} = \exp\bigg[-\Theta\bigg(\sum_{r=1}^m(\la\wb_{1,r}^{(\tau)},\vb\ra)^2\bigg)\bigg].
\end{align*}
Therefore, let $a_\tau: = \sum_{r=1}^m(\la\wb_{1,r}^{(\tau)},\vb\ra)^2$, we have the following according to \eqref{eq:update_feature_learning_phase3_proof}
\begin{align}\label{eq:update_at_correct_strongfeat}
a_{\tau+1}\le a_\tau \cdot\bigg[1 +  \Theta(\eta)\cdot e^{-c a_\tau}+\Theta\bigg(\frac{ b\alpha^2\eta}{n} \bigg)\cdot\sum_{i=1}^n |\ell_{1,i}^{(\tau)}|\bigg],
\end{align}
where $c$ is an absolute positive constant. Let $T=\polylog(n)$ be the total iteration number, then we will show that $a_t\le 3c^{-1}\log(T)$ for all $\tau\le t$. Particularly, we will prove that either (1) $a_\tau < 2c^{-1}\log(T)$ or (2) $a_{\tau}>2\log(T)>a_{\tau-1}$ but it will not reach $3\log(T)$ as $\tau$ increases before it becomes less than $2c^{-1}\log(T)$ again. The first case immediately implies that $a_\tau<3c^{-1}\log(T)$, so we will only need to focus on case (2). In this case, we have $a_\tau\le a_{\tau-1}+\Theta(\eta)\le 2.1c^{-1}\log(T)$. Then before $a_{\tau}$ becomes less than $2c^{-1}\log(T)$, we have for any $\tau'\in[\tau, t]$ that
\begin{align*}
a_{\tau'} \le a_\tau + \sum_{s=\tau}^{\tau'-1} a_s \cdot\bigg[ \Theta(\eta)\cdot e^{-c a_s}+\Theta\bigg(\frac{ b\alpha^2\eta}{n} \bigg)\cdot\sum_{i=1}^n |\ell_{1,i}^{(s)}|\bigg].
\end{align*}
Note that $a_s\cdot e^{-c a_s} \le 2c^{-1}\log(T)/T^2\le 0.1c^{-1}/T$ if $T=\omega(1)$, then using the fact that $\eta=o(1)$,
\begin{align*}
a_{\tau'} \le a_\tau + \sum_{s=\tau}^{\tau'-1} \bigg[\frac{\eta}{T}+\Theta\bigg(\frac{ b\alpha^2\eta}{n} \bigg)\cdot a_s\sum_{i=1}^n |\ell_{1,i}^{(s)}|\bigg]\le 2.2 c^{-1}\log(T) + \Theta\bigg(\frac{ b\alpha^2\eta}{n} \bigg)\cdot \sum_{s=\tau}^{\tau'-1}  a_s\sum_{i=1}^n |\ell_{1,i}^{(s)}|.
\end{align*}
Then as long as $a_s<10c^{-1}\log(T)$ for $s\in[\tau,\tau']$, we have the following according to Lemma \ref{lemma:bound_sum_lossderivative},
\begin{align*}
\Theta\bigg(\frac{ b\alpha^2\eta}{n} \bigg)\cdot \sum_{s=\tau}^{\tau'-1}  a_s\sum_{i=1}^n |\ell_{1,i}^{(s)}| = \tilde O\bigg(b\alpha^2+\frac{\rho n b\alpha^2}{d\sigma_p^2}\bigg) = o(1)\le 0.1c^{-1}\log(T),
\end{align*}
where we use the fact that $b\alpha^2=o(1)$ and $d\sigma_p^2=\omega(\rho n)$. Therefore, we can conclude that before $\alpha_\tau'$ reaches $10c^{-1}\log(T)$, it must satisfy
\begin{align*}
a_{\tau'}\le 2.3 c^{-1}\log(T),
\end{align*}
for any $\tau'\le t$. This further implies that
\begin{align*}
a_{t+1}\le a_t + \tilde O(\eta)\le 3c^{-1}\log(T) = O\big(\polylog(n)\big),
\end{align*}
which completes the proof of $\sum_{r=1}^m(\la\wb_{1,r}^{(t+1)},\vb\ra)^2=\tilde O(1)$. 

The next step is to show that $\sum_{r=1}^m(\la\wb_{1,r}^{(t+1)},\vb\ra)^2=\tilde \Omega(1)$. Similar to \eqref{eq:update_at_correct_strongfeat}, we can get that
\begin{align*}
a_{\tau+1}\ge a_\tau \cdot\bigg[ 1 + \Theta(\eta)\cdot e^{-C a_\tau}-\Theta\bigg(\frac{ b\alpha^2\eta}{n} \bigg)\cdot\sum_{i=1}^n |\ell_{1,i}^{(\tau)}|\bigg],
\end{align*}
where $C$ is an absolute positive constant. In fact, we must have $a_\tau\ge \frac{1}{\polylog(n)}$ since otherwise,
\begin{align*}
a_{\tau+1}\ge a_\tau \cdot\bigg[ 1 + \Theta(\eta)\cdot e^{-C a_\tau}-\Theta\bigg(\frac{ b\alpha^2\eta}{n} \bigg)\cdot\sum_{i=1}^n |\ell_{1,i}^{(\tau)}|\bigg]\ge a_\tau\cdot\big[ 1 +\Theta(\eta)\big],
\end{align*}
where the first inequality is due to $e^{-ca_\tau}=\Theta(1)$ if $a_\tau=O(1)$ and the second inequality is due to $|\ell_{1,i}^{(\tau)}|\le 1$ and $b\alpha^2=o(1)$. This implies that $a_{\tau+1}$ will keep increase, which will at least continue to the case that $a_\tau>1$. This completes the proof that $a_{t+1}=\tilde\Omega(1)$. 

The proof for $\sum_{r=1}^m(\la\wb_{2,r}^{(t+1)},\ub\ra)^2=\tilde \Theta(1)$ will be basically the same so we omit it here.

\end{proof}

\begin{lemma}\label{lemma:weakfeat_correctnoise_phase3}
Let Hypothesis \ref{hypothesis} holds for all $\tau \le t$, then we have $\sum_{r=1}^m\sum_{q\in\cP_s(\bxi)}(\la\wb_{1,r}^{(t+1)},\bxi_s^{(q)}\ra)^2=\tilde\Theta(1)$ for all $s\in\cS_1^+$, and $\sum_{r=1}^m\sum_{q\in\cP_s(\bxi)}(\la\wb_{2,r}^{(t+1)},\bxi_i^{(q)}\ra)^2=\tilde\Theta(1)$ for all $s\in\cS_1^-$. 
\end{lemma}
\begin{proof}[Proof of Lemma \ref{lemma:weakfeat_correctnoise_phase3}]
Note that $P,m=\Theta(\polylog(n))$, it suffices to prove that $\max_{q,r}(\la\wb_{1,r}^{(t+1)},\bxi_s^{(q)}\ra)^2=\tilde \Theta(1)$ for all $s\in\cS_1^+$
 and $\max_{q,r}(\la\wb_{2,r}^{(t+1)},\bxi_s^{(q)}\ra)^2=\tilde \Theta(1)$ all $s\in\cS_1^-$. In the following proof we will only consider $s\in\cS_1^+$ as the proof for $s\in\cS_1^-$ will exactly the same.

By \eqref{eq:update_noise_simplified}, we have for all $s\in\cS_1^+$,
\begin{align}\label{eq:noise_weakdata_phase3}
\la\wb_{1,r}^{(\tau+1)},\bxi_s^{(q)}\ra 
& = \la\wb_{1,r}^{(\tau)},\bxi_s^{(q)}\ra \cdot\bigg[ 1 + \frac{2\eta}{n}\cdot \ell_{1,s}^{(\tau)}\cdot\|\bxi_s^{(q)}\|_2^2\bigg] \pm \frac{2\eta}{n}\cdot\tilde O\big(d^{1/2}\sigma_p^2\big)\cdot\sum_{i\neq s || p\neq q} |\ell_{k,i}^{(\tau)}|\cdot |\la\wb_{1,r}^{(\tau)},\bxi_i^{(p)}\ra|.
\end{align}
We first prove the upper bound of $\sum_{r=1}^m\sum_{q\in\cP_s(\bxi)}(\la\wb_{2,r}^{(\tau)},\bxi_s^{(q)}\ra)^2$. Then, using the Hypothesis \ref{hypothesis} \ref{item:weakfeat_noise_correct}, we have for any $i\in[n]$,  $s\in\cS_1^+$, $r\in[m]$, and $p\in[P]$
\begin{align*}
(\la\wb_{1,r}^{(\tau)},\bxi_i^{(p)}\ra)^2\le O\bigg(\polylog(n)\bigg)\cdot O(mP)\cdot \max_{r,q}(\la\wb_{1,r}^{(\tau)},\bxi_s^{(q)}\ra)^2 =O\bigg(\polylog(n)\bigg)\cdot  \max_{r,q}(\la\wb_{1,r}^{(\tau)},\bxi_s^{(q)}\ra)^2.
\end{align*}
Then \eqref{eq:noise_weakdata_phase3} implies that
\begin{align*}
\max_{r,q}(\la\wb_{1,r}^{(\tau+1)},\bxi_s^{(q)}\ra)^2 
& \le  \max_{r,q}(\la\wb_{1,r}^{(\tau)},\bxi_s^{(q)}\ra)^2 \cdot\bigg[ 1 + \Theta\bigg(\frac{\eta d\sigma_p^2}{n}\bigg)\cdot \ell_{1,s}^{(\tau)} + \tilde O\bigg(\frac{\eta P d^{1/2}\sigma_p^2}{n}\bigg)\cdot\sum_{i=1}^n |\ell_{1,i}^{(\tau)}|\bigg]. 
\end{align*}

Then by Hypothesis \ref{hypothesis}, we can further get that the quantity $\sum_{r=1}^m\sum_{q\in\cP_s(\bxi)}(\la\wb_{1,r}^{(\tau+1)},\bxi_s^{(q)}\ra)^2$ will be the dominating term in the neural network output function, so that $\ell_{1,s}^{(\tau)}\ge e^{-c \max_{r,q}(\la\wb_{1,r}^{(\tau)},\bxi_s^{(q)}\ra)^2 }$ for some constant $c$. Therefore, let $a_\tau = \max_{r,q}(\la\wb_{1,r}^{(\tau)},\bxi_s^{(q)}\ra)^2$, we can follow the similar derivation of \eqref{eq:update_at_correct_strongfeat}. Thus, it follows that
\begin{align*}
a_{\tau+1}\le a_\tau\cdot\bigg[ 1 + \Theta\bigg(\frac{\eta d\sigma_p^2}{n}\bigg)\cdot e^{-c a_\tau} + \tilde O\bigg(\frac{\eta P d^{1/2}\sigma_p^2}{n}\bigg)\cdot\sum_{i=1}^n |\ell_{1,i}^{(\tau)}|\bigg]
\end{align*}
Then we can follow the exact proof technique in Lemma \ref{lemma:strongfeat_correct_phase3} to conclude that $a_{t+1}=\tilde O(1)$, while it only requires to verify that
\begin{align*}
\tilde O\bigg(\frac{\eta P d^{1/2}\sigma_p^2}{n}\bigg)\cdot\sum_{\tau=T_1}^t\sum_{i=1}^n |\ell_{1,i}^{(\tau)}|=o(1),
\end{align*}
which clearly holds by Lemma \ref{lemma:bound_sum_lossderivative} and the fact that $P d^{1/2}\sigma_p^2=o(1)$. 

The lower bound can be similarly obtained as the following can be deduced by \eqref{eq:noise_weakdata_phase3}:
\begin{align*}
\max_{r,q}(\la\wb_{1,r}^{(\tau+1)},\bxi_s^{(q)}\ra)^2 
& \ge  \max_{r,q}(\la\wb_{1,r}^{(\tau)},\bxi_s^{(q)}\ra)^2 \cdot\bigg[ 1 + \Theta\bigg(\frac{\eta d\sigma_p^2}{n}\bigg)\cdot \ell_{1,s}^{(\tau)} - \tilde O\bigg(\frac{\eta P d^{1/2}\sigma_p^2}{n}\bigg)\cdot\sum_{i=1}^n |\ell_{1,i}^{(\tau)}|\bigg],
\end{align*}
which leads to 
\begin{align*}
a_{\tau+1}\ge a_\tau\cdot\bigg[ 1 + \Theta\bigg(\frac{\eta d\sigma_p^2}{n}\bigg)\cdot e^{-C a_\tau} - \tilde O\bigg(\frac{\eta P d^{1/2}\sigma_p^2}{n}\bigg)\cdot\sum_{i=1}^n |\ell_{1,i}^{(\tau)}|\bigg]
\end{align*}
for some absolute constant $C$. Then following the same proof of Lemma \ref{lemma:strongfeat_correct_phase3}, we can get that $a_{t+1}=\tilde\Omega(1)$. This completes the proof.
\end{proof}

\subsection{Proof of Theorem \ref{thm:std_training}}\label{sec:proof_main_std}
\begin{proof}[Proof of Theorem \ref{thm:std_training}.]
We first show that $\|\wb_{k,r}^{(T)}\|_2 = \tilde O(n)$ for all $k\in[2]$ and $r\in[m]$. In particular, note that the update of standard training is always the linear combination of all critical vectors, i.e., $\vb$, $\ub$, $\vb'$, $\ub'$, and $\bxi_i^{(p)}$'s. Therefore, we have
\begin{align*}
\wb_{k,r}^{(t)} = \wb_{k,r}^{(0)} + \rho_{k,r}^{(t)}(\vb) \cdot \vb + \rho_{k,r}^{(t)}(\ub) \cdot \ub + \rho_{k,r}^{(t)}(\vb') \cdot \vb' + \rho_{k,r}^{(t)}(\ub') \cdot \ub' + \sum_{i=1}^n\sum_{p\in\cP_i(\bxi)}\rho_{k,r}^{(t)}(\bxi_i^{(p)}) \cdot \bxi_i^{(p)}.
\end{align*}
Here we use $\rho_{k,r}^{(t)}(\ab)$ to denote the coefficient of $\ab$ for all $\ab\in\{\vb,\ub,\vb',\ub'\}\cup\{\bxi\}$. Then by Lemma \ref{lemma:standard_learning_phase3_main} and using the fact that $\|\vb\|_2,\|\ub\|_2, \|\vb'\|_2, \|\ub'\|_2=1$, we have
\begin{align*}
|\rho_{k,r}^{(t)}(\vb)|, |\rho_{k,r}^{(t)}(\ub)| =\tilde O(1), \quad |\rho_{k,r}^{(t)}(\vb')|, |\rho_{k,r}^{(t)}(\ub')| =o\bigg(\frac{1}{\polylog(n)}\bigg).
\end{align*}
Moreover, using the fact that $|\la\bxi_i^{(p)},\bxi_j^{(q)}\ra|=o(1/\polylog(n))$ for any $i\neq j$ or $p\neq q$, applying Lemma \ref{lemma:standard_learning_phase3_main} and the fact that $\|\bxi_i^{(p)}\|_2^2=\Omega(1)$ for all $i\in[n]$ and $p\in[P]$, we have
\begin{align*}
\bigg\|\sum_{i=1}^n\sum_{p\in\cP_i(\bxi)}\rho_{k,r}^{(t)}(\bxi_i^{(p)}) \cdot \bxi_i^{(p)}\bigg\|_2^2\le \tilde O(n^2).
\end{align*}
Combining the above results, we can readily conclude that $\|\wb_{k,r}^{(t)}\|_2=\tilde O(n)$.

Then we will characterize the test errors for common
feature data and rare feature data separately. 
Regarding the common feature data, we can take a positive common feature data $(\xb, 1)$ as an example and obtain the following by Lemma \ref{lemma:standard_learning_phase3_main},
\begin{align}\label{eq:correct_output_std_training_strong}
F_1(\Wb^{(t)};\xb) = \sum_{r=1}^m \sum_{p=1}^P\big(\la\wb_{1,r}^{(t)},\xb^{(p)}\ra\big)^2 \ge \sum_{r=1}^m\sum_{p:\xb^{(p)}=\vb} \big(\la\wb_{1,r}^{(t)},\vb\ra\big)^2=\tilde\Theta(1).
\end{align}
Besides, we have the following regarding $F_2(\Wb^{(t)};\xb)$:
\begin{align}\label{eq:wrong_output_std_training_strong}
F_2(\Wb^{(t)};\xb) &= \sum_{r=1}^m\sum_{p:\xb^{(p)}=\vb} \big(\la\wb_{2,r}^{(t)},\vb\ra\big)^2 + \sum_{r=1}^m\sum_{p:\xb^{(p)}\neq \vb} \big(\la\wb_{2,r}^{(t)},\xb^{(p)}\ra\big)^2 \notag\\
&= \sum_{r=1}^m\sum_{p:\xb^{(p)}\neq \vb} \big(\la\wb_{2,r}^{(t)},\xb^{(p)}\ra\big)^2 + o\bigg(\frac{1}{\polylog(n)}\bigg).
\end{align}
where we use the result  $|\la\wb_{2,r}^{(t)},\vb\ra|=o\big(1/\polylog(n)\big)$. Then, note that if $\xb^{(p)}\neq \vb$, it can be either feature noise (i.e., $\alpha\ub$ or $\alpha\vb$) or random noise $\bzeta_i^{(p)}$, which is independent of the random noise vectors in the training data points (i.e., $\{\bxi\}$). Therefore, using the result that $\|\wb_{k,r}^{(t)}\|_2=\tilde O(n)$, we can obtain with probability at least $1- \exp(-\Omega(d^{1/2}))$, it holds that for all $r\in[m]$
\begin{align}\label{eq:noise_test_std_training}
(\la\wb_{2,r}^{(t)},\bzeta_i^{(p)}\ra)^2 = \tilde O(\sigma_p^2n^2).
\end{align}
Besides, note that there are at most $b$ patches within the total $P$ patches that are feature noise, we have
\begin{align*}
\sum_{r=1}^m\sum_{p:\xb^{(p)}\neq \vb} \big(\la\wb_{2,r}^{(t)},\xb^{(p)}\ra\big)^2\le O(mb\alpha^2) + \tilde O(mP\sigma_p^2 n^2) = o\bigg(\frac{1}{\polylog(n)}\bigg),
\end{align*}
where the last equality is by the data model in Definition \ref{def:data_distribution_new}: $b\alpha^2=o\big(1/\polylog(n)\big)$ and $\sigma_p = o(d^{-1/2}n^{1/2})$.
Therefore, comparing \eqref{eq:correct_output_std_training_strong} and \eqref{eq:wrong_output_std_training_strong}, we can get $F_1(\Wb^{(t)};\xb)>F_2(\Wb^{(t)};\xb)$ with probability at least $1-1/\poly(n)$. 

Then we will move on to study  the rare feature data. In particular, we consider the rare feature data with incorrect feature noise. Without loss of generality, we take a positive data $(\xb,1)$ as an example, which contains rare feature $\vb$ and incorrect feature noise $\alpha\ub$. Then we can get the following results for $F_k(\Wb^{(t)};\xb)$
\begin{align*}
F_k(\Wb^{(t)};\xb) &= \sum_{r=1}^m\sum_{p:\xb^{(p)}=\vb'} \big(\la\wb_{k,r}^{(t)},\vb'\ra\big)^2 + \sum_{r=1}^m\sum_{p:\xb^{(p)}=\alpha\ub} \big(\la\wb_{k,r}^{(t)},\alpha\ub\ra\big)^2 + \sum_{r=1}^m\sum_{p:\xb^{(p)}\not\in\{\vb',\alpha\ub\}} \big(\la\wb_{k,r}^{(t)},\xb^{(p)}\ra\big)^2.
\end{align*}
Note that if $\xb^{(p)}\not\in\{\vb',\alpha\ub\}$, then $\xb^{(p)}$ must be a random noise vector that is independent of $\wb_{k,r}^{(t)}$. To begin with, the first two terms of the above equation for different $k$'s can be bounded by applying Lemma \ref{lemma:standard_learning_phase3_main} (particularly $\sum_{r=1}^m(\la\wb_{2,r}^{(t)},\ub\ra)^2=\tilde\Omega(1)$), we have
\begin{align*}
&\sum_{r=1}^m\sum_{p:\xb^{(p)}=\vb'} \big(\la\wb_{1,r}^{(t)},\vb'\ra\big)^2 = \tilde O(\sigma_0^2), \quad \sum_{r=1}^m\sum_{p:\xb^{(p)}=\alpha\ub}\big(\la\wb_{1,r}^{(t)},\alpha\ub\ra\big)^2 = \tilde O(b\alpha^2\sigma_0^2),\notag\\
&\sum_{r=1}^m\sum_{p:\xb^{(p)}=\vb'} \big(\la\wb_{2,r}^{(t)},\vb'\ra\big)^2 = \tilde O(\sigma_0^2), \quad \sum_{r=1}^m\sum_{p:\xb^{(p)}=\alpha\ub}\big(\la\wb_{2,r}^{(t)},\alpha\ub\ra\big)^2 = \tilde\Omega(\alpha^2).
\end{align*}
Moreover, by \eqref{eq:noise_test_std_training}, we can further get that with probability at least $1-\exp(-\Omega(d^{1/2}))>1-1/\poly(n)$, we have
\begin{align*}
\sum_{r=1}^m\sum_{p:\xb^{(p)}\not\in\{\vb',\alpha\ub\}} \big(\la\wb_{k,r}^{(t)},\xb^{(p)}\ra\big)^2 = \tilde O(mP\sigma_p^2 n^2) = o(\alpha^2). 
\end{align*}
where the last equality is by our data model in Definition \ref{def:data_distribution_new}. This further implies that 
 conditioning on $\Wb^{(t)}$, with probability at least $1-1/\poly(n)$, we have
 \begin{align*}
 F_2(\Wb^{(t)};\xb)>  F_1(\Wb^{(t)};\xb)
 \end{align*}
on the positive rare feature data that has incorrect feature noise. 
\begin{align*}
\PP_{(\xb, y)\sim \cD_{\mathrm{rare}}}[ \argmax_k F_k(\Wb^{(t)};\xb)\neq y] \ge \frac{1}{2} - \frac{1}{\poly(n)}\ge \frac{1}{2.01}.
\end{align*}
Therefore, combining the test error analysis for common feature data and rare feature data and using the fact that the fraction of rare feature data is $\rho$, we can finally obtain:
\begin{align*}
\PP_{(\xb, y)\sim \cD}[ \argmax_k F_k(\Wb^{(t)};\xb)\neq y] \ge \rho \cdot \PP_{(\xb, y)\sim \cD_{\mathrm{rare}}}[ \argmax_k F_k(\Wb^{(t)};\xb)\neq y]  \ge \frac{\rho}{2.01}.
\end{align*}
This completes the proof.

\end{proof}

\section{Mixup data}

\subsection{Characterization of the mixup dataset}

\paragraph{Category of different Mixup data patches.}
First recall the category of different Mixup training data points:
\begin{itemize}
    \item Mix between two common feature data points, including $\cS_{0,0}^{+, +}$, $\cS_{0,0}^{-, -}$, $\cS_{0,0}^{+, -}$, $\cS_{0,0}^{-, +}$, each of them is of size $\Theta(n^2)$.
    \item Mix between  common feature and rare feature data points with the same label, including  $\cS_{0,1}^{+, +}$, $\cS_{0,1}^{-, -}$, $\cS_{1,0}^{+,+}$, and $\cS_{1,0}^{-,-}$, each of them is of size $\Theta(\rho n^2)$.
    \item Mix between common feature  and rare feature data points with different labels, including  $\cS_{0,1}^{+, -}$, $\cS_{0,1}^{-, +}$, $\cS_{1, 0}^{+, -}$, and $\cS_{1, 0}^{-, +}$, each of them is of size $\Theta(\rho n^2)$.
    \item Mix between two rare feature data points, including $\cS_{1,1}^{+, +}$,$\cS_{1,1}^{-, -}$, $\cS_{1,1}^{+, -}$ and $\cS_{1, 1}^{-, +}$, each of them is of size $\Theta(\rho^2 n^2)$.
    
\end{itemize}

Then, given $n^2$ mixed data points, we have in total $n^2P$ data patches. Besides, note that in the original dataset that consists of $n$ training data points, each data patch $\xb_i^{(p)}$ satisfies
\begin{align*}
\xb_i^{(p)}\in\big\{\vb,\ub, \alpha \ub, \alpha \vb, \vb', \ub', \bxi_i^{(p)}\big\}.
\end{align*}
Moreover, by the data distribution defined in Definition \ref{def:data_distribution_new}, we have
\begin{itemize}
\item $\vb$ and $\ub$ will appear in $\Theta(n)$ data and $\Theta(n)$ data patches.
\item $\alpha\vb$ and $\alpha\ub$ will appear in $n$ data and $\Theta(b n)$ data patches.
\item $\vb'$ and $\ub'$ will appear in $\Theta(\rho n)$ data and $\Theta(\rho n)$ data patches.
\item $\bxi_i^{(p)}$, if it is not zero, will appear in one data and one data patch.
\end{itemize}

Then based on the above facts, we provide the following lemma that characterizes the number of different types of data patches on the mixup dataset.

\begin{lemma}\label{lemma:occurance_mixed_patch}
Let $\cP:=\{\xb_{i,j}^{(p)}\}_{i,j\in[n], p\in[P]}$ be the collection of all data patches of the mixup dataset, then among these $n^2P$ data patches, with probability at least $1-1/\poly(n)$, let $\xb_{i,j}^{(p)}=\lambda\ab+(1-\lambda)\bb$, we have
\begin{itemize}
\item 
The vector with $\ab\in\{\vb,\ub\}$ and $\bb\in\{\vb,\ub\}$
will appear in $\Theta(n^2/P)$ data patches.
\item The vector with $\ab\in\{\vb,\ub\}$ and $\bb\in\{\vb',\ub'\}$   will appear in $\Theta(\rho n^2/P)$ patches.
\item The vector with $\ab\in\{\vb,\ub\}$ and $\bb\in\{\alpha\vb,\alpha\ub\}$ will appear in $O(bn^2/P)$ patches.
\item The vector with $\ab\in\{\vb,\ub\}$ and $\bb\in\{\bxi\}$ will appear in $\Theta(n^2)$ patches.
\item The vector with $\ab\in\{\vb',\alpha\ub'\}$ and $\bb\in\{\vb',\ub'\}$ will appear in $\Theta(\rho^2n^2/P)$ data patches.
\item The vector with $\ab\in\{\vb',\alpha\ub'\}$ and $\bb\in\{\alpha\vb,\alpha\ub\}$ will appear in $O(\rho bn^2/P)$ patches.
\item The vector with $\ab\in\{\vb',\alpha\ub'\}$ and $\bb\in\{\bxi\}$ will appear in $\Theta(\rho n^2)$ patches.
\item The vector with $\ab\in\{\alpha\vb,\alpha\ub\}$ and $\bb\in\{\alpha\vb,\alpha\ub\}$ will appear in $O(b^2n^2/P)$ patches.
\item The vector with $\ab\in\{\alpha\vb,\alpha\ub\}$ and $\bb\in\{\bxi\}$  will appear in $O(bn^2)$ patches.

\end{itemize}
Besides, regarding any non-zero noise vector $\bxi_i^{(p)}$, we have, among the collection of data patches $\{\xb_{i,j}^{(p)}\}_{j\in[n]}$, with probability at least $1-1/\poly(n)$, 
\begin{itemize}
\item $\xb_{i,j}^{(p)}=\lambda \bxi_{i,j}^{(p)}+(1-\lambda)\bb$ with $\bb\in\{\vb,\ub\}$ will appear in $\Theta(n/P)$ patches.
\item $\xb_{i,j}^{(p)}=\lambda \bxi_{i,j}^{(p)}+(1-\lambda)\bb$ with $\bb\in\{\alpha\vb,\alpha\ub\}$ will appear in $O(bn/P)$ patches.
\item $\xb_{i,j}^{(p)}=\lambda \bxi_{i,j}^{(p)}+(1-\lambda)\bb$ with $\bb\in\{\vb',\ub'\}$ will appear in $\Theta(\rho n/P)$ patches.
\item $\xb_{i,j}^{(p)}=\lambda \bxi_{i,j}^{(p)}+(1-\lambda)\bb$ with $\bb\in\{\bxi\}$ will appear in $\Theta(n)$ patches.
\end{itemize}

\end{lemma}
\begin{proof}[Proof of Lemma \ref{lemma:occurance_mixed_patch}]
We first consider a fixed $\xb_i$ and the corresponding collection of data patches $\{\xb_{i,j}^{(p)}\}_{j\in[n], p\in[P]}$.  Then by Definition \ref{def:data_distribution_new}, conditioning on $\xb_i^{(p)} = \vb$, we have for any $j\neq i$
\begin{align*}
\PP[\xb_j^{(p)}=\vb|\xb_i^{(p)} = \vb] =  \PP[\xb_j^{(p)}=\vb] = \Theta\bigg(\frac{1}{P}\bigg).
\end{align*}
Therefore, we can further get that conditioning on $\xb_{i}^{(p)}=\vb$, the summation $\sum_{j\neq i}\ind[\xb_{i,j}^{(p)}=\vb]$ follows Binomial distribution $\mathrm{Binom}(n-1, p)$ with probability parameter $p=\Theta(1/P)$. Then by Hoeffding's inequality, we can get that with probability at least $1-\exp(-n^2/P^2)$, it holds that 
\begin{align*}
\sum_{j\in[n]}\ind[\xb_j^{(p)}=\vb|\xb_i^{(p)}=\vb] = \Theta\bigg(\frac{n}{P}\bigg).
\end{align*}
Note that we have at least $\Theta(n)$ number of $\xb_i$'s that consist of the common feature vector $\vb$, then applying union bound over these $\xb_i$'s, we can further get with probability at least $1-1/\poly(n)$, it holds that
\begin{align*}
\sum_{i,j\in[n]}\sum_{p\in[P]}\ind[\xb_{i,j}^{(p)}=\vb]& \ge \sum_{i,j\in[n]}\ind[\xb_{i,j}^{(p_i)}=\vb|\xb_i^{(p_i)}=\vb]\cdot \ind[\xb_i^{(p_i)}=\vb]\notag\\
&\ge \Theta(n)\cdot\Theta\bigg(\frac{n}{P}\bigg)\notag\\
& = \Theta\bigg(\frac{n^2}{P}\bigg).
\end{align*}
Here we define $p_i$ as the index of the data patch that is $\vb$ if the data $\xb_i$ has such a common feature vector, otherwise, $p_i$ is arbitrarily chosen.
On the other hand, we can also get
\begin{align*}
\sum_{i,j\in[n]}\sum_{p\in[P]}\ind[\xb_{i,j}^{(p)}=\vb]\le \sum_{i,j\in[n]}\sum_{p\in[P]}\ind[\xb_{i,j}^{(p)}=\vb|\xb_i^{(p)}=\vb]\cdot \ind[\xb_i^{(p)}=\vb]\le n\cdot \Theta(1)\cdot\Theta\bigg(\frac{n}{P}\bigg)=\Theta\bigg(\frac{n^2}{P}\bigg),
\end{align*}
where the second inequality is due to that each data will have at most $\Theta(1)$ patches being $\vb$. Similarly, we can also prove the same results for the case of $\xb_{i,j}^{(p)} = \lambda\ab + (1-\lambda)\bb$ with $\ab,\bb\in\{\ub,\vb\}$.

The proof for the case of $\xb_{i,j}^{(p)} = \lambda\ab + (1-\lambda)\bb$ with $\ab\in\{\ub,\vb\}$ and $\bb\in\{\alpha\vb,\alpha\ub\}$ will be also similar, the only difference is that conditioning on $\xb_i^{(p)}=\vb$, the probability of $\xb_j^{(p)}=\alpha\vb$
or $\xb_j^{(p)}=\alpha\ub$ will be $O(b/P)$. Finally, we can get that (here we take $\ab=\vb$ and $\bb=\vb$ as an example)
\begin{align*}
\sum_{i,j\in[n]}\sum_{p\in[P]}\ind[\xb_{i,j}^{(p)}=\lambda\vb + \alpha(1-\lambda)\vb] =\Theta(n)\cdot\Theta\bigg(\frac{bn}{P}\bigg)=\Theta\bigg(\frac{bn^2}{P}\bigg).
\end{align*}

The proof for the case of $\xb_{i,j}^{(p)} = \lambda\ab + (1-\lambda)\bb$  with $\ab\in\{\ub,\vb\}$ and $\bb\in\{\ub',\vb'\}$ will also be similar, where we only need to use the fact that $\PP[\xb_j^{(p)}=\vb'|\xb_i^{(p)} = \vb]=\Theta(\rho/P)$. Here we take $\ab=\vb$ and $\bb=\vb'$ as an example.

Regarding the case of $\xb_{i,j}^{(p)} = \lambda\ab + (1-\lambda)\bb$ with $\ab\in\{\ub,\vb\}$ and $\bb\in\{\bxi\}$, we only need to use the fact that $\PP[\xb_{i,j}^{(p)}=\lambda\vb+(1-\lambda)\bxi_j^{(p)}|\xb_i^{(p)} = \vb]=\Theta(1)$, where we take $\ab=\vb$ as an example. Then the desired result can be proved in a similar way.

 When $\ab\in\{\vb', \ub'\}$ we will also need to use the fact that we have in total $\Theta(\rho n)$ number of $\xb_i$'s that consist of $\vb'$ or $\ub'$. Then take $\ab=\vb'$ and $\bb=\vb'$ as an example, conditioning on $\xb_i^{(p)}=\vb'$, we have for any $j\neq i$
\begin{align*}
\PP[\xb_j^{(p)}=\vb'|\xb_i^{(p)} = \vb'] =  \PP[\xb_j^{(p)}=\vb'] = \Theta\bigg(\frac{\rho}{P}\bigg).
\end{align*} 
Therefore, we can get that with probability at least $1-1/\poly(n)$,
\begin{align*}
\sum_{j\in[n]}\ind[\xb_j^{(p)}=\vb'|\xb_i^{(p)}=\vb'] = \Theta\bigg(\frac{\rho n}{P}\bigg).
\end{align*}
Accordingly, we can further obtain
\begin{align*}
\sum_{i,j\in[n]}\sum_{p\in[P]}\ind[\xb_{i,j}^{(p)}=\vb']= \sum_{i,j\in[n]}\sum_{p\in[P]}\ind[\xb_{i,j}^{(p)}=\vb'|\xb_i^{(p)}=\vb']\cdot \ind[\xb_i^{(p)}=\vb']=\Theta(\rho n)\cdot \Theta\bigg(\frac{\rho n}{P}\bigg) = \Theta\bigg(\frac{\rho^2 n^2}{P}\bigg).
\end{align*}

The proof for the case of $\xb_{i,j}^{(p)} = \lambda\ab + (1-\lambda)\bb$  with $\ab\in\{\ub',\vb'\}$ and $\bb\in\{\alpha\vb,\alpha\vb\}$ or  $\bb\in\{\bxi\}$ will also be similar, where we only need to use the fact that $\PP[\xb_j^{(p)}=\alpha\vb|\xb_i^{(p)} = \vb']=O(b/P)$ and $\PP[\xb_j^{(p)}=\alpha\bxi_j^{(p)}|\xb_i^{(p)} = \vb']=\Theta(1)$.

When $\ab\in\{\alpha\vb, \alpha\ub\}$ we only need to use the fact that we have in total $\Theta(n)$ number of $\xb_i$'s that consist of $\Theta(b)$ number of $\vb'$ or $\ub'$. The remaining proof will be similar to previous ones based on the fact that $\PP[\xb_j^{(p)}=\alpha\vb|\xb_i^{(p)} = \alpha\vb]=O(b/P)$ and $\PP[\xb_j^{(p)}=\alpha\bxi_j^{(p)}|\xb_i^{(p)} = \alpha\vb]=\Theta(1)$, where we take $\ab=\alpha\vb$ and $\bb=\alpha\vb$ as an example.

Lastly, we will move on to the case of $\ab = \bxi_i^{(p)}$. In this case, we only need to use the facts that for any $j\neq i$,
\begin{align*}
\PP[\xb_j^{(p)}=\vb|\xb_i^{(p)}= \bxi_i^{(p)}]&=\PP[\xb_j^{(p)}=\vb] = \Theta(1/P)\notag\\
\PP[\xb_j^{(p)}=\ub|\xb_i^{(p)} = \bxi_i^{(p)}]&=\PP[\xb_j^{(p)}=\ub] = \Theta(1/P)\notag\\
\PP[\xb_j^{(p)}=\vb'|\xb_i^{(p)} = \bxi_i^{(p)}]&=\PP[\xb_j^{(p)}=\vb'] = \Theta(\rho/P) \notag\\
\PP[\xb_j^{(p)}=\ub'|\xb_i^{(p)} = \bxi_i^{(p)}]&=\PP[\xb_j^{(p)}=\ub'] = \Theta(\rho/P)\notag\\
\PP[\xb_j^{(p)}=\alpha\vb|\xb_i^{(p)} = \bxi_i^{(p)}]&=\PP[\xb_j^{(p)}=\alpha\vb] = O(b/P)\notag\\
\PP[\xb_j^{(p)}=\alpha\ub|\xb_i^{(p)} = \bxi_i^{(p)}]&=\PP[\xb_j^{(p)}=\alpha\ub] = O(b/P)\notag\\
\PP[\xb_j^{(p)}\in\{\bxi\}|\xb_i^{(p)} = \bxi_i^{(p)}]&=\PP[\xb_j^{(p)}\in\{\bxi\}] = \Theta(1).
\end{align*}
Then applying the standard concentration argument for binomial distribution yields the desired results.

\end{proof}

\subsection{Learning Dynamics of Feature and Noise vectors}

Now, we will seek to study the learning of feature and noise vectors. Particularly, the update formulas of all feature vectors are provided as follows: for any $\ab\in\{\ub,\vb,\ub',\vb'\}\cup\{\bxi\}$, we have
\begin{align}\label{eq:update_allfeatures_mixup}
\la\wb_{k,r}^{(t+1)}, \ab\ra &= \la\wb_{k,r}^{(t)}, \ab\ra - \eta\cdot\la\nabla_{\wb_{k,r}} L(\Wb^{(t)}),\ab\ra\notag\\
&=\la\wb_{k,r}^{(t)}, \ab\ra + \frac{\eta}{n^2}\cdot   \sum_{i,j\in[n]} \ell_{k,(i,j)}^{(t)} \sum_{p\in[P]}\la\wb_{k,r}^{(t)},\xb_{i,j}^{(p)}\ra\cdot\la \xb_{i,j}^{(p)},\ab\ra
\end{align}

More specifically, we summarize the update of all critical vectors (e.g., common features, rare features, and data noise vectors) in the following Proposition.
\begin{proposition}\label{prop:update_mixup}
For any critical vector $\ab\in\{\vb, \ub, \vb', \ub'\}\cup\{\bxi\}$, we have
\begin{align*}
-\la\nabla_{\wb_{k,r}} L_{\cS}(\Wb^{(t)}),\ab\ra &=  \gamma_k^{(t)}(\vb,\ab)\cdot\la\wb_{k,r}^{(t)}, \vb\ra + \gamma_k^{(t)}(\ub,\ab)\cdot\la\wb_{k,r}^{(t)},\ub\ra +\gamma_k^{(t)}(\vb',\ab)\cdot\la\wb_{k,r}^{(t)},\vb'\ra \notag\\
&\qquad + \gamma_k^{(t)}(\ub',\ab)\cdot\la\wb_{k,r}^{(t)},\ub'\ra +  \sum_{i=1}^n\sum_{p\in[P]}\gamma_k^{(t)}(\bxi_i^{(p)},\ab)\cdot\la\wb_{k,r}^{(t)},\bxi_i^{(p)}\ra,
\end{align*}
where $\gamma_k^{(t)}(\bb,\ab)$ is a scalar output function that depends on $\bb, \ab\in\{\vb, \ub, \vb', \ub'\}\cup\{\bxi\}$. More specifically, let 
\begin{align}\label{eq:linear_expansion_data}
\xb_{i,j}^{(p)} = \theta_{i,j}^{(p)}(\vb) \cdot\vb + \theta_{i,j}^{(p)}(\ub)\cdot \ub +\theta_{i,j}^{(p)}(\vb') \cdot\vb' + \theta_{i,j}^{(p)}(\ub')\cdot \ub' + \sum_{s=1}^n\sum_{q\in[P]}\theta_{i,j}^{(p)}(\bxi_s^{(q)})\cdot \bxi_s^{(q)} 
\end{align}
be a linear expansion of $\xb_{i,j}^{(p)}$ on the space spanned by $\{\vb, \ub, \vb', \ub'\}\cup\{\bxi\}$, we have
\begin{align}\label{eq:expansion_mixed_data}
\gamma_k^{(t)}(\bb, \ab) = \frac{1}{n^2}\sum_{i,j\in[n]}\ell_{k,(i,j)}^{(t)} \sum_{p\in[P]} \theta_{i,j}^{(p)}(\bb)\cdot\la\xb_{i,j}^{(p)},\ab\ra.
\end{align}

\end{proposition}
\begin{proof}[Proof of Proposition \ref{prop:update_mixup}]
Recall \eqref{eq:update_allfeatures_mixup} and the decomposition of $\xb_{i,j}^{(p)}$ in \eqref{eq:linear_expansion_data}, we have
\begin{align*}
-\la\nabla_{\wb_{k,r}}L_\cS(\Wb^{(t)}),\ab\ra &= \frac{1}{n^2}\sum_{i,j\in[n]} \ell_{k,(i,j)}^{(t)} \sum_{p\in[P]}\la\wb_{k,r}^{(t)},\xb_{i,j}^{(p)}\ra\cdot\la \xb_{i,j}^{(p)},\ab\ra\notag\\
& = \frac{1}{n^2}\sum_{i,j\in[n]}\ell_{k,(i,j)}^{(t)} \sum_{p\in[P]} \sum_{\bb\in\{\vb,\ub,\vb',\ub'\}\cup\{\bxi\}}\theta_{i,j}^{(p)}(\bb)\cdot \la\wb_{k,r}^{(t)},\bb\ra\cdot \la\xb_{i,j}^{(p)},\ab\ra\notag\\
& =  \sum_{\bb\in\{\vb,\ub,\vb',\ub'\}\cup\{\bxi\}}\bigg[\frac{1}{n^2}\sum_{i,j\in[n]}\ell_{k,(i,j)}^{(t)} \sum_{p\in[P]} \theta_{i,j}^{(p)}(\bb)\cdot \la\xb_{i,j}^{(p)},\ab\ra\bigg]\cdot \la\wb_{k,r}^{(t)},\bb\ra. 
\end{align*}
Therefore, it is easy to see that using the definition of $\gamma_k^{(t)}(\bb,\ab)$ in \eqref{eq:expansion_mixed_data}, we have
\begin{align*}
-\la\nabla_{\wb_{k,r}}L_\cS(\Wb^{(t)}),\ab\ra = \sum_{\bb\in\{\vb,\ub,\vb',\ub'\}\cup\{\bxi\}} \gamma_k^{(t)}(\bb,\ab)\cdot \la\wb_{k,r}^{(t)},\bb\ra,
\end{align*}
which completes the proof.

\end{proof}

Note that the neural network outputs are in the order of $o(1)$ in the first few iterations, which implies that the output logits are within the range $[0.5-o(1), 0.5+o(1)]$. Further note that the loss derivatives $\ell_{k;(i,j)}^{(t)}$ satisfies
\begin{align*}
|\ell_{k;(i,j)}^{(t)}|\in\big\{1 - \logit_k(\Wb^{(t)};\xb_{i,j}), \logit_k(\wb^{(t)};\xb_{i,j}), \lambda - \logit_k(\Wb^{(t)};\xb_{i,j}), \logit_k(\Wb^{(t)};\xb_{i,j})+\lambda - 1\big\},
\end{align*}
which will also be in the constant order. Then similar to the previous analysis on the standard training, we will directly take $|\ell_{k,(i,j)}^{(t)}|=\Theta(1)$ when characterizing the learning of feature and noise vectors in the initial phase.

Then, the challenging part in the analysis is the characterization of the mixed data patches $\{\xb_{i,j}^{(p)}\}_{p\in[P]}$, since it can be: mixture of common features, mixture of rare features, mixture of common and rare features, mixture of feature and noise, which will produce different gradients. For any mixed data $\xb_{i,j}=\lambda \xb_i+(1-\lambda)\xb_j$, we will denote it as the positive mixed data if $y_i=1$ and the negative mixed data if $y_i=-1$. The following lemma gives the characterization of the data patch of all mixed data.

\subsection{Characterizing the Coefficient $\gamma_k^{(t)}(\cdot,\cdot)$}

\subsubsection{Correct Common Feature Learning}

\begin{lemma}\label{lemma:feature_learning_coefficients_v_mixup}
Assume $\max_{k\in[2], (i,j)\in\cS} |F_k(\Wb^{(t)}; \xb_{i,j})|\le \zeta\in\big[\omega(b\alpha),o\big(\frac{1}{\polylog(n)}\big)\big]$, then recalling the update form in Proposition \ref{prop:update_mixup}, we have
\begin{align*}
&\gamma_1^{(t)}(\vb, \vb) = \Theta(1), \quad |\gamma_1^{(t)}(\ub, \vb)| =  O(\zeta+\alpha), \quad |\gamma_1^{(t)}(\vb',\vb)| = O(\rho /P), \notag\\
&|\gamma_1(\ub',\vb)| = O(\zeta\rho/P), \quad |\gamma_1^{(t)}(\bxi_s^{(q)}, \vb)| = \tilde O\big(1/(Pn)\big).
\end{align*}

\end{lemma}
\begin{proof}[Proof of Lemma \ref{lemma:feature_learning_coefficients_v_mixup}]
We will prove all the arguments in order. 
\paragraph{Proof for $\gamma_1^{(t)}(\vb,\vb)$.}
We first prove the bound for $\gamma_1^{(t)}(\vb, \vb)$. By \eqref{eq:expansion_mixed_data}, we have
\begin{align}\label{eq:coefficient_update_v_v_mixup}
\gamma_k^{(t)}(\vb, \vb) = \frac{1}{n^2}\sum_{i,j\in[n]}\ell_{k,(i,j)}^{(t)} \sum_{p\in[P]} \theta_{i,j}^{(p)}(\vb)\cdot\la\xb_{i,j}^{(p)},\vb\ra,
\end{align}
where $\theta_{i,j}^{(p)}(\vb) = \la\xb_{i,j}^{(p)},\vb\ra$. Therefore, we only need to consider the data patches that contain $\vb$ (including common feature $\vb$ and feature noise $\alpha\vb$). The regarding the mixed data $\xb_{i,j}$, we consider the following cases
\begin{itemize}
    \item $i\in\cS_0^+$ and $j\in\cS_0^+$;
    \item $i\in\cS_0^+$ and $j\in\cS_1^+$, and $i\in\cS_1^+$ and $j\in\cS_0^+$;
    \item $i\in\cS_0^+$ and $j\in\cS_0^-\cup\cS_1^-$, and $i\in\cS_0^-\cup\cS_1^-$ and $j\in\cS_0^+$
    \item $i\in\cS_0^-\cup\cS_1^+\cup\cS_1^-$ and $j\in\cS_0^-\cup\cS_1^+\cup\cS_1^-$.
\end{itemize}

\textit{Analysis on the data  $i\in\cS_0^+$ and $j\in\cS_0^+$} In particular, note that before the mixup, both the data $\xb_i$ and $\xb_j$ have a constant number of common feature patches. Therefore, let $\cP_{i,j}^*(\vb)$ denote the set of patches with the common feature $\vb$ (which appears in either $\xb_i$ or $\xb_j$), we have
\begin{align}\label{eq:decomposition_case1_v_mixup}
\sum_{p\in[P]}\theta_{i,j}^{(p)}(\vb)\cdot\la\xb_{i,j}^{(p)},\vb\ra  = \sum_{p\in \cP_{i,j}^*(\vb)}\theta_{i,j}^{(p)}(\vb)\cdot\la\xb_{i,j}^{(p)},\vb\ra  + \sum_{p\in \cP_{i,j}(\vb)\backslash\cP_{i,j}^*(\vb)}\theta_{i,j}^{(p)}(\vb)\cdot\la\xb_{i,j}^{(p)},\vb\ra.
\end{align}
Regarding the first term on the R.H.S. of the above equation, by Definition \ref{def:data_distribution_new}, we know that there exists at least one common feature patch in both $\xb_i$ and $\xb_j$, which leads to $\theta_{i,j}^{(p)}\ge \lambda$ for at least one $p\in\cP_{i,j}^*(vb)$. This further gives 
\begin{align*}
 \sum_{p\in \cP_{i,j}^*(\vb)}\theta_{i,j}^{(p)}(\vb)\cdot\la\xb_{i,j}^{(p)},\vb\ra=   \sum_{p\in \cP_{i,j}^*(\vb)}[\theta_{i,j}^{(p)}(\vb)]^2\ge \lambda^2.
\end{align*}
Besides, we also have that the number of common feature patches are upper bounded by some constant (i.e., $|\cP_{i,j}^*(\vb)|=\Theta(1)$), this further leads to
\begin{align*}
\sum_{p\in \cP_{i,j}^*(\vb)}\theta_{i,j}^{(p)}(\vb)\cdot\la\xb_{i,j}^{(p)},\vb\ra \le \Theta(1).
\end{align*}

Regarding the second term on the R.H.S. of \eqref{eq:decomposition_case1_v_mixup}, we have $\theta_{i,j}^{(p)}\le \alpha$ since $\vb$ can only appear in the form of feature noise. Besides, by Definition \ref{def:data_distribution_new}, we know that the number of patches containing feature noise is at most $b$, then
\begin{align*}
\sum_{p\in \cP_{i,j}(\vb)\backslash\cP_{i,j}^*(\vb)}\theta_{i,j}^{(p)}(\vb)\cdot\la\xb_{i,j}^{(p)},\vb\ra =  \sum_{p\in \cP_{i,j}(\vb)\backslash\cP_{i,j}^*(\vb)}[\theta_{i,j}^{(p)}(\vb)]^2 \le b\alpha^2=o\bigg(\frac{1}{\polylog(n)}\bigg).
\end{align*}
Moreover, note that in the initial phase we have $\ell_{1,(i,j)}^{(t)}=\Theta(1)$ for $(i,j)\in \cS_{0,0}^{+, +}$, we can further get that
\begin{align*}
\ell_{1,(i,j)}^{(t)} \sum_{p\in[P]} \theta_{i,j}^{(p)}(\vb)\cdot\la\xb_{i,j}^{(p)},\vb\ra =\Theta(1).
\end{align*}

\textit{Analysis on  the data $i\in\cS_0^+$ and $j\in\cS_1^+$.}
The analysis for this type of data will be similar. In fact, we will  consider two types of data: $i\in\cS_0^+$ and $j\in\cS_1^+$, and $i\in\cS_1^+$ and $j\in\cS_0^+$ since two original training data will give two mixed data. 

In particular, note that $\ell_{i,j}^{(t)}=\Theta(1)$ for these two types of data, we can immediately get that there is a constant number of patches that satisfy $\theta_{i,j}^{(p)}\ge 1-\lambda$, while the remaining patches $p\in\cP_{i,j}(\vb)$ satisfy $\theta_{i,j}^{(p)}\le\alpha$. Therefore, we can follow the same proof technique as that for the data $(i,j)\in\cS_{0,0}^{+, +}$ and get that for all $(i,j)\in\cS_{0,1}^{+, +}\cup\cS_{1, 0}^{+, +}$,
\begin{align}\label{eq:}
\ell_{k,(i,j)}^{(t)} \sum_{p\in[P]} \theta_{i,j}^{(p)}(\vb)\cdot\la\xb_{i,j}^{(p)},\vb\ra =\ell_{k,(i,j)}^{(t)} \sum_{p\in\cP_{i,j}^*(\vb)} [\theta_{i,j}^{(p)}(\vb)]^2 + \sum_{p\in\cP_{i,j}(\vb)\backslash\cP_{i,j}^*(\vb)} [\theta_{i,j}^{(p)}(\vb)]^2 =\Theta(1).
\end{align}

\textit{ Analysis on  the data $i\in\cS_0^+$ and $j\in\cS_0^-\cup\cS_1^-$.}
In this part, we will handle data $\xb_{i,j}$ and $\xb_{j,i}$ together. Different from the previous cases where the loss derivatives $\ell_{1, (i,j)}^{(t)}$ are positive, here the loss derivative $\ell_{1, (i,j)}^{(t)}$ will become negative for $(i,j)\in\cS_{0, 0}^{-, +}\cup\cS_{0,1}^{-, +}$. Particularly, for any $(i,j)\in\cS_{0, 0}^{+, -}$, we have $(j,i)\in\cS_{0,0}^{-, +}$, then
\begin{align}\label{eq:coefficient_differentlabel_data_v_mixup}
&\ell_{1,(i,j)}^{(t)} \sum_{p\in[P]} \theta_{i,j}^{(p)}(\vb)\cdot\la\xb_{i,j}^{(p)},\vb\ra + \ell_{1,(j,i)}^{(t)} \sum_{p\in[P]} \theta_{j,i}^{(p)}(\vb)\cdot\la\xb_{j,i}^{(p)},\vb\ra\notag\\
&=\ell_{1,(i,j)}^{(t)}\sum_{p\in\cP_{i,j}(\vb)}\big[[\theta_{i,j}^{(p)}(\vb)]^2-[\theta_{j,i}^{(p)}(\vb)]^2\big] + \big[\ell_{1,(i,j)}^{(t)}+\ell_{1,(j,i)}^{(t)}\big]\cdot \sum_{p\in\cP_{i,j}(\vb)}\big[\theta_{j,i}^{(p)}(\vb)\big]^2,
\end{align}
where we use the fact that $\cP_{i,j}(\vb) = \cP_{j,i}(\vb)$ and $\la\xb_{i,j}^{(p)},\vb\ra = \theta_{i,j}^{(p)}(\vb)$.
Recall that the neural network output is upper bounded by $\zeta$, then it is easy to see
\begin{align*}
|\ell_{1,(i,j)}^{(t)} + \ell_{1,(j,i)}^{(t)}| = |\lambda - 0.5 \pm O(\zeta) + 0.5-\lambda \pm O(\zeta)| = O(\zeta). 
\end{align*}
Besides, note that
\begin{align*}
\xb_{i,j}^{(p)} = \lambda \xb_i^{(p)} + (1-\lambda)\xb_j^{(p)},\quad \xb_{j,i}^{(p)} + (1-\lambda)\xb_i^{(p)} + \lambda\xb_j^{(p)}.
\end{align*}
Then we will also define $\cP_{i,j}^*(\vb)$ as the set of patches with common feature. Note that $\xb_j^{(p)}$ does not have the common feature patch since $j\in\cS_0^-\cup\cS_1^-$, we can immediately get that $\cP_{i,j}^*(\vb) = \cP_i^*(\vb)$, where $\cP_i^*(\vb)$ denotes the set of common feature patches of $\xb_i$. Besides, it is also clear that all data patches in $\cP_{i,j}(\vb)$ only contain the feature noise $\alpha\vb$. Then it follows that
\begin{align*}
&\ell_{1,(i,j)}^{(t)}\sum_{p\in\cP_{i,j}(\vb)}\big[[\theta_{i,j}^{(p)}(\vb)]^2-[\theta_{j,i}^{(p)}(\vb)]^2\big] \notag\\
&= \Theta(1)\cdot\bigg[ \sum_{p\in\cP_{i,j}^*(\vb)}[\lambda^2-(1-\lambda)^2] +  \sum_{p\in\cP_{i,j}(\vb)\backslash\cP_{i,j}^*(\vb)}\big[[\theta_{i,j}^{(p)}(\vb)]^2-[\theta_{j,i}^{(p)}(\vb)]^2\big]\bigg]\notag\\
&=\Theta(1)  \pm O(b\alpha^2)\notag\\
&=\Theta(1).
\end{align*}
Similarly, we can also get $\sum_{p\in\cP_{i,j}(\vb)}[\theta_{j,i}^{(p)}]^2=\Theta(1)$. Therefore, putting everything to \eqref{eq:coefficient_differentlabel_data_v_mixup}, we can finally obtain the following 
\begin{align*}
\ell_{1,(i,j)}^{(t)} \sum_{p\in[P]} \theta_{i,j}^{(p)}(\vb)\cdot\la\xb_{i,j}^{(p)},\vb\ra + \ell_{1,(j,i)}^{(t)} \sum_{p\in[P]} \theta_{j,i}^{(p)}(\vb)\cdot\la\xb_{j,i}^{(p)},\vb\ra = \Theta(1) \pm \Theta(1)\cdot O(\zeta) = \Theta(1).
\end{align*}

\textit{Analysis on the data $i,j\in\cS_0^-\cup\cS_1^+\cup\cS_1^-$} In this case, we can observe that there is no common feature patches in $\xb_i$ and $\xb_j$, while the vector $\vb$ will only appear in at most $2b$ patches of $\xb_{i,j}$ in the form of feature noise. Therefore, we have $|\Theta_{i,j}^{(p)}|\in [(1-\lambda)\alpha, \alpha]$ for at most $2b$ patches and the remaining patches will give $|\Theta_{i,j}^{(p)}|=0$. Consequently, we have
\begin{align*}
\bigg|\ell_{k,(i,j)}^{(t)} \sum_{p\in[P]} \theta_{i,j}^{(p)}(\vb)\cdot\la\xb_{i,j}^{(p)},\vb\ra\bigg| =\bigg|\ell_{k,(i,j)}^{(t)} \sum_{p\in\cP_{i,j}(\vb)} [\theta_{i,j}^{(p)}(\vb)]^2\bigg|=O(b\alpha^2)=o\bigg(\frac{1}{\polylog(n)}\bigg).
\end{align*}

\textit{Completing the analysis for $\gamma_1^{(t)}(\vb, \vb)$.} Now we are able to complete the analysis on $\gamma_1^{(t)}(\vb,\vb)$ based on \eqref{eq:coefficient_update_v_v_mixup}:
\begin{align*}
\gamma_1^{(t)}(\vb, \vb) &= \frac{1}{n^2}\sum_{i,j\in[n]}\ell_{1,(i,j)}^{(t)} \sum_{p\in[P]} \theta_{i,j}^{(p)}(\vb)\cdot\la\xb_{i,j}^{(p)},\vb\ra\notag\\
&= \frac{1}{n^2}\bigg[\sum_{(i,j)\in\cS_{0,0}^{+,+}}\ell_{1,(i,j)}^{(t)} \sum_{p\in[P]} \theta_{i,j}^{(p)}(\vb)\cdot\la\xb_{i,j}^{(p)},\vb\ra \notag\\
&\qquad + \sum_{(i,j)\in\cS_{0,1}^{+,+}\cup\cS_{1,0}^{+,+}}\ell_{1,(i,j)}^{(t)} \sum_{p\in[P]} \theta_{i,j}^{(p)}(\vb)\cdot\la\xb_{i,j}^{(p)},\vb\ra\notag\\
&\qquad + \sum_{(i,j)\in\cS_{0,0}^{+,-}\cup\cS_{0, 1}^{+, -}\cup\cS_{0,0}^{-, +}\cup\cS_{0,1}^{-,+}}\ell_{1,(i,j)}^{(t)} \sum_{p\in[P]} \theta_{i,j}^{(p)}(\vb)\cdot\la\xb_{i,j}^{(p)},\vb\ra\notag\\
&\qquad + \sum_{i,j\in\cS_0^-\cup\cS_1^+\cup\cS_1^-}\ell_{1,(i,j)}^{(t)} \sum_{p\in[P]} \theta_{i,j}^{(p)}(\vb)\cdot\la\xb_{i,j}^{(p)},\vb\ra\bigg]\notag\\
& = \frac{1}{n^2}\bigg[\Theta(1)\cdot|\cS_{0,0}^{+,+}| + \Theta(1)\cdot |\cS_{0,1}^{+,+}\cup\cS_{1,0}^{+,+}| + \Theta(1)\cdot |\cS_{0,0}^{+,-}\cup\cS_{0,1}^{+,-}\cup\cS_{0,0}^{-,+}\cup\cS_{0,1}^{-,+}|\notag\\
&\qquad \pm o\bigg(\frac{1}{\polylog(n)}\bigg)\cdot|\cS_{0,0}^{-,-}\cup\cS_{0,1}^{-,+}\cup\cS_{0,1}^{-,-}\cup\cS_{1,0}^{+,-}\cup\cS_{1,1}^{+,+}\cup\cS_{1,1}^{+,-}\cup\cS_{1,0}^{-,-}\cup\cS_{1,1}^{-,+}\cup\cS_{1,1}^{-,-}|\bigg]\notag\\
& = \frac{1}{n^2}\bigg[\Theta(n^2) \pm o\bigg(\frac{n^2}{\polylog(n)}\bigg)\bigg]\notag\\
&=\Theta(1).
\end{align*}

\paragraph{Proof for $\gamma_k^{(t)}(\ub,\vb)$.}
The next step is to characterize $\gamma_k^{(t)}(\ub,\vb)$. We will split the entire mixed training dataset into the following classes:
\begin{itemize}
    \item $i\in\cS_0^+$ and $j\in\cS_0^-$, and $i\in\cS_0^-$ and $j\in\cS_0^+$, i.e., $\cS_{0,0}^{+,-}\cup\cS_{0,0}^{-,+}$.
    \item all $(i,j)\not\in\cS_{0,0}^{+,-}\cup\cS_{0,0}^{-,+}$.
\end{itemize}

We first recall the formula of $\gamma_1^{(t)}(\ub,\vb)$ (see Proposition \ref{prop:update_mixup}):
\begin{align}\label{eq:formula_correct_u_v_mixup}
\gamma_1^{(t)}(\ub,\vb) = \frac{1}{n^2}\sum_{i,j\in[n]}\ell_{1,(i,j)}^{(t)}\sum_{p\in[P]}\theta_{i,j}^{(p)}(\ub)\cdot\la\xb_{i,j}^{(p)},\vb\ra.
\end{align}

\textit{Analysis on the data $(i,j)\in\cS_{0,0}^{+,-}\cup\cS_{0,0}^{-,+}$.}
Since $\cS_{0,0}^{+,-}$ and $\cS_{0,0}^{-,+}$ are symmetric: i.e., for any $(i,j)\in\cS_{0,0}^{+,-}$, we have $(j,i)\in\cS_{0,0}^{-,+}$ and vise versa. Then we will handle data $\xb_{i,j}$ and  $\xb_{j,i}$ together by studying the following quantity:
\begin{align*}
*&:=\ell_{1,(i,j)}^{(t)} \sum_{p\in[P]}\theta_{i,j}^{(p)}(\ub)\cdot\la\xb_{i,j}^{(p)},\vb\ra + \ell_{1,(j,i)}^{(t)} \sum_{p\in[P]}\theta_{j,i}^{(p)}(\ub)\cdot\la\xb_{j,i}^{(p)},\vb\ra\notag\\
& = \ell_{1,(i,j)}^{(t)}\sum_{p\in[P]}\theta_{i,j}^{(p)}(\ub)\cdot \theta_{i,j}^{(p)}(\vb)+ \ell_{1,(j,i)}^{(t)} \sum_{p\in[P]}\theta_{j,i}^{(p)}(\ub)\cdot \theta_{j,i}^{(p)}(\vb).
\end{align*}
Note that we will only consider the patch that contains both $\ub$ and $\vb$. Then consider a data patch $\xb_{i,j}^{(p)}$ satisfy this condition: $\xb_i^{(p)}=\alpha_i\vb$ and $\xb_j^{(p)}=\alpha_j\ub$, where $\alpha_i,\alpha_j\in\{\alpha, 1\}$, which further leads to $\xb_{i,j}^{(p)} = \lambda\alpha_i\vb+(1-\lambda)\alpha_j\ub$ and $\xb_{j,i}^{(p)} = \lambda\alpha_j\ub+(1-\lambda)\alpha_i\vb$. Accordingly, it further gives
\begin{align*}
\theta_{i,j}^{(p)}(\ub)\cdot\theta_{i,j}^{(p)}(\vb) = (1-\lambda)\alpha_j\cdot \alpha_i\lambda_i = \lambda\alpha_j\cdot(1-\lambda)\alpha_i =  \theta_{j,i}^{(p)}(\ub)\cdot\theta_{j,i}^{(p)}(\vb).
\end{align*}
Additionally, for any $p\in\cP_{i,j}(\vb)$, we have at most $\Theta(1)$ among them satisfy $\theta_{j,i}^{(p)}(\ub)=\Theta(1)$ and at most $\Theta(1)$ among them satisfy $\theta_{j,i}^{(p)}(\vb)=\Theta(1)$, while the remaining, with size at most $2b$, can only give $\theta_{j,i}^{(p)}(\ub),\theta_{j,i}^{(p)}(\vb)=\Theta(\alpha)$. This implies that
\begin{align*}
\sum_{p\in[P]}\theta_{j,i}^{(p)}(\ub)\cdot\theta_{j,i}^{(p)}(\vb) = O(1) + O(\alpha) +O(b\alpha^2) = O(1).
\end{align*}
Therefore, applying the above equations, we can get that
\begin{align*}
* &= \ell_{1,(i,j)}^{(t)}\sum_{p\in\cP_{i,j}(\vb)}\big[\theta_{i,j}^{(p)}(\ub)\cdot\theta_{i,j}^{(p)}(\vb) - \theta_{j,i}^{(p)}(\ub)\cdot\theta_{j,i}^{(p)}(\vb)\big] + \big[\ell_{1,(i,j)}^{(t)}+\ell_{1,(j,i)}^{(t)}\big]\cdot \sum_{p\in[P]}\theta_{j,i}^{(p)}(\ub)\cdot\theta_{j,i}^{(p)}(\ub)\notag\\
& = \big[\ell_{1,(i,j)}^{(t)}+\ell_{1,(j,i)}^{(t)}\big]\cdot O(1).
\end{align*}
Further note that in the initial phase we have $\ell_{1,(i,j)}^{(t)}+\ell_{1,(j,i)}^{(t)} = O(\zeta)$, we consequently get 
\begin{align*}
|*|=\bigg|\ell_{1,(i,j)}^{(t)} \sum_{p\in[P]}\theta_{i,j}^{(p)}(\ub)\cdot\la\xb_{i,j}^{(p)},\vb\ra + \ell_{1,(j,i)}^{(t)} \sum_{p\in[P]}\theta_{j,i}^{(p)}(\ub)\cdot\la\xb_{j,i}^{(p)},\vb\ra\bigg| = O(\zeta).
\end{align*}

\textit{Analysis on the remaining data $(i,j)\not\in\cS_{0,0}^{+,-}\cup\cS_{0,0}^{-,+}$.} 
In this case, we note that there are no data patches that satisfy $\theta_{j,i}^{(p)}(\vb)=\Theta(1)$ and $\theta_{j,i}^{(p)}(\ub)=\Theta(1)$ simultaneously. Therefore, for any data $\xb_{i,j}$, there will exist at most $\Theta(1)$ patches that satisfy $\theta_{j,i}^{(p)}(\vb)\cdot\theta_{j,i}^{(p)}(\ub)=\alpha $ and at most $2b$ patches satisfying $\theta_{j,i}^{(p)}(\vb)\cdot\theta_{j,i}^{(p)}(\ub)=\alpha^2 $, while the remaining patches will give $\theta_{j,i}^{(p)}(\vb)\cdot\theta_{j,i}^{(p)}(\ub)=0 $. Therefore, we can get that
\begin{align*}
\sum_{p\in[P]}\theta_{i,j}^{(p)}(\ub)\cdot\la\xb_{i,j}^{(p)},\vb\ra & = \sum_{p\in[P]}\theta_{i,j}^{(p)}(\ub)\cdot \theta_{i,j}^{(p)}(\ub)=\Theta(1)\cdot \alpha + 2b\alpha^2 = O(\alpha),
\end{align*}
where the last equality follows from the setting of the data distribution that $b\alpha<1$. 

\textit{Completing the analysis for $\gamma_1^{(t)}(\ub,\vb)$.} By \eqref{eq:formula_correct_u_v_mixup} and using the fact that $|\ell_{1,(i,j)}^{(t)}|\le 1$, we have
\begin{align*}
|\gamma_1^{(t)}(\ub,\vb)| &= \frac{1}{n^2}\bigg|\sum_{i,j\in[n]}\ell_{1,(i,j)}^{(t)}\sum_{p\in[P]}\theta_{i,j}^{(p)}(\ub)\cdot\theta_{i,j}^{(p)}(\vb)\bigg| \notag\\
&= \frac{1}{n^2}\cdot \big[|\cS_{0,0}^{+,-}\cup\cS_{0,0}^{-,+}|\cdot O(\zeta) + \big(n^2-|\cS_{0,0}^{+,-}\cup\cS_{0,0}^{-,+}|\big)\cdot O(\alpha)\big]\notag\\
& = O(\zeta + \alpha).
\end{align*}

\paragraph{Proof for $\gamma_1^{(t)}(\vb', \vb)$.} We then tend to characterize $\gamma_1^{(t)}(\vb', \vb)$. We will consider the following two classes of data:
\begin{itemize}
    \item $(i,j)\in\cS_{0,1}^{+, +}\cup\cS_{1,0}^{+, +}$
    \item all $(i,j)\not\in \cS_{0,1}^{+, +}\cup\cS_{1,0}^{+, +}$.
\end{itemize}

\textit{Analysis on the data $(i,j)\in\cS_{0,1}^{+, +}\cup\cS_{1,0}^{+, +}$}
First, it is easy to see that with probability at least $1-1/\poly(n)$, we have $|\cS_{0,1}^{+, +}\cup\cS_{1,0}^{+, +}|=O(\rho n^2)$.
For this class of data, with probability $\Theta(1/P)$ we have the data $\xb_{i,j}^{(p)}$ has a constant number of patches that satisfy $\theta_{i,j}^{(p)}(\vb)\cdot \theta_{i,j}^{(p)}(\vb')=\Theta(1)$. Besides, by Lemma \ref{lemma:occurance_mixed_patch}, we have with probability at least $1-1/\poly(n)$, there are $\Theta(b\rho n^2/P)$ patches are the mixture of $\alpha\vb$ and $\vb'$, leading to $\theta_{i,j}^{(p)}(\vb)\cdot \theta_{i,j}^{(p)}(\vb')=\Theta(\alpha)$. The remaining patches will give $\theta_{i,j}^{(p)}(\vb)\cdot \theta_{i,j}^{(p)}(\vb')=0$. Combine the above results, we can get
\begin{align*}
\sum_{(i,j)\in\cS_{0,1}^{+, +}\cup\cS_{1,0}^{+, +}}\ell_{1,(i,j)}^{(t)}\sum_{p\in\cP} \theta_{i,j}^{(p)}(\vb)\cdot \theta_{i,j}^{(p)}(\vb') = \Theta\bigg(\frac{\rho n^2}{P}\bigg)  + \Theta\bigg(\frac{b\alpha\rho n^2}{P}\bigg) = \Theta\bigg(\frac{\rho n^2}{P}\bigg)
\end{align*}
where we use the fact that $b\alpha=o(1)$.  

\textit{Analysis on the remaining data} Particular, we will only consider the data $(i,j)\in\cS_{0,1}^{-,+}\cup\cS_{1,1}^{-,+}\cup\cS_{1, 0}^{+,-}$ since otherwise there is no data containing the rare feature vector $\vb'$. Moreover, note that for this class of data we only have $\theta_{i,j}^{(p)}(\vb)\cdot \theta_{i,j}^{(p)}(\vb')=O(\alpha)$ since there is no data consisting of common feature patch (but only contain feature noise $\alpha\vb$). Therefore, similar to the previous analysis, we can get that, by Lemma \ref{lemma:occurance_mixed_patch}, with probability at least $1-1/\poly(n)$, there are $\Theta(b\rho n^2/P)$  patches that give $\theta_{i,j}^{(p)}(\vb)\cdot \theta_{i,j}^{(p)}(\vb')=\Theta(\alpha)$, which consequently leads to
\begin{align*}
\sum_{(i,j)\in\cS_{0,1}^{-,+}\cup\cS_{1,1}^{-,+}\cup\cS_{1, 0}^{+,-}}\bigg|\ell_{1,(i,j)}^{(t)}\sum_{p\in\cP}\theta_{i,j}^{(p)}(\vb)\cdot \theta_{i,j}^{(p)}(\vb') \bigg|= O\bigg(\frac{b\alpha\rho n^2}{P}\bigg).
\end{align*}

\textit{Completing the analysis for $\gamma_1^{(t)}(\vb',\vb)$.} Completing the previous analysis, we have
\begin{align*}
\gamma_1^{(t)}(\vb',\vb) &= \frac{1}{n^2}\sum_{i,j\in[n]}\ell_{1,(i,j)}^{(t)}\sum_{p\in[P]}\theta_{i,j}^{(p)}(\vb')\cdot\theta_{i,j}^{(p)}(\vb) = \Theta\bigg(\frac{\rho }{P}\bigg)\pm O\bigg(\frac{b\alpha\rho }{P}\bigg)= \Theta\bigg(\frac{\rho }{P}\bigg).
\end{align*}

\paragraph{Proof for $\gamma_1^{(t)}(\ub',\vb)$.} Regarding the coefficient $\gamma_1^{(t)}(\ub',\vb)$, we consider two cases (1) mixup between $\ub'$ and $\vb$; (2) mixup between $\ub'$ and $\alpha\vb$. Then it can be seen that the first cases cover the data $(i,j)\in\cS_{1,0}^{-, +}$ and $(i,j)\in\cS_{0,1}^{+, -}$, which is equivalent to the dataset $\{(i,j), (j,i): (i,j)\in\cS_{1,0}^{+,-}\}$. Therefore, we will handle the data $(i,j)$ and $(j,i)$ together in this case. In particular, we have
\begin{align*}
&\ell_{1,(i,j)}^{(t)}\sum_{p\in[P]}\theta_{i,j}^{(p)}(\ub')\cdot \theta_{i,j}^{(p)}(\vb) + \ell_{1,(j,i)}^{(t)}\sum_{p\in[P]}\theta_{j,i}^{(p)}(\ub')\cdot \theta_{j,i}^{(p)}(\vb)\notag\\
& = \ell_{1,(i,j)}^{(t)}\sum_{p\in[P]}\big[\theta_{i,j}^{(p)}(\ub')\cdot \theta_{i,j}^{(p)}(\vb) - \theta_{j,i}^{(p)}(\ub')\cdot \theta_{j,i}^{(p)}(\vb)\big] + \big[\ell_{1,(i,j)}^{(t)}+\ell_{1,(j,i)}^{(t)}\big]\cdot\sum_{p\in[P]}\theta_{j,i}^{(p)}(\ub')\cdot \theta_{j,i}^{(p)}(\vb).
\end{align*}
It is clear that the first term on the R.H.S. of the above equation is zero since in case (1) 
\begin{align*}
\theta_{i,j}^{(p)}(\ub')\cdot \theta_{i,j}^{(p)}(\vb) = \theta_{j,i}^{(p)}(\ub')\cdot \theta_{j,i}^{(p)}(\vb) = \lambda(1-\lambda).
\end{align*}
Regarding the second term, we can use Lemma \ref{lemma:occurance_mixed_patch} and get that the number of patches falling in case (1) is $\Theta(\rho n^2/P)$. Then using the fact that $|\ell_{1,(i,j)}^{(t)} + \ell_{1,(i,j)}^{(t)}|=O(\zeta)$ can lead to the final bound for case (1). 

Regarding case (2), we can follow the analysis for $\gamma_1^{(t)}(\vb',\vb)$, which relies on the fact that $\theta_{i,j}^{(p)}(\ub)\cdot\theta_{i,j}^{(p)}(\vb')=\Theta(\alpha)$. Therefore, we can finally get
\begin{align*}
|\gamma_1^{(t)}(\ub,\vb')|&=\bigg|\frac{1}{n^2}\sum_{(i,j)\in\cS_{0,1}^{-,+}\cup\cS_{1,0}^{+,-}} \ell_{1,(i,j)}^{(t)}\sum_{p\in[P]}\theta_{i,j}^{(p)}(\ub)\cdot \theta_{i,j}^{(p)}(\vb')\bigg|\notag\\
&= O\bigg(\frac{\rho }{P}\bigg)\cdot\Theta(\zeta) + O\bigg(\frac{b\alpha\rho }{P}\bigg)\notag\\
&= \Theta\bigg(\frac{\zeta\rho  }{P}\bigg),
\end{align*}
where we use the fact that $\zeta = \omega(b\alpha)$. 


\paragraph{Proof for $\gamma_1^{(t)}(\bxi_s^{(q)},\vb)$.}
Finally, we will study $\gamma_1^{(t)}(\bxi_s^{(q)},\vb)$. Recall its formula in \eqref{eq:expansion_mixed_data} we can get 
\begin{align*}
\gamma_1^{(t)}(\bxi_s^{(q)},\vb) = \frac{1}{n^2}\sum_{i,j\in[n]}\ell_{1,(i,j)}^{(t)}\sum_{p\in[P]}\theta_{i,j}^{(p)}(\bxi_s^{(q)})\cdot\theta_{i,j}^{(p)}(\vb).
\end{align*}

Then it can be seen that the noise vector $\bxi_s^{(p)}$ will appear in $2n-1$ mixup data patches. By Lemma \ref{lemma:occurance_mixed_patch}, we have with probability at least $1-1/\poly(n)$,  $\Theta(1/P)$ fraction of them are mixed with $\vb$ and $O(b/P)$ fraction of them are mixed with $\alpha\vb$. Therefore, we can get that
\begin{align*}
\gamma_1^{(t)}(\bxi_s^{(q)},\vb) = \underbrace{\frac{1}{n^2}\sum_{p=q,i=s || p=q, j=s}\ell_{1,(i,j)}^{(t)}\theta_{i,j}^{(p)}(\bxi_s^{(q)})\cdot\theta_{i,j}^{(p)}(\vb)}_{I_1} + \underbrace{\frac{1}{n^2}\sum_{p\neq q|| i\neq s, j\neq s}\ell_{1,(i,j)}^{(t)}\sum_{p\in[P]}\theta_{i,j}^{(t)}(\bxi_s^{(q)})\cdot\theta_{i,j}^{(p)}(\vb)}_{I_2},
\end{align*}
where it holds that
\begin{align*}
|I_1| \le \frac{1}{n^2}\cdot \big[ \Theta(n/P) + \Theta(b\alpha/P)\big] = \Theta\bigg(\frac{1}{Pn}\bigg),
\end{align*}
and
\begin{align*}
|I_2| \le \tilde O\bigg(\frac{P}{d^{1/2}}\bigg),
\end{align*}
where we use the fact that $b\alpha=o(1)$ and $\theta_{i,j}^{(t)}(\bxi_s^{(q)})=\tilde O(d^{-1/2})$ for all $i\neq s $ and $j\neq s$. This further implies that 
\begin{align*}
|\gamma_1^{(t)}(\bxi_s^{(q)},\vb)| = O\bigg(\frac{1}{Pn}\bigg)
\end{align*}
since we have assumed that $d\ge P^4n^2$.
\end{proof}

We can also get a similar result for the learning of common feature $\ub$.
\begin{lemma}\label{lemma:feature_learning_coefficients_u_mixup}
Assume $\max_{k\in[2], (i,j)\in\cS} |F_k(\Wb^{(t)}; \xb_{i,j})|\le \zeta=o\big(\frac{1}{\polylog(n)}\big)$, then recalling the update form in Proposition \ref{prop:update_mixup}, we have for any $r\in[m]$, $q\in[P]$, and $s\in[n]$,
\begin{align*}
&\gamma_2^{(t)}(\ub, \ub) = \Theta(1), \quad |\gamma_2^{(t)}(\vb, \ub)| =  O(\zeta+\alpha), \quad |\gamma_2^{(t)}(\vb',\ub)| = O(\zeta\rho /P), \notag\\
&|\gamma_2(\ub',\ub)| = O(\rho/P), \quad |\gamma_2^{(t)}(\bxi_s^{(q)}, \ub)| = \tilde O\big(1/(Pn)\big).
\end{align*}

\end{lemma}

\subsubsection{Incorrect Common Feature Learning}
In this part, we will study the incorrect common feature learning, i.e., quantifying the inner products $\la\wb_{2,r}^{(t)},\vb\ra$ and $\la\wb_{1,r}^{(t)},\ub\ra$.

\begin{lemma}\label{lemma:feature_learning_coefficients_v_incorrect_mixup}
Assume $\max_{k\in[2], (i,j)\in\cS} |F_k(\Wb^{(t)}; \xb_{i,j})|\le \zeta=o\big(\frac{1}{\polylog(n)}\big)$, then recalling the update form in Proposition \ref{prop:update_mixup}, we have
\begin{align*}
&\gamma_2^{(t)}(\vb, \vb) = -\Theta(1), \quad |\gamma_2^{(t)}(\ub, \vb)| =  O(\zeta+\alpha), \quad |\gamma_2^{(t)}(\vb',\vb)| = O(\rho /P), \notag\\
&|\gamma_2^{(t)}(\ub',\vb)| = O(\zeta\rho/P), \quad |\gamma_2^{(t)}(\bxi_s^{(q)}, \vb)| = \tilde O\big(1/(Pn)\big).
\end{align*}

\end{lemma}
\begin{proof}[Proof of Lemma \ref{lemma:feature_learning_coefficients_v_incorrect_mixup}]
Recall the definition of $\gamma_2^{(t)}(\vb, \vb)$, we have
\begin{align*}
\gamma_2^{(t)}(\vb,\vb) = \frac{1}{n^2}\sum_{i,j\in[n]}\ell_{2,(i,j)}^{(t)}\sum_{p\in[P]}[\theta_{i,j}^{(p)}(\vb)]^2.
\end{align*}
Then comparing with the previous analysis on $\gamma_2^{(t)}(\vb, \vb)$, the only difference is to replace $\ell_{1,(i,j)}^{(t)}$ to $\ell_{2,(i,j)}^{(t)} = - \ell_{2,(i,j)}^{(t)}$. Therefore, we can immediately get that $\gamma_2^{(t)}(\vb,\vb)=-\gamma_1^{(t)}(\vb,\vb)=-\Theta(1)$.

Regarding other terms that are bounded in terms of their absolute values, we can get the same results as in Theorem \ref{lemma:feature_learning_coefficients_v_mixup}. This completes the proof.
\end{proof}
Similarly, we can get the following results for $\ub$.
\begin{lemma}\label{lemma:feature_learning_coefficients_u_incorrect_mixup}
Assume $\max_{k\in[2], (i,j)\in\cS} |F_k(\Wb^{(t)}; \xb_{i,j})|\le \zeta=o\big(\frac{1}{\polylog(n)}\big)$, then recalling the update form in Proposition \ref{prop:update_mixup}, we have
\begin{align*}
&\gamma_1^{(t)}(\ub, \ub) = -\Theta(1), \quad |\gamma_1^{(t)}(\vb, \ub)| =  O(\zeta+\alpha), \quad |\gamma_1^{(t)}(\vb',\ub)| = O(\rho /P), \notag\\
&|\gamma_1(\ub',\ub)| = O(\zeta\rho/P), \quad |\gamma_1^{(t)}(\bxi_s^{(q)}, \ub)| = \tilde O\big(1/(Pn)\big).
\end{align*}

\end{lemma}

\subsubsection{Rare Feature Learning}
In this part, we will study the rare feature learning, i.e., quantifying the inner products $\la\wb_{1,r}^{(t)},\vb'\ra$ and $\la\wb_{2,r}^{(t)},\ub'\ra$. 
\begin{lemma}\label{lemma:feature_learning_coefficients_v'_mixup}
Assume $\max_{k\in[2], (i,j)\in\cS} |F_k(\Wb^{(t)}; \xb_{i,j})|\le \zeta$ for some $\zeta =o\big(\frac{1}{\polylog(n)}\big)$ and $\zeta > b\alpha$, then recalling the update form in Proposition \ref{prop:update_mixup}, we have
\begin{align*}
&\gamma_1^{(t)}(\vb', \vb') = \Theta(\rho), \quad \gamma_1^{(t)}(\vb, \vb') = \Theta(\rho/P), \quad|\gamma_1^{(t)}(\ub,\vb')| = O(\zeta\rho /P), \notag\\
&|\gamma_1^{(t)}(\ub',\vb')| = O(\zeta\rho^2/P), \quad |\gamma_1^{(t)}(\bxi_s^{(q)}, \vb')| = \tilde O\big(\rho/(Pn)\big).
\end{align*}

\end{lemma}

\begin{proof}[Proof of Lemma \ref{lemma:feature_learning_coefficients_v'_mixup}]
Recalling the definition of $\gamma_1^{(t)}(\vb',\vb')$:
\begin{align*}
\gamma_1^{(t)}(\vb',\vb') = \frac{1}{n^2}\sum_{i,j\in[n]}\ell_{1,(i,j)}^{(t)}\sum_{p\in[P]}[\theta_{i,j}^{(p)}(\vb')]^2.
\end{align*}
Note that the rare feature $\vb'$ will not appear in the form of feature noise, then we will only need to focus on the mixed data $(i,j)$ with either $i\in\cS_1^+$ or $j\in\cS_1^+$, where the rare feature can only appear in the form of $\vb$, $\lambda\vb$, or $(1-\lambda)\vb$. Particularly, regarding the data $(i,j)\in\cS_{1,1}^{+,+}\cup\cS_{1,0}^{+, +}\cup_{0,1}^{+,+}$, let $\cP_{i,j}^*(\vb')$ be the set of patches that contain the feature $\vb'$, we have $|\cP_{i,j}^*(\vb')|=\Theta(1)$ and then
\begin{align*}
\ell_{1, (i,j)}^{(t)}\cdot\sum_{p\in[P]}[\theta_{i,j}^{(p)}(\vb')]^2 = \ell_{1, (i,j)}^{(t)}\cdot\sum_{p\in \cP_{i,j}^*(\vb)}[\theta_{i,j}^{(p)}(\vb')]^2 = \Theta(1),
\end{align*}
where we use the fact that $\ell_{1,(i,j)}^{(t)}=\Theta(1)$ for any $(i,j)\in\cS_{1,1}^{+, +}$.

Regarding the data $(i,j)\in\cS_{1,0}^{+, -}\cup\cS_{1,1}^{+,-}$, we will consider $(i,j)$ and $(j,i)$ together. Particularly, we have
\begin{align*}
\ell_{1, (i,j)}^{(t)}\cdot\sum_{p\in[P]}[\theta_{i,j}^{(p)}(\vb')]^2 + \ell_{1, (j,i)}^{(t)}\cdot\sum_{p\in[P]}[\theta_{j,i}^{(p)}(\vb')]^2 &= \underbrace{\ell_{1,(i,j)}^{(t)}\cdot\sum_{p\in[P]}\Big[[\theta_{i,j}^{(p)}(\vb')]^2 - [\theta_{j,i}^{(p)}(\vb')]^2\Big]}_{I_1}\notag\\
&\qquad + \underbrace{\big[\ell_{1,(i,j)}^{(t)}+\ell_{1,(j,i)}^{(t)}\big]\cdot\sum_{p\in[P]}[\theta_{j,i}^{(p)}(\vb')]^2}_{I_2}.
\end{align*}
Then using the same definition of $\cP_{i,j}^*(\vb)$, we have for any $p\in\cP_{i,j}^*(\vb)$, it holds that $\theta_{i,j}^{(p)}(\vb')=\lambda$ and $\theta_{j,i}^{(p)}(\vb')=1-\lambda$, then
\begin{align*}
I_1 = \Theta(1)\cdot |\cP_{i,j}^*(\vb')|\cdot [\lambda^2 - (1-\lambda)^2] = \Theta(1).
\end{align*}
Regarding $I_2$, we can use the condition that the neural network output is upper bounded by $\zeta$, then 
\begin{align*}
|I_2| = \bigg|\big[\lambda - 0.5 + 0.5 - \lambda \pm O(\zeta)\big]\cdot \sum_{p\in\cP_{i,j}^*(\vb')}[\theta_{j,i}^{(p)}(\vb')]^2\bigg| = O(\zeta).
\end{align*}
Therefore, combining these results for $I_1$ and $I_2$, we can get
\begin{align*}
\ell_{1, (i,j)}^{(t)}\cdot\sum_{p\in[P]}[\theta_{i,j}^{(p)}(\vb')]^2 + \ell_{1, (j,i)}^{(t)}\cdot\sum_{p\in[P]}[\theta_{j,i}^{(p)}(\vb')]^2 = I_1 + I_2 = \Theta(1).
\end{align*}
To complete the analysis, we have
\begin{align*}
\gamma_1^{(t)}(\vb',\vb') &= \frac{1}{n^2}\sum_{i,j\in[n]}\sum_{p\in[P]}[\theta_{i,j}^{(p)}(\vb')]^2\notag\\
& = \frac{1}{n^2}\sum_{i\in\cS_1^+,j\in[n] || i\in[n],j\in\cS_1^+}\sum_{p\in[P]}[\theta_{i,j}^{(p)}(\vb')]^2\notag\\
& = |\cS_{1,1}^{+,+}\cup\cS_{1,0}^{+, +}\cup_{0,1}^{+,+}| + |\cS_{1,0}^{+, -}\cup\cS_{1,1}^{+,-}|\notag\\
& = \Theta(\rho).
\end{align*}

The characterization of $\gamma_1(\vb,\vb')$ and $\gamma_1(\ub,\vb')$ will be exactly the same as $\gamma_1(\vb',\vb)$ and $\gamma_1(\vb', \ub)$ due to the  fact that $\gamma_1(\ab,\bb)=\gamma_1(\bb,\ab)$. Therefore, we can apply Lemmas \ref{lemma:feature_learning_coefficients_v_mixup} and \ref{lemma:feature_learning_coefficients_u_incorrect_mixup} to get the desired results.

Regarding the proof for $\gamma_1^{(t)}(\ub',\vb')$, we will follow a similar proof for $\gamma_1^{(t)}(\ub',\vb)$ in Lemma \ref{lemma:feature_learning_coefficients_v_mixup}, while two differences need to be considered: (1) the rare feature vectors $\ub'$ and $\vb'$ will not appear in the form of feature noise, thus we only need to consider the data $(i,j)\in\cS_{1,1}^{+,-}\cup\cS_{1,1}^{-,+}$; (2) the cardinality of the critical subset of data satisfies $|\cS_{1,1}^{+,-}\cup\cS_{1,1}^{-,+}|=\rho^2 n^2$. Therefore, for any $(i,j)\in\cS_{1,1}^{+,-}$, we have
\begin{align*}
&\ell_{1,(i,j)}^{(t)}\sum_{p\in[P]}\theta_{i,j}^{(p)}(\ub')\cdot \theta_{i,j}^{(p)}(\vb') + \ell_{1,(j,i)}^{(t)}\sum_{p\in[P]}\theta_{j,i}^{(p)}(\ub')\cdot \theta_{j,i}^{(p)}(\vb')\notag\\
& = \ell_{1,(i,j)}^{(t)}\sum_{p\in[P]}\big[\theta_{i,j}^{(p)}(\ub')\cdot \theta_{i,j}^{(p)}(\vb') - \theta_{j,i}^{(p)}(\ub')\cdot \theta_{j,i}^{(p)}(\vb')\big] + \big[\ell_{1,(i,j)}^{(t)}+\ell_{1,(j,i)}^{(t)}\big]\cdot\sum_{p\in[P]}\theta_{j,i}^{(p)}(\ub')\cdot \theta_{j,i}^{(p)}(\vb').
\end{align*}
It is easy to see that $\theta_{i,j}^{(p)}(\ub')\cdot \theta_{i,j}^{(p)}(\vb') = \theta_{j,i}^{(p)}(\ub')\cdot \theta_{j,i}^{(p)}(\vb')=\lambda(1-\lambda)$. Besides, we have in total $\rho^2n^2/P$ patches that consist of both $\ub'$ and $\vb'$. This further implies that
\begin{align*}
|\gamma_1^{(t)}(\ub',\vb')| &= \bigg|\frac{1}{n^2}\sum_{(i,j)\in\cS_{1,1}^{+,-}\cup\cS_{1,1}^{-,+}}\big[\ell_{1,(i,j)}^{(t)}+\ell_{1,(j,i)}^{(t)}\big]\cdot\sum_{p\in[P]}\theta_{j,i}^{(p)}(\ub')\cdot\theta_{j,i}^{(p)}(\vb')\bigg|\notag\\
& = O\bigg(\frac{\rho^2}{P}\bigg)\cdot O(\zeta) \notag\\
& = O\bigg(\frac{\zeta\rho^2}{P}\bigg),
\end{align*}
where we use the fact that $|\ell_{1,(i,j)}^{(t)}+\ell_{1,(j,i)}^{(t)}|=O(\zeta)$.

Lastly, we will characterize $\gamma_1^{(t)}(\bxi_s^{(q)},\vb')$. First recall its definition:
\begin{align*}
\gamma_1^{(t)}(\bxi_s^{(q)},\vb') &= \frac{1}{n^2}\sum_{i,j\in[n]}\ell_{1,(i,j)}^{(t)}\sum_{p\in[P]}\theta_{i,j}^{(p)}(\bxi_s^{(q)})\cdot\theta_{i,j}^{(p)}(\vb')\notag\\
& = \underbrace{\frac{1}{n^2}\sum_{p=q,i=s||p=q,j=s}\ell_{1,(i,j)}^{(t)}\theta_{i,j}^{(p)}(\bxi_s^{(q)})\cdot\theta_{i,j}^{(p)}(\vb')}_{I_1} +\underbrace{\frac{1}{n^2}\sum_{p\neq q||i\neq s, j\neq s}\ell_{1,(i,j)}^{(t)}\sum_{p\in[P]}\theta_{i,j}^{(p)}(\bxi_s^{(q)})\cdot\theta_{i,j}^{(p)}(\vb')}_{I_2} .
\end{align*}
Note that for any fixed $\bxi_s^{(p)}$, it will be mixed with $n$ data patches in total, while, by Lemma \ref{lemma:occurance_mixed_patch}, we know that there are only $\Theta(\rho/P)$ fraction among them are $\vb'$. Using the fact that $|\ell_{1,(i,j)}^{(t)}|\le 1$, we have
\begin{align*}
|I_1|\le \frac{1}{n} \cdot \Theta(\rho/P) = O\bigg(\frac{\rho}{Pn}\bigg).
\end{align*}
Besides, note that $|\theta_{i,j}^{(p)}(\bxi_s^{(q)})| = \tilde O(d^{-1/2})$ if $i,j\neq s$ or $p\neq q$, we have
\begin{align*}
|I_2|\le \tilde O\bigg(\frac{\rho P}{d^{1/2}}\bigg) = O\bigg(\frac{\rho}{Pn}\bigg) ,
\end{align*}
where the last equality is by the assumption that $d\ge P^4n^2$.
Combining the above results for $I_1$ and $I_2$, we can get
\begin{align*}
|\gamma_1^{(t)}(\bxi_s^{(q)},\vb')| = O\bigg(\frac{\rho}{Pn}\bigg).
\end{align*}
\end{proof}

Following the exactly same procedure, we can get the following results regarding the learning of $\ub'$.
\begin{lemma}\label{lemma:feature_learning_coefficients_u'_mixup}
Assume $\max_{k\in[2], (i,j)\in\cS} |F_k(\Wb^{(t)}; \xb_{i,j})|\le \zeta$ for some $\zeta =o\big(\frac{1}{\polylog(n)}\big)$ and $\zeta > b\alpha$, then recalling the update form in Proposition \ref{prop:update_mixup}, we have
\begin{align*}
&\gamma_2^{(t)}(\ub', \ub') = \Theta(\rho), \quad \gamma_2^{(t)}(\ub, \ub') = \Theta(\rho/P), \quad|\gamma_2^{(t)}(\vb,\ub')| = O(\zeta\rho /P), \notag\\
&|\gamma_2^{(t)}(\vb',\ub')| = O(\zeta\rho^2/P), \quad |\gamma_2^{(t)}(\bxi_s^{(q)}, \ub')| = \tilde O\big(\rho/(Pn)\big).
\end{align*}

\end{lemma}

\subsubsection{Incorrect Rare Feature Learning}
In contrast to the previous section that studies $\la\wb_{1,r}^{(t)},\vb'\ra$ and $\la\wb_{2,r}^{(t)},\ub'\ra$, the incorrect rare feature learning aims to characterize the quantities $\la\wb_{2,r}^{(t)},\vb'\ra$ and $\la\wb_{1,r}^{(t)},\ub'\ra$. Similar to the proof of Lemmas \ref{lemma:feature_learning_coefficients_v_incorrect_mixup} and  \ref{lemma:feature_learning_coefficients_u_incorrect_mixup}, we only need to replace $\ell_{1,(i,j)}^{(t)}$ with $\ell_{2,(i,j)}^{(t)}=-\ell_{1,(i,j)}^{(t)}$ or $\ell_{2,(i,j)}^{(t)}$ with $\ell_{1,(i,j)}^{(t)}=-\ell_{2,(i,j)}^{(t)}$. Based on this, the update of $\la\wb_{2,r}^{(t)},\vb'\ra$ and $\la\wb_{1,r}^{(t)},\ub'\ra$ in each iteration are characterized in the following lemmas.

\begin{lemma}\label{lemma:feature_learning_coefficients_v'_incorrect_mixup}
Assume $\max_{k\in[2], (i,j)\in\cS} |F_k(\Wb^{(t)}; \xb_{i,j})|\le \zeta$ for some $\zeta =o\big(\frac{1}{\polylog(n)}\big)$ and $\zeta > b\alpha$, then recalling the update form in Proposition \ref{prop:update_mixup}, we have
\begin{align*}
&\gamma_2^{(t)}(\vb', \vb') = -\Theta(\rho), \quad \gamma_2^{(t)}(\vb, \vb') = -\Theta(\rho/P), \quad|\gamma_2^{(t)}(\ub,\vb')| = O(\zeta\rho /P), \notag\\
&|\gamma_2^{(t)}(\ub',\vb')| = O(\zeta\rho^2/P), \quad |\gamma_2^{(t)}(\bxi_s^{(q)}, \vb')| = \tilde O\big(\rho/(Pn)\big).
\end{align*}

\end{lemma}
\begin{lemma}\label{lemma:feature_learning_coefficients_u'_incorrect_mixup}
Assume $\max_{k\in[2], (i,j)\in\cS} |F_k(\Wb^{(t)}; \xb_{i,j})|\le \zeta$ for some $\zeta =o\big(\frac{1}{\polylog(n)}\big)$ and $\zeta > b\alpha$, then recalling the update form in Proposition \ref{prop:update_mixup}, we have
\begin{align*}
&\gamma_1^{(t)}(\ub', \ub') = -\Theta(\rho), \quad \gamma_1^{(t)}(\ub, \ub') = -\Theta(\rho/P), \quad|\gamma_1^{(t)}(\vb,\ub')| = O(\zeta\rho /P), \notag\\
&|\gamma_1^{(t)}(\vb',\ub')| = O(\zeta\rho^2/P), \quad |\gamma_1^{(t)}(\bxi_s^{(q)},\ub'| = \tilde O\big(\rho/(Pn)\big).
\end{align*}

\end{lemma}

\subsubsection{Noise Learning}

\begin{lemma}\label{lemma:noise_learning_coefficients_incorrect_mixup}
Assume $\max_{k\in[2], (i,j)\in\cS} |F_k(\Wb^{(t)}; \xb_{i,j})|\le \zeta$ for some $\zeta =o\big(\frac{1}{\polylog(n)}\big)$ and $\zeta > b\alpha$, then recalling the update form in Proposition \ref{prop:update_mixup}, for any $\bxi_s^{(q)}$ with $y_s=1$, we have
\begin{align*}
&\gamma_1^{(t)}(\bxi_s^{(q)}, \bxi_s^{(q)}) = \frac{d\sigma_p^2\cdot [n\lambda^3 - (2\lambda-1)(1-\lambda)^2]}{2n^2} \pm \tilde O\bigg(\frac{\zeta d\sigma_p^2}{n}\bigg)\notag\\
&|\gamma_1^{(t)}(\vb, \bxi_s^{(q)})| = O\big(d\sigma_p^2/(Pn)\big), \quad|\gamma_1^{(t)}(\ub, \bxi_s^{(q)})| = O\big(d\sigma_p^2/(Pn)\big),\quad |\gamma_1^{(t)}(\vb', \bxi_s^{(q)})| = O\big(d\sigma_p^2\rho/(Pn)\big), \notag\\
&|\gamma_1^{(t)}(\ub', \bxi_s^{(q)})| = O\big(d\sigma_p^2\rho/(Pn)\big), \quad |\gamma_1^{(t)}(\bxi_i^{(q)}, \bxi_s^{(q)})| = \ind[y_i=y_s] \cdot \frac{\lambda(1-\lambda)d\sigma_p^2}{n^2} \pm O\bigg(\frac{\zeta d\sigma_p^2}{n^2}\bigg).
\end{align*}

\end{lemma}

\begin{proof}[Proof of Lemma \ref{lemma:noise_learning_coefficients_incorrect_mixup}]
Without loss of generality, we assume $y_s=1$. According to the definition of $\gamma_1^{(t)}\big(\bxi_s^{(q)},\bxi_s^{(q)}\big)$, we have
\begin{align*}
\gamma_1^{(t)}\big(\bxi_s^{(q)},\bxi_s^{(q)}\big) &= \frac{1}{n^2}\sum_{i,j\in[n]}\ell_{1,(i,j)}^{(t)}\sum_{p\in[P]}[\theta_{i,j}^{(p)}(\bxi_s^{(q)})]^2\cdot \|\bxi_s^{(q)}\|_2^2\notag\\
& = \frac{ \|\bxi_s^{(q)}\|_2^2}{n^2}\cdot \bigg(\sum_{i\in[n]}\ell_{1,(s, i)}^{(t)}[\theta_{s, i}^{(q)}(\bxi_s^{(q)})]^2+\sum_{i\neq s}\ell_{1,(i, s)}^{(t)}[\theta_{i, s}^{(q)}(\bxi_s^{(q)})]^2\bigg)\notag\\
& = \frac{ \|\bxi_s^{(q)}\|_2^2}{n^2}\cdot \bigg(\lambda^2\cdot\sum_{i\in[n]}\ell_{1,(s,i)}^{(t)}+(1-\lambda)^2\cdot\sum_{i\neq s}\ell_{1,(i,s)}^{(t)}\bigg)\notag\\
& = \frac{ \|\bxi_s^{(q)}\|_2^2}{n^2}\cdot\big[ 0.5n\lambda^3 - (\lambda-0.5)(1-\lambda)^2 \pm O(n\zeta)  \big]\notag\\
& = \frac{ \|\bxi_s^{(q)}\|_2^2\cdot[n\lambda^3 - (2\lambda-1)(1-\lambda)^2]}{2n^2} \pm O\bigg(\frac{\zeta \|\bxi_s^{(q)}\|_2^2}{n}\bigg),
 \end{align*}
 where the second equation is due to the fact that only $\xb_{i,s}$ or $\xb_{s,i}$ will contain the component of $\bxi_s^{(q)}$, the fourth inequality holds since we assume there have $n/2$ positive samples and $n/2$ negative samples in the training data.
Moreover, note that $\bxi_s^{(q)}\sim N(0,\sigma_p^2\Ib)$,  applying union bound over all $s\in[n]$ and $p\in[P]$, we can get that with probability at least $1-1/\poly(n)$, we have
\begin{align*}
\big|\|\bxi_s^{(q)}\|_2^2-d\sigma_p^2\big| \le \polylog(n)\cdot d^{1/2}\sigma_p^2. 
\end{align*}
Therefore, it follows that for all $s\in[n]$ and $p\in[P]$, with probability at least $1-1/\poly(n)$,
\begin{align*}
\gamma_1^{(t)}\big(\bxi_s^{(q)},\bxi_s^{(q)}\big)& = \frac{d\sigma_p^2\cdot [n\lambda^3 - (2\lambda-1)(1-\lambda)^2]}{2n^2} \pm \tilde O\bigg(\frac{d^{1/2}\sigma_p^2}{n}+\frac{\zeta d\sigma_p^2}{n}\bigg)\notag\\
&= \frac{d\sigma_p^2\cdot [n\lambda^2-(1-\lambda)^2]}{2n^2} \pm \tilde O\bigg(\frac{\zeta d\sigma_p^2}{n}\bigg),
\end{align*}
where we use the fact that $\zeta = \omega(d^{-1/2})$.

Regarding $\gamma_1^{(t)}(\bxi_i^{(q)}, \bxi_s^{(q)})$, we have
\begin{align*}
\gamma_1^{(t)}(\bxi_i^{(q)}, \bxi_s^{(q)}) &= \frac{\lambda(1-\lambda)}{n^2}\cdot[\ell_{1,(i,s)}^{(t)}+\ell_{1,(s,i)}^{(t)}]\cdot \|\bxi_s^{(q)}\|_2^2\notag\\
& =\ind[y_i=y_s] \cdot \frac{\lambda(1-\lambda)d\sigma_p^2}{n^2} \pm O\bigg(\frac{\zeta d\sigma_p^2}{n^2}\bigg).
\end{align*}

Regarding the remaining quantities, we can directly apply the aforementioned lemmas on the learning of common and rare features, since the following holds 
\begin{align*}
\gamma_1^{(t)}(\ab, \bxi_s^{(q)}) = \gamma_1^{(t)}(\bxi_s^{(q)},\ab)\cdot \|\bxi_s^{(q)}\|_2^2 = \gamma_1^{(t)}(\bxi_s^{(q)},\ab)\cdot\tilde O(d\sigma_p^2),
\end{align*}
where $\ab\in\{\vb,\ub,\vb',\ub'\}$.
This completes the proof.

\end{proof}


\subsection{Outcome of Phase 1 Mixup Training.}

In this part, we will provide the outcome of Phase 1 mixup training. 


We first recall Proposition \ref{prop:update_mixup} and Lemma \ref{lemma:feature_learning_coefficients_v_mixup} to obtain the learning dynamics of the common feature vector $\vb$.
\begin{align*}
\la\wb_{1,r}^{(t+1)},\vb\ra  &= \la\wb_{1,r}^{(t)},\vb\ra - \eta\cdot\la\nabla_{\wb_{1,r}}L_\cS(\Wb^{(t)}),\vb\ra\notag\\
& = \big[1 + \eta\gamma_1^{(t)}(\vb,\vb)\big]\cdot\la\wb_{1,r}^{(t)},\vb\ra + \eta\gamma_1^{(t)}(\ub,\vb)\cdot\la\wb_{1,r}^{(t)},\ub\ra + \eta\gamma_1^{(t)}(\vb',\vb)\cdot\la\wb_{1,r}^{(t)},\vb'\ra \notag\\
&\qquad + \eta\gamma_1^{(t)}(\ub',\vb)\cdot\la\wb_{1,r}^{(t)}, \ub'\ra + \sum_{i=1}^n\sum_{p\in[P]}\eta\gamma_1^{(t)}(\bxi_i^{(p)},\vb)\cdot\la\wb_{1,r}^{(t)},\bxi_i^{(p)}\ra.
\end{align*}

Then it can be seen that the most complicated part in the above update form is the composition of noise learning, i.e., $\la\wb_{1,r}^{(t)},\bxi_i^{(p)}\ra$.The following lemma provides an upper bound on the term $\sum_{i=1}^n\sum_{p\in[P]}\eta\gamma_1^{(t)}(\bxi_i^{(p)},\vb)\cdot\la\wb_{1,r}^{(t)},\bxi_i^{(p)}\ra$, which will leverage the randomness of $\bxi_i^{(p)}$ at the initialization.
\begin{lemma}\label{lemma:update_sum_noise}
Assume $\max_{k\in[2], (i,j)\in\cS} |F_k(\Wb^{(t)}; \xb_{i,j})|\le \zeta$ for some $\zeta\in\big[\omega\big((nP)^{-1/2}\big), o\big(\frac{1}{\polylog(n)}\big)\big]$. Let $z_t:=\sum_{i=1}^n\sum_{p\in[P]}\gamma_1^{(t)}(\bxi_i^{(p)},\vb)\cdot\la\wb_{1,r}^{(t)},\bxi_i^{(p)}\ra$, then we have with probability at least $1-1/\poly(n)$, for all $t = O\big(n\eta^{-1}/(d\sigma_p^2)\big)$, we have
\begin{align*}
|z_t|&\le O\bigg(\frac{d^{1/2}\sigma_0\sigma_p}{P^{1/2}n^{1/2}}\bigg) +O\bigg(\frac{\zeta}{Pn}\bigg)\cdot\sum_{s=1}^n\sum_{p\in[P]}|\la\wb_{1,r}^{(t)},\bxi_s^{(p)}\ra|+ O\bigg(\frac{\eta d\sigma_p^2}{Pn}\bigg)\cdot\sum_{\tau=0}^{t-1}\big[|\la\wb_{1,r}^{(\tau)},\vb\ra|+|\la\wb_{1,r}^{(\tau)},\ub\ra|\big] \notag\\
&\qquad+ O\bigg(\frac{\eta\rho d\sigma_p^2}{Pn}\bigg)\cdot\sum_{\tau=0}^{t-1}\big[|\la\wb_{1,r}^{(\tau)},\vb'\ra|+|\la\wb_{1,r}^{(\tau)},\ub'\ra|\big]+ O\bigg(\frac{\eta\zeta d\sigma_p^2}{n^2}\bigg)\cdot\sum_{\tau=0}^{t-1}\sum_{p\in[P]}\sum_{s=1}^n |\la\wb_{1,r}^{(\tau)},\bxi_s^{(p)}\ra|.
\end{align*}
\end{lemma}
\begin{proof}
Based on the definition of $z_t$, we can conduct the following decomposition:
\begin{align*}
z_t &= \sum_{i=1}^n\sum_{p\in[P]}\gamma_1^{(t)}(\bxi_i^{(p)},\vb)\cdot\la\wb_{1,r}^{(t)},\bxi_i^{(p)}\ra\notag\\
& = \frac{1}{n^2}\sum_{i'=1}^n\sum_{q\in[P]}\sum_{i,j\in[n]}\ell_{1,(i,j)}^{(t)}\sum_{p\in[P]}\theta_{i,j}^{(p)}(\bxi_{i'}^{(q)})\cdot\theta_{i,j}^{(p)}(\vb)\cdot\la\wb_{1,r}^{(t)},\bxi_{i'}^{(q)}\ra.
\end{align*}
Note that during the initial training phase $\ell_{1, (i,j)}^{(t)}$ is close to the constant $l_{1,(i,j)}\in\{0.5, -0.5, 0.5-\lambda, \lambda-0.5\}$, which is independent of the random noise vectors $\{\bxi\}$ and random initial weights $\{\wb_{1,r}^{(0)}\}_{r\in[m]}$. Then using the fact that $|\ell_{1, (i,j)}^{(t)}-l_{1,(i,j)}| = O(\zeta)$, we can get
\begin{align*}
|z_0| & \le  \bigg|\underbrace{\frac{1}{n^2}\sum_{i'=1}^n\sum_{q\in[P]}\sum_{i,j\in[n]}l_{1,(i,j)}\sum_{p\in[P]}\theta_{i,j}^{(p)}(\bxi_{i'}^{(q)})\cdot\theta_{i,j}^{(p)}(\vb)\cdot\la\wb_{1,r}^{(0)},\bxi_{i'}^{(q)}\ra}_{I_1}\bigg| \notag\\
&\qquad + \underbrace{\bigg|\frac{1}{n^2}\sum_{i'=1}^n\sum_{q\in[P]}\sum_{i,j\in[n]}[\ell_{1,(i,j)}^{(0)}-l_{1,(i,j)}]\sum_{p\in[P]}\theta_{i,j}^{(p)}(\bxi_{i'}^{(q)})\cdot\theta_{i,j}^{(p)}(\vb)\cdot\la\wb_{1,r}^{(0)},\bxi_{i'}^{(q)}\ra\bigg|}_{I_2}.
\end{align*}
Regarding $I_1$, note that $\ell_{1,(i,j)}$, $\theta_{i,j}^{(p)}(\bxi_{i'}^{(q)})$, and $\theta_{i,j}^{(p)}(\vb)$ are independent of the random noise vectors $\{\bxi\}$ and random initial weights $\{\wb_{1,r}^{(0)}\}_{r\in[m]}$. Besides, note that the inner products $\{\la\wb_{1,r}^{(0)},\bxi_{i'}^{(q)}\ra\}_{i'\in[n], q\in[P]}$ are independent conditioning on $\wb_{1,r}^{(0)}$ and for all $i'\in[n]$ and $q\in[P]$, . We can apply standard concentration arguments to get the upper bound of $I_1$. Before approaching this, we first apply Lemma \ref{lemma:occurance_mixed_patch} and follow the similar proof of Lemma \ref{lemma:feature_learning_coefficients_v_mixup}, and obtain that with probability at least $1-1/\poly(n)$
\begin{align}\label{eq:bound_sum_coefficients_noise}
\bigg|\frac{1}{n^2}\sum_{i,j\in[n]}l_{1,(i,j)}\sum_{p\in[P]}\theta_{i,j}^{(p)}(\bxi_{i'}^{(q)})\cdot\theta_{i,j}^{(p)}(\vb)\bigg| = O\bigg(\frac{1}{nP}\bigg).
\end{align}
Then performing the following decomposition on $I_1$ according to the value of $y_{i'}$:
\begin{align*}
I_1 &= \underbrace{\frac{1}{n^2}\sum_{i':y_{i'}=1}\sum_{q\in[P]}\sum_{i,j\in[n]}l_{1,(i,j)}\sum_{p\in[P]}\theta_{i,j}^{(p)}(\bxi_{i'}^{(q)})\cdot\theta_{i,j}^{(p)}(\vb)\cdot\la\wb_{1,r}^{(0)},\bxi_{i'}^{(q)}\ra}_{I_1^{(1)}}\notag\\
&\qquad + \underbrace{\frac{1}{n^2}\sum_{i':y_{i'}=2}\sum_{q\in[P]}\sum_{i,j\in[n]}l_{1,(i,j)}\sum_{p\in[P]}\theta_{i,j}^{(p)}(\bxi_{i'}^{(q)})\cdot\theta_{i,j}^{(p)}(\vb)\cdot\la\wb_{1,r}^{(0)},\bxi_{i'}^{(q)}\ra}_{I_1^{(2)}}.
\end{align*}
Therefore, note that conditioning on $\wb_{1,r}^{(0)}$, the quantity $\la\wb_{1,r}^{(0)},\bxi_{i'}^{(q)}\ra$ is $\|\wb_{1,r}^{(0)}\|_2\cdot \sigma_p$-subGaussian, by \eqref{eq:bound_sum_coefficients_noise}, we can immediately get that both $I_1^{(1)}$ and $I_1^{(2)}$ are $\|\wb_{1,r}^{(0)}\|_2\cdot \sigma_p\cdot (nP)^{-1/2}$-subGuassian. Then using the fact that $\wb_{1,r}^{(0)}\in N(0,\sigma_0^2\Ib)$, we can get that with probability at least $1-1/\poly(n)$, 
\begin{align}\label{eq:bound_I1_noise_sum}
|I_1^{(1)}|,|I_1^{(2)}|  \le \tilde O\bigg(\frac{d^{1/2}\sigma_0\sigma_p}{(nP)^{1/2}}\bigg).
\end{align}

Regarding $I_2$, we can also apply Lemma \ref{lemma:occurance_mixed_patch} and follow the similar proof of Lemma \ref{lemma:feature_learning_coefficients_v_mixup}, then with probability at least $1-1/\poly(n)$,
\begin{align*}
\bigg|\frac{1}{n^2}\sum_{i,j\in[n]}[\ell_{1,(i,j)}^{(0)}-l_{1,(i,j)}]\sum_{p\in[P]}\theta_{i,j}^{(p)}(\bxi_{i'}^{(q)})\cdot\theta_{i,j}^{(p)}(\vb)\bigg| = O\bigg(\frac{\max_{i,j}|\ell_{1,(i,j)}^{(0)}-l_{1,(i,j)}|}{nP}\bigg)=O\bigg(\frac{\zeta}{nP}\bigg).
\end{align*}
This further implies that
\begin{align}\label{eq:bound_I2_noise_sum}
I_2 \le O\bigg(\frac{\zeta}{nP}\bigg) \cdot \sum_{i'=1}^n\sum_{q\in[P]}|\la\wb_{1,r}^{(0)},\bxi_{i'}^{(q)}\ra|\le \tilde O\big(\zeta d^{1/2}\sigma_0\sigma_p\big).
\end{align}
where we use the fact that $\wb_{1,r}^{(0)}\sim N(0,\sigma_0^2\Ib)$ and $\bxi_{i'}^{(q)}\sim N(0,\sigma_p^2\Ib)$. Combining \eqref{eq:bound_I1_noise_sum} and \eqref{eq:bound_I2_noise_sum} leads to
\begin{align*}
|z_0|\le \tilde O\bigg(\frac{d^{1/2}\sigma_0\sigma_p}{(nP)^{1/2}} + \zeta d^{1/2}\sigma_0\sigma_p\bigg)=O\big(\zeta d^{1/2}\sigma_0\sigma_p\big).
\end{align*}
where we use the condition that $\zeta = \omega\big((nP)^{-1/2}\big)$. 

Next we will move on to study the update of $z_t$ using the update results of $\la\wb_{1,r}^{(t)},\bxi_i^{(p)}\ra$ in Lemma \ref{lemma:noise_learning_coefficients_incorrect_mixup}. Particularly, we can again use the quantities $l_{1,(i,j)}$'s and get the following decomposition
\begin{align*}
z_t &= \underbrace{\frac{1}{n^2}\sum_{i'=1}^n\sum_{q\in[P]}\sum_{i,j\in[n]}l_{1,(i,j)}\sum_{p\in[P]}\theta_{i,j}^{(p)}(\bxi_{i'}^{(q)})\cdot\theta_{i,j}^{(p)}(\vb)\cdot\la\wb_{1,r}^{(t)},\bxi_{i'}^{(q)}\ra}_{I_3} \notag\\
& \qquad + \underbrace{\frac{1}{n^2}\sum_{i'=1}^n\sum_{q\in[P]}\sum_{i,j\in[n]}\big[\ell_{1,(i,j)}^{(t)} - l_{1,(i,j)}\big]\sum_{p\in[P]}\theta_{i,j}^{(p)}(\bxi_{i'}^{(q)})\cdot\theta_{i,j}^{(p)}(\vb)\cdot\la\wb_{1,r}^{(t)},\bxi_{i'}^{(q)}\ra}_{I_4}.
\end{align*}
Recall the update results of $\la\wb_{1,r}^{(t)},\bxi_i^{(p)}\ra$ in Lemma \ref{lemma:noise_learning_coefficients_incorrect_mixup}: for any $y_i=1$,
\begin{align}\label{eq:update_positive_noise}
\la\wb_{1,r}^{(t+1)},\bxi_i^{(p)}\ra   &= \bigg[1 + \eta\cdot\bigg(\frac{d\sigma_p^2\cdot [n\lambda^3 - (2\lambda-1)(1-\lambda)^2]}{2n^2} \pm \tilde O\bigg(\frac{\zeta d\sigma_p^2}{n}\bigg)\bigg)\bigg]\cdot \la\wb_{1,r}^{(t)},\bxi_i^{(p)}\ra\pm O\bigg(\frac{\eta d\sigma_p^2}{Pn}\bigg)\cdot\la\wb_{1,r}^{(t)},\vb\ra  \notag\\
& \qquad\pm O\bigg(\frac{\eta d\sigma_p^2}{Pn}\bigg)\cdot\la\wb_{1,r}^{(t)},\ub\ra \pm O\bigg(\frac{\eta d\sigma_p^2\rho}{Pn}\bigg)\cdot\la\wb_{1,r}^{(t)},\vb'\ra \pm O\bigg(\frac{\eta d\sigma_p^2\rho}{Pn}\bigg)\cdot\la\wb_{1,r}^{(t)},\ub'\ra\notag\\
&\qquad +  O\bigg(\frac{\eta\lambda(1-\lambda)d\sigma_p^2}{n^2}\bigg)\sum_{s:y_s=1} \la\wb_{1,r}^{(t)},\bxi_s^{(p)}\ra \pm O\bigg(\frac{\eta\zeta d\sigma_p^2}{n^2}\bigg)\cdot\sum_{s=1}^n |\la\wb_{1,r}^{(t)},\bxi_s^{(p)}\ra| .
\end{align}
For any $y_i=2$, we have
\begin{align}\label{eq:update_negative_noise}
\la\wb_{1,r}^{(t+1)},\bxi_i^{(p)}\ra   &= \bigg[1 - \eta\cdot\bigg(\frac{d\sigma_p^2\cdot [n\lambda^3 - (2\lambda-1)(1-\lambda)^2]}{2n^2} \pm \tilde O\bigg(\frac{\zeta d\sigma_p^2}{n}\bigg)\bigg)\bigg]\cdot \la\wb_{1,r}^{(t)},\bxi_i^{(p)}\ra \pm O\bigg(\frac{\eta d\sigma_p^2}{Pn}\bigg)\cdot\la\wb_{1,r}^{(t)},\vb\ra\notag\\
& \qquad \pm O\bigg(\frac{\eta d\sigma_p^2}{Pn}\bigg)\cdot\la\wb_{1,r}^{(t)},\ub\ra \pm O\bigg(\frac{\eta d\sigma_p^2\rho}{Pn}\bigg)\cdot\la\wb_{1,r}^{(t)},\vb'\ra\pm O\bigg(\frac{\eta d\sigma_p^2\rho}{Pn}\bigg)\cdot\la\wb_{1,r}^{(t)},\ub'\ra\notag\\
&\qquad - O\bigg(\frac{\eta\lambda(1-\lambda)d\sigma_p^2}{n^2}\bigg)\cdot\sum_{s:y_s=1} \la\wb_{1,r}^{(t)},\bxi_s^{(p)}\ra \pm O\bigg(\frac{\eta\zeta d\sigma_p^2}{n^2}\bigg)\cdot\sum_{s=1}^n |\la\wb_{1,r}^{(t)},\bxi_s^{(p)}\ra| .
\end{align}

We first prove the bound of the quantity $ \sum_{p\in[P]}\sum_{s:y_s=1}\la\wb_{1,r}^{(t)},\bxi_s^{(p)}\ra$. First, using the standard concentration result gives $|\sum_{p\in[P]}\sum_{s:y_s=1}\la\wb_{1,r}^{(0)},\bxi_s^{(p)}\ra|=\tilde O\big(d^{1/2}\sigma_0\sigma_pP^{1/2}n^{1/2}\big)$. Then, by the above update rule, we can get
\begin{align*}
\sum_{p\in[P]}\sum_{s:y_s=1}\la\wb_{1,r}^{(t+1)},\bxi_s^{(p)}\ra &= \bigg[1 + \Theta\bigg(\frac{\eta d\sigma_p^2}{n}\bigg)\bigg]\cdot \sum_{p\in[P]}\sum_{s:y_s=1}\la\wb_{1,r}^{(t)},\bxi_s^{(p)}\ra\notag\\
& \qquad \pm O\big(\eta d\sigma_p^2\big)\cdot\la\wb_{1,r}^{(t)},\vb\ra\pm O\big(\eta d\sigma_p^2\big)\cdot\la\wb_{1,r}^{(t)},\ub\ra \pm O\big(\eta\rho d\sigma_p^2\big)\cdot\la\wb_{1,r}^{(t)},\vb'\ra\notag\\
&\qquad\pm O\big(\eta\rho d\sigma_p^2\big)\cdot\la\wb_{1,r}^{(t)},\ub'\ra  \pm O\bigg(\frac{\eta\zeta d\sigma_p^2}{n}\bigg)\cdot\sum_{p\in[P]}\sum_{s=1}^n |\la\wb_{1,r}^{(t)},\bxi_s^{(p)}\ra|. 
\end{align*}
Then we can get that for any $t = O\big(n\eta^{-1}/(d\sigma_p^2)\big)$, we have
\begin{align*}
\bigg|\sum_{p\in[P]}\sum_{s:y_s=1}\la\wb_{1,r}^{(t)},\bxi_s^{(p)}\ra\bigg|&\le O\big(n^{1/2}P^{1/2}d^{1/2}\sigma_0\sigma_p\big)+ O(\eta d\sigma_p^2)\cdot\sum_{\tau=0}^{t-1}\big[|\la\wb_{1,r}^{(\tau)},\vb\ra|+|\la\wb_{1,r}^{(\tau)},\ub\ra|\big] \notag\\
&\qquad+ O(\eta\rho d\sigma_p^2)\cdot\sum_{\tau=0}^{t-1}\big[|\la\wb_{1,r}^{(\tau)},\vb'\ra|+|\la\wb_{1,r}^{(\tau)},\ub'\ra|\big] \notag\\
&\qquad + O\bigg(\frac{\eta\zeta d\sigma_p^2}{n}\bigg)\cdot\sum_{\tau=0}^{t-1}\sum_{p\in[P]}\sum_{s=1}^n |\la\wb_{1,r}^{(\tau)},\bxi_s^{(p)}\ra|.
\end{align*}
Moreover, similar result can be obtained for $\sum_{s:y_s=2}\la\wb_{1,r}^{(t)},\bxi_s^{(p)}\ra$ and we omit the proof here. 

Now we are ready to upper bound $I_3$. Particularly, let $\alpha_1^{(t)}$ and $\alpha_2^{(t)}$ be denoted as follows:
\begin{align*}
\alpha_1^{(t)} &= \frac{1}{n^2}\sum_{i': y_{i'}=1}^n\sum_{q\in[P]}\sum_{i,j\in[n]}l_{1,(i,j)}\sum_{p\in[P]}\theta_{i,j}^{(p)}(\bxi_{i'}^{(q)})\cdot\theta_{i,j}^{(p)}(\vb)\cdot\la\wb_{1,r}^{(t)},\bxi_{i'}^{(q)}\ra\notag\\
\alpha_2^{(t)} &= \frac{1}{n^2}\sum_{i': y_{i'}=2}^n\sum_{q\in[P]}\sum_{i,j\in[n]}l_{1,(i,j)}\sum_{p\in[P]}\theta_{i,j}^{(p)}(\bxi_{i'}^{(q)})\cdot\theta_{i,j}^{(p)}(\vb)\cdot\la\wb_{1,r}^{(t)},\bxi_{i'}^{(q)}\ra.
\end{align*}
Then it is clear that $I_3 = \alpha_1^{(t)} + \alpha_2^{(t)}$. Then by \eqref{eq:bound_sum_coefficients_noise} and \eqref{eq:update_positive_noise}, 
we can get
\begin{align*}
\alpha_1^{(t+1)} & = \bigg[1 + \Theta\bigg(\frac{\eta d\sigma_p^2}{n}\bigg)\bigg]\cdot \alpha_1^{(t)}\pm O\bigg(\frac{\eta d\sigma_p^2}{Pn}\bigg)\cdot\la\wb_{1,r}^{(t)},\vb\ra\pm O\bigg(\frac{\eta d\sigma_p^2}{Pn}\bigg)\cdot\la\wb_{1,r}^{(t)},\ub\ra \notag\\
& \qquad \pm O\bigg(\frac{\eta\rho d\sigma_p^2}{Pn}\bigg)\cdot\la\wb_{1,r}^{(t)},\vb'\ra\pm O\bigg(\frac{\eta\rho d\sigma_p^2}{Pn}\bigg)\cdot\la\wb_{1,r}^{(t)},\ub'\ra\notag\\
&\qquad  \pm O\bigg(\frac{\eta d\sigma_p^2}{n^2P}\bigg)\cdot \bigg|\sum_{p\in[P]}\sum_{s:y_s=1}\la\wb_{1,r}^{(t)},\bxi_s^{(p)}\ra\bigg|\pm O\bigg(\frac{\eta \zeta d\sigma_p^2}{n^2P}\bigg)\cdot \sum_{p\in[P]}\sum_{s=1}^n|\la\wb_{1,r}^{(t)},\bxi_s^{(p)}\ra|.
\end{align*}
Similarly, we can also obtain
\begin{align*}
\alpha_2^{(t+1)} & = \bigg[1 - \Theta\bigg(\frac{\eta d\sigma_p^2}{n}\bigg)\bigg]\cdot \alpha_2^{(t)}\pm O\bigg(\frac{\eta d\sigma_p^2}{Pn}\bigg)\cdot\la\wb_{1,r}^{(t)},\vb\ra\pm O\bigg(\frac{\eta d\sigma_p^2}{Pn}\bigg)\cdot\la\wb_{1,r}^{(t)},\ub\ra \notag\\
& \qquad \pm O\bigg(\frac{\eta\rho d\sigma_p^2}{Pn}\bigg)\cdot\la\wb_{1,r}^{(t)},\vb'\ra\pm O\bigg(\frac{\eta\rho d\sigma_p^2}{Pn}\bigg)\cdot\la\wb_{1,r}^{(t)},\ub'\ra\notag\\
&\qquad  \pm O\bigg(\frac{\eta d\sigma_p^2}{n^2P}\bigg)\cdot \bigg|\sum_{p\in[P]}\sum_{s:y_s=2}\la\wb_{1,r}^{(t)},\bxi_s^{(p)}\ra\bigg|\pm O\bigg(\frac{\eta \zeta d\sigma_p^2}{n^2P}\bigg)\cdot \sum_{p\in[P]}\sum_{s=1}^n|\la\wb_{1,r}^{(t)},\bxi_s^{(p)}\ra|.
\end{align*}
Then using the previous results on $\big|\sum_{p\in[P]}\sum_{s:y_s=1}\la\wb_{1,r}^{(t)},\bxi_s^{(p)}\ra\big|$ and $\big|\sum_{p\in[P]}\sum_{s:y_s=2}\la\wb_{1,r}^{(t)},\bxi_s^{(p)}\ra\big|$ and \eqref{eq:bound_I1_noise_sum}, we can get that for any $t=O\big(n\eta^{-1}/(d\sigma_p^2)\big)$,
\begin{align*}
|\alpha_1^{(t)}| &\le O\bigg(\frac{d^{1/2}\sigma_0\sigma_p}{P^{1/2}n^{1/2}}\bigg) + O\bigg(\frac{\eta d\sigma_p^2}{Pn}\bigg)\cdot\sum_{\tau=0}^{t-1}\big[|\la\wb_{1,r}^{(\tau)},\vb\ra|+|\la\wb_{1,r}^{(\tau)},\ub\ra|\big] \notag\\
&\qquad+ O\bigg(\frac{\eta\rho d\sigma_p^2}{Pn}\bigg)\cdot\sum_{\tau=0}^{t-1}\big[|\la\wb_{1,r}^{(\tau)},\vb'\ra|+|\la\wb_{1,r}^{(\tau)},\ub'\ra|\big]+ O\bigg(\frac{\eta\zeta d\sigma_p^2}{n^2P}\bigg)\cdot\sum_{\tau=0}^{t-1}\sum_{p\in[P]}\sum_{s=1}^n |\la\wb_{1,r}^{(\tau)},\bxi_s^{(p)}\ra|,
\end{align*}
where we use the upper bound of $|\alpha_1^{(0)}|$ provided in 
Similarly, we can obtain the same results for $\alpha_2^{(t)}$ as follows:
\begin{align*}
|\alpha_2^{(t)}| &\le O\bigg(\frac{d^{1/2}\sigma_0\sigma_p}{P^{1/2}n^{1/2}}\bigg) + O\bigg(\frac{\eta d\sigma_p^2}{Pn}\bigg)\cdot\sum_{\tau=0}^{t-1}\big[|\la\wb_{1,r}^{(\tau)},\vb\ra|+|\la\wb_{1,r}^{(\tau)},\ub\ra|\big] \notag\\
&\qquad+ O\bigg(\frac{\eta\rho d\sigma_p^2}{Pn}\bigg)\cdot\sum_{\tau=0}^{t-1}\big[|\la\wb_{1,r}^{(\tau)},\vb'\ra|+|\la\wb_{1,r}^{(\tau)},\ub'\ra|\big]+ O\bigg(\frac{\eta\zeta d\sigma_p^2}{n^2P}\bigg)\cdot\sum_{\tau=0}^{t-1}\sum_{p\in[P]}\sum_{s=1}^n |\la\wb_{1,r}^{(\tau)},\bxi_s^{(p)}\ra|.
\end{align*}
Combining the above results leads to the bound of $I_3$.

We will finally bound $I_4$ as follows: using the fact that $|\ell_{1,(i,j)}-l_{1,(i,j)}|=O(\zeta)$ and a similar characterization of \eqref{eq:bound_sum_coefficients_noise}, we can get
\begin{align*}
I_4 \le O\bigg(\frac{\zeta}{Pn}\bigg)\cdot\sum_{s=1}^n\sum_{p\in[P]}|\la\wb_{1,r}^{(t)},\bxi_s^{(p)}\ra|.
\end{align*}
Combining the above bounds on $I_3$ and $I_4$, we can finally get
\begin{align*}
|z_t| &\le |I_3| + |I_4|\notag\\
&\le O\bigg(\frac{d^{1/2}\sigma_0\sigma_p}{P^{1/2}n^{1/2}}\bigg) +O\bigg(\frac{\zeta}{Pn}\bigg)\cdot\sum_{s=1}^n\sum_{p\in[P]}|\la\wb_{1,r}^{(t)},\bxi_s^{(p)}\ra|+ O\bigg(\frac{\eta d\sigma_p^2}{Pn}\bigg)\cdot\sum_{\tau=0}^{t-1}\big[|\la\wb_{1,r}^{(\tau)},\vb\ra|+|\la\wb_{1,r}^{(\tau)},\ub\ra|\big] \notag\\
&\qquad+ O\bigg(\frac{\eta\rho d\sigma_p^2}{Pn}\bigg)\cdot\sum_{\tau=0}^{t-1}\big[|\la\wb_{1,r}^{(\tau)},\vb'\ra|+|\la\wb_{1,r}^{(\tau)},\ub'\ra|\big]+ O\bigg(\frac{\eta\zeta d\sigma_p^2}{n^2P}\bigg)\cdot\sum_{\tau=0}^{t-1}\sum_{p\in[P]}\sum_{s=1}^n |\la\wb_{1,r}^{(\tau)},\bxi_s^{(p)}\ra|.
\end{align*}

This completes the proof.

\end{proof}

Then the following lemma characterizes the growth of  common feature learning.
\begin{lemma}\label{lemma:dominance_v}
Assume $\max_{k\in[2], (i,j)\in\cS} |F_k(\Wb^{(t)}; \xb_{i,j})|\le \zeta$ for some $\zeta =o\big(d^{-1/2}\sigma_p^{-1}\big)$. Then for any $t=O\big(\polylog(n)/\eta\big)$ that satisfies this condition, we have with probability at least $1-1/\poly(n)$, there exists at least one $r\in[m]$ such that
\begin{align*}
\la\wb_{1,r}^{(t+1)},\vb\ra= \big[1 + \Theta(\eta)\big]\cdot \la\wb_{1,r}^{(t)},\vb\ra.
\end{align*}

\end{lemma}
\begin{proof} 
First, note that $\la\wb_{1,r}^{(0)},\vb\ra$ follows $N(0, \sigma_0^2)$, then it is easy to get that 
\begin{align}\label{eq:prob_largest_neuron}
\mathbb P\bigg[\max_{r\in[m]}|\la\wb_{1,r}^{(0)},\vb\ra|\ge \sigma_0\bigg] = 1 - \big(\mathbb P_{\xi\sim N(0,\sigma_0^2)}[|\xi|\le \sigma_0]\big)^m\ge 1-0.7^m \ge 1-1/\poly(n),
\end{align}
where the last inequality is by our assumption that $m=\polylog(n)>C\log(n)$ for some sufficiently large constant $C$.

Recall the update rule of $\la\wb_{1,r}^{(t)},\vb\ra$:
\begin{align*}
\la\wb_{1,r}^{(t+1)},\vb\ra  
& = \big[1 + \eta\gamma_1^{(t)}(\vb,\vb)\big]\cdot\la\wb_{1,r}^{(t)},\vb\ra + \eta\gamma_1^{(t)}(\ub,\vb)\cdot\la\wb_{1,r}^{(t)},\ub\ra + \eta\gamma_1^{(t)}(\vb',\vb)\cdot\la\wb_{1,r}^{(t)},\vb'\ra \notag\\
&\qquad + \eta\gamma_1^{(t)}(\ub',\vb)\cdot\la\wb_{1,r}^{(t)}, \ub'\ra + \sum_{i=1}^n\sum_{p\in[P]}\eta\gamma_1^{(t)}(\bxi_i^{(p)},\vb)\cdot\la\wb_{1,r}^{(t)},\bxi_i^{(p)}\ra.
\end{align*}
Taking absolute value on both sides leads to
\begin{align*}
|\la\wb_{1,r}^{(t+1)},\vb\ra|&\ge \big[1 + \eta\gamma_1^{(t)}(\vb,\vb)\big]\cdot|\la\wb_{1,r}^{(t)},\vb\ra| - \eta\big|\gamma_1^{(t)}(\ub,\vb)\cdot\la\wb_{1,r}^{(t)},\ub\ra\big| - \eta\big|\gamma_1^{(t)}(\vb',\vb)\cdot\la\wb_{1,r}^{(t)},\vb'\ra\big| \notag\\
&\qquad - \eta\big|\gamma_1^{(t)}(\ub',\vb)\cdot\la\wb_{1,r}^{(t)}, \ub'\ra\big| - \bigg|\sum_{i=1}^n\sum_{p\in[P]}\eta\gamma_1^{(t)}(\bxi_i^{(p)},\vb)\cdot\la\wb_{1,r}^{(t)},\bxi_i^{(p)}\ra\bigg|.
\end{align*}
Therefore, the next step is to show that these ``negative'' terms in the above inequality are dominated by $\eta\gamma_1^{(t)}(\vb,\vb)\cdot |\la\wb_{1,r}^{(t)},\vb\ra|$, i.e., showing that 
\begin{align*}
&\big|\gamma_1^{(t)}(\ub,\vb)\cdot\la\wb_{1,r}^{(t)},\ub\ra\big|, \big|\gamma_1^{(t)}(\vb',\vb)\cdot\la\wb_{1,r}^{(t)},\vb'\ra\big|, \big|\gamma_1^{(t)}(\ub',\vb)\cdot\la\wb_{1,r}^{(t)}, \ub'\ra\big| \ll |\la\wb_{1,r}^{(t)},\vb\ra|;\notag\\
&\bigg|\sum_{i=1}^n\sum_{p\in[P]}\eta\gamma_1^{(t)}(\bxi_i^{(p)},\vb)\cdot\la\wb_{1,r}^{(t)},\bxi_i^{(p)}\ra\bigg|\ll |\la\wb_{1,r}^{(t)},\vb\ra|,
\end{align*}
where we use our result in Lemma \ref{lemma:feature_learning_coefficients_v_mixup} that $\gamma_1^{(t)}(\vb,\vb)=\Theta(1)$. Then we are able to get that
\begin{align}\label{eq:updates_v_tmp}
|\la\wb_{1,r}^{(t+1)},\vb\ra|&=\big[1 + \eta\gamma_1^{(t)}(\vb,\vb) \pm o\big(1/\polylog(n)\big)\big]\cdot|\la\wb_{1,r}^{(t)},\vb\ra|\ge \big[1+\Theta(\eta)\big]\cdot|\la\wb_{1,r}^{(t)},\vb\ra|. 
\end{align}

Regarding the first three terms, we will prove them by mathematical induction on a stronger argument (recall that $\gamma_1^{(t)}(\vb,\vb)=\Theta(1)$, $|\gamma_1^{(t)}(\ub,\vb)|,|\gamma_1^{(t)}(\vb',\vb)|,|\gamma_1^{(t)}(\ub',\vb)|=o\big(1/\polylog(n)\big)$, according to Lemma \ref{lemma:feature_learning_coefficients_v_mixup}): we aim to verify the hypothesis
\begin{align}\label{eq:hypothesis_dominance_v}
 & |\la\wb_{1,r}^{(t)},\ub\ra\big|, \quad |\la\wb_{1,r}^{(t)},\vb'\ra\big| ,\quad |\la\wb_{1,r}^{(t)},\ub'\ra\big| \le  c\cdot\log^2(n)|\la\wb_{1,r}^{(t)},\vb\ra|,
\end{align}
where $c$ is some sufficiently small constant.

In particular, we can first consider the initialization where $t=0$, then by \eqref{eq:prob_largest_neuron} and standard concentration bound of Gaussian random variable, we have with probability at least $1-1/\poly(n)$,
\begin{align*}
&|\la\wb_{1,r}^{(0)},\vb\ra| = \Omega(\sigma_0),\quad \big|\la\wb_{1,r}^{(0)},\ub\ra\big| = O\big(\log(n)\sigma_0\big), \notag\\
&\big|\la\wb_{1,r}^{(0)},\vb'\ra\big| = O\big(\log(n)\sigma_0\big),\quad\big|\la\wb_{1,r}^{(0)},\vb'\ra\big| = O\big(\log(n)\sigma_0\big).
\end{align*}
Therefore, using the fact that $\zeta = o\big(d^{-1/2}\sigma_p^{-1}\big)$, it is easy to verify the hypothesis. We will then assume the hypothesis holds for all $\tau\le t$ and aim to verify it for $t+1$. Particularly, recall the update rules of $\la\wb_{1,r}^{(t)},\ub\ra$, we have
\begin{align}\label{eq:upperbound_w_u}
|\la\wb_{1,r}^{(t+1)},\ub\ra| &\le \big[1 - \eta\gamma_1^{(t)}(\ub,\ub)\big]\cdot|\la\wb_{1,r}^{(t)},\ub\ra| + \eta\big|\gamma_1^{(t)}(\vb,\ub)\cdot\la\wb_{1,r}^{(t)},\vb\ra\big| + \eta\big|\gamma_1^{(t)}(\vb',\vb)\cdot\la\wb_{1,r}^{(t)},\vb'\ra\big| \notag\\
&\qquad + \eta\big|\gamma_1^{(t)}(\ub',\ub)\cdot\la\wb_{1,r}^{(t)}, \ub'\ra\big| + \bigg|\sum_{i=1}^n\sum_{p\in[P]}\eta\gamma_1^{(t)}(\bxi_i^{(p)},\ub)\cdot\la\wb_{1,r}^{(t)},\bxi_i^{(p)}\ra\bigg|\notag\\
&\le |\la\wb_{1,r}^{(0)},\ub\ra| + \eta\sum_{\tau=0}^{t}\big|\gamma_1^{(\tau)}(\vb,\ub)\cdot\la\wb_{1,r}^{(\tau)},\vb\ra\big| + \eta\sum_{\tau=0}^{t}\big|\gamma_1^{(\tau)}(\vb',\vb)\cdot\la\wb_{1,r}^{(\tau)},\vb'\ra\big| \notag\\
&\qquad + \eta\sum_{\tau=0}^{t}\big|\gamma_1^{(\tau)}(\ub',\ub)\cdot\la\wb_{1,r}^{(\tau)}, \ub'\ra\big| + \sum_{\tau=0}^{t}\bigg|\sum_{i=1}^n\sum_{p\in[P]}\eta\gamma_1^{(\tau)}(\bxi_i^{(p)},\ub)\cdot\la\wb_{1,r}^{(\tau)},\bxi_i^{(p)}\ra\bigg|\notag\\
&\le O\big(\log(n)\sigma_0\big) + \tilde O\big(\eta(\zeta+\alpha)\big)\cdot\sum_{\tau=0}^t|\la\wb_{1,r}^{(\tau)},\vb\ra| + O\bigg(\frac{\eta\rho}{P}+\frac{t\rho\eta^2 d\sigma_p^2}{Pn}\bigg)\sum_{\tau=0}^t\big[|\la\wb_{1,r}^{(\tau)},\vb'\ra| +|\la\wb_{1,r}^{(\tau)},\ub'\ra| \big]\notag\\
& \qquad+ O\bigg(\frac{t \eta d^{1/2}\sigma_0\sigma_p}{P^{1/2}n^{1/2}}\bigg)+ O\bigg(\frac{t\eta^2 d\sigma_p^2}{Pn}\bigg)\cdot\sum_{\tau=0}^t\big[|\la\wb_{1,r}^{(\tau)},\vb\ra| +|\la\wb_{1,r}^{(\tau)},\ub\ra| \big]\notag\\
&\qquad  + O\bigg(\frac{\eta\zeta}{Pn}+\frac{t\eta^2\zeta d\sigma_p^2}{n^2P}\bigg)\cdot\sum_{\tau=0}^{t}\sum_{s=1}^n\sum_{p\in[P]}|\la\wb_{1,r}^{(\tau)},\bxi_s^{(p)}\ra|.
\end{align}
where the last inequality is by Lemma \ref{lemma:update_sum_noise}.
Then by \eqref{eq:update_positive_noise}, we have the following results regarding $|\la\wb_{1,r}^{(\tau)},\bxi_s^{(p)}\ra|$
\begin{align*}
\max_{i\in[n],p\in[P]}|\la\wb_{1,r}^{(\tau+1)}, \bxi_i^{(p)}\ra| &\le \bigg[1 + O\bigg(\frac{d\sigma_p^2}{n}\bigg)\bigg]\cdot \max_{i\in[n],p\in[P]}|\la\wb_{1,r}^{(\tau)}, \bxi_i^{(p)}\ra| +O\bigg(\frac{\eta d\sigma_p^2}{Pn}\bigg)\cdot\big[|\la\wb_{1,r}^{(\tau)},\vb\ra| + |\la\wb_{1,r}^{(\tau)},\ub\ra|\big]\notag\\
& \qquad+ O\bigg(\frac{\eta\rho d\sigma_p^2}{Pn}\bigg)\cdot\big[|\la\wb_{1,r}^{(\tau)},\vb'\ra| + |\la\wb_{1,r}^{(\tau)},\ub'\ra|\big]\notag\\
& =  O\big(d^{1/2}\sigma_0\sigma_p\big) + O\bigg(\frac{\eta d\sigma_p^2}{Pn}\bigg)\cdot\sum_{s=0}^{\tau}\big[|\la\wb_{1,r}^{(s)},\vb\ra| + |\la\wb_{1,r}^{(s)},\ub\ra|\big] \notag\\
&\qquad + O\bigg(\frac{\eta\rho d\sigma_p^2}{Pn}\bigg)\cdot\sum_{s=0}^{\tau}\big[|\la\wb_{1,r}^{(s)},\vb'\ra| + |\la\wb_{1,r}^{(s)},\ub'\ra|\big].
\end{align*}
Therefore,  we can accordingly get the following upper bound regarding the last term in the RHS of \eqref{eq:upperbound_w_u},
\begin{align}\label{eq:upperbound_sum_aboslute_noise}
\frac{1}{nP}\sum_{\tau=0}^{t}\sum_{s=1}^n\sum_{p\in[P]}|\la\wb_{1,r}^{(\tau)},\bxi_s^{(p)}\ra|&\le  \sum_{\tau=0}^t\max_{i\in[n],p\in[P]}|\la\wb_{1,r}^{(\tau)}, \bxi_i^{(p)}\ra|\notag\\
& \le O\big(td^{1/2}\sigma_0\sigma_p\big) + O\bigg(\frac{t\eta d\sigma_p^2}{Pn}\bigg)\cdot\sum_{\tau=0}^{t-1}\big[|\la\wb_{1,r}^{(\tau)},\vb\ra| + |\la\wb_{1,r}^{(\tau)},\ub\ra|\big] \notag\\
&\qquad + O\bigg(\frac{t\eta\rho d\sigma_p^2}{Pn}\bigg)\cdot\sum_{\tau=0}^{t-1}\big[|\la\wb_{1,r}^{(\tau)},\vb'\ra| + |\la\wb_{1,r}^{(\tau)},\ub'\ra|\big].
\end{align}
Then using the fact that $t\eta =O\big(\polylog(n)\big)$ and $n=\omega(d\sigma_p^2)$, we can further get the following on \eqref{eq:upperbound_w_u}
\begin{align*}
|\la\wb_{1,r}^{(t+1)},\ub\ra|&\le  O\big(\log(n)\sigma_0\big) + \tilde O\big(\eta(\zeta+\alpha)\big)\cdot\sum_{\tau=0}^t|\la\wb_{1,r}^{(\tau)},\vb\ra| + O\bigg(\frac{\eta\rho }{P}\bigg)\sum_{\tau=0}^t\big[|\la\wb_{1,r}^{(\tau)},\vb'\ra| +|\la\wb_{1,r}^{(\tau)},\ub'\ra| \big]\notag\\
& \qquad+ O\bigg(\frac{t \eta d^{1/2}\sigma_0\sigma_p}{P^{1/2}n^{1/2}}+t\eta\zeta d^{1/2}\sigma_0\sigma_p\bigg)+ O\bigg(\frac{\eta }{P}\bigg)\cdot\sum_{\tau=0}^t\big[|\la\wb_{1,r}^{(\tau)},\vb\ra| +|\la\wb_{1,r}^{(\tau)},\ub\ra| \big].
\end{align*}
Then according to the Hypothesis \ref{eq:hypothesis_dominance_v} for any $\tau\le t$, it is easy to get that
\begin{align*}
|\la\wb_{1,r}^{(t+1)},\ub\ra|\ge |\la\wb_{1,r}^{(t)},\ub\ra|\ge\ldots\ge|\la\wb_{1,r}^{(0)},\ub\ra|.
\end{align*}
Then we can get $|\la\wb_{1,r}^{(t+1)},\ub\ra|=\Omega(\sigma_0)$, applying  the fact that $t\eta=O(\polylog(n))$ further gives
\begin{align}\label{eq:characterize1}
O\big(\log(n)\sigma_0\big) +O\bigg(\frac{t \eta d^{1/2}\sigma_0\sigma_p}{P^{1/2}n^{1/2}}+t\eta\zeta d^{1/2}\sigma_0\sigma_p\bigg)\bigg) \bigg]=o(\log^2(n)\sigma_0) = o\big(\log^2(n)|\la\wb_{1,r}^{(t+1)},\ub\ra|\big).
\end{align}
Besides, note that the Hypothesis \ref{eq:hypothesis_dominance_v} holds for all $\tau\le t$, we have
\begin{align}\label{eq:characterize2}
|\la\wb_{1,r}^{(\tau)},\ub\ra|,|\la\wb_{1,r}^{(\tau)},\vb'\ra|,|\la\wb_{1,r}^{(\tau)},\ub'\ra|\le c\cdot \log^2(n)\cdot |\la\wb_{1,r}^{(\tau)},\vb\ra|\le \log^2(n)\cdot |\la\wb_{1,r}^{(t+1)},\vb\ra|,
\end{align}
we can immediately get that
\begin{align}\label{eq:characterize3}
&\tilde O\big(\eta(\zeta+\alpha)\big)\cdot\sum_{\tau=0}^t|\la\wb_{1,r}^{(\tau)},\vb\ra|\le \tilde O\big(t\eta(\zeta+\alpha)\big)\cdot|\la\wb_{1,r}^{(t+1)},\vb\ra| =o\big( \log^2(n)\cdot |\la\wb_{1,r}^{(t+1)},\vb\ra|\big)\notag\\
 &O\bigg(\frac{\eta\rho}{P}\bigg)\cdot\sum_{\tau=0}^t\big[|\la\wb_{1,r}^{(\tau)},\vb'\ra| +|\la\wb_{1,r}^{(\tau)},\ub'\ra| \big]\le \tilde O\bigg(\frac{t\eta\rho}{P}\bigg)\cdot|\la\wb_{1,r}^{(t+1)},\vb\ra| =o\big( \log^2(n)\cdot |\la\wb_{1,r}^{(t+1)},\vb\ra|\big)\notag\\
&O\bigg(\frac{\eta}{P}\bigg)\cdot\sum_{\tau=0}^t\big[|\la\wb_{1,r}^{(\tau)},\vb\ra| +|\la\wb_{1,r}^{(\tau)},\ub\ra| \big] \le \tilde O\bigg(\frac{t\eta}{P}\bigg)\cdot|\la\wb_{1,r}^{(t+1)},\vb\ra| =o\big( \log^2(n)\cdot |\la\wb_{1,r}^{(t+1)},\vb\ra|\big).
\end{align}
Putting the above results together, we can verify that
\begin{align*}
|\la\wb_{1,r}^{(t+1)},\ub\ra| = o\big( \log^2(n)\cdot |\la\wb_{1,r}^{(t+1)},\vb\ra|\big).
\end{align*}

We will then verify the Hypothesis for $\la\wb_{1,r}^{(t+1)},\vb'\ra$. By its update rule, Lemma \ref{lemma:feature_learning_coefficients_v'_mixup}, and Lemma \ref{lemma:update_sum_noise}, we have
\begin{align*}
|\la\wb_{1,r}^{(t+1)},\vb'\ra| &\le \big[1 + \eta\gamma_1^{(t)}(\vb',\vb')\big]\cdot|\la\wb_{1,r}^{(t)},\vb'\ra| + \eta\big|\gamma_1^{(t)}(\vb,\vb')\cdot\la\wb_{1,r}^{(t)},\vb\ra\big| + \eta\big|\gamma_1^{(t)}(\ub,\vb')\cdot\la\wb_{1,r}^{(t)},\ub\ra\big| \notag\\
&\qquad + \eta\big|\gamma_1^{(t)}(\ub',\vb')\cdot\la\wb_{1,r}^{(t)}, \ub'\ra\big| + \bigg|\sum_{i=1}^n\sum_{p\in[P]}\eta\gamma_1^{(t)}(\bxi_i^{(p)},\vb')\cdot\la\wb_{1,r}^{(t)},\bxi_i^{(p)}\ra\bigg|\notag\\
&\le2|\la\wb_{1,r}^{(0)},\vb'\ra| + \eta\sum_{\tau=0}^{t}\big|\gamma_1^{(\tau)}(\vb,\vb')\cdot\la\wb_{1,r}^{(\tau)},\vb\ra\big| + \eta\sum_{\tau=0}^{t}\big|\gamma_1^{(\tau)}(\ub,\vb')\cdot\la\wb_{1,r}^{(\tau)},\ub\ra\big| \notag\\
&\qquad + \eta\sum_{\tau=0}^{t}\big|\gamma_1^{(\tau)}(\ub',\vb')\cdot\la\wb_{1,r}^{(\tau)}, \ub'\ra\big| + \sum_{\tau=0}^{t}\bigg|\sum_{i=1}^n\sum_{p\in[P]}\eta\gamma_1^{(\tau)}(\bxi_i^{(p)},\vb')\cdot\la\wb_{1,r}^{(\tau)},\bxi_i^{(p)}\ra\bigg|\notag\\
&\le O\big(\log(n)\sigma_0\big) + O\bigg(\frac{\eta\rho}{P}+\frac{t\rho\eta^2 d\sigma_p^2}{Pn}\bigg)\cdot\sum_{\tau=0}^t|\la\wb_{1,r}^{(\tau)},\vb\ra| + O\bigg(\frac{\eta\zeta\rho}{P}+\frac{t\eta^2\rho d\sigma_p^2}{Pn}\bigg)\cdot\sum_{\tau=0}^t|\la\wb_{1,r}^{(\tau)},\ub\ra|\notag\\
&\qquad + O\bigg(\frac{\eta\zeta\rho^2}{P}+\frac{t\rho^2\eta^2 d\sigma_p^2}{Pn}\bigg)\cdot\sum_{\tau=0}^t|\la\wb_{1,r}^{(\tau)},\ub'\ra| + O\bigg(\frac{t\eta\rho d^{1/2}\sigma_0\sigma_p}{P^{1/2}n^{1/2}}\bigg)\notag\\
&\qquad+ O\bigg(\frac{\eta\rho\zeta}{Pn}+\frac{t\eta^2\zeta\rho d\sigma_p^2}{n^2P}\bigg)\cdot\sum_{\tau=0}^t\sum_{s=1}^n\sum_{p\in[P]}|\la\wb_{1,r}^{(\tau)},\bxi_s^{(p)}\ra|.
\end{align*}
Then by \eqref{eq:upperbound_sum_aboslute_noise} and using the fact that $t\eta=O(\polylog(n))$ and $n=\omega(d\sigma_p^2)$, we can finally get
\begin{align*}
|\la\wb_{1,r}^{(t+1)},\vb'\ra| &\le O\big(\log(n)\sigma_0\big) + O\bigg(\frac{\eta\rho}{P}\bigg)\cdot\sum_{\tau=0}^t|\la\wb_{1,r}^{(\tau)},\vb\ra| + O\bigg(\frac{\eta\rho}{P}\bigg)\cdot\sum_{\tau=0}^t|\la\wb_{1,r}^{(\tau)},\ub\ra| \notag\\
&\qquad  + O\bigg(\frac{\eta \rho^2}{P}\bigg)\cdot\sum_{\tau=0}^t|\la\wb_{1,r}^{(\tau)},\ub'\ra|+ O\bigg(\frac{t\eta\rho d^{1/2}\sigma_0\sigma_p}{P^{1/2}n^{1/2}}+t\eta\rho\zeta d^{1/2}\sigma_0\sigma_p\bigg).
\end{align*}
Then applying \eqref{eq:characterize1}, \eqref{eq:characterize2}, and \eqref{eq:characterize3}, we can also verify that 
\begin{align*}
|\la\wb_{1,r}^{(t+1)},\vb'\ra| = o\big(\log^2(n)\cdot|\la\wb_{1,r}^{(t+1)},\vb\ra|\big).
\end{align*}
The using exactly the same proof, we are also able to verify that 
\begin{align*}
|\la\wb_{1,r}^{(t+1)},\ub'\ra| = o\big(\log^2(n)\cdot|\la\wb_{1,r}^{(t+1)},\vb\ra|\big).
\end{align*}

Lastly, we will prove that 
\begin{align}\label{eq:verification_upperbound_allnoise_v}
\bigg|\sum_{i=1}^n\sum_{p\in[P]}\gamma_1^{(t)}(\bxi_i^{(p)},\vb)\cdot\la\wb_{1,r}^{(t)},\bxi_i^{(p)}\ra\bigg| \le c\cdot |\la\wb_{1,r}^{(t)}, \vb\ra|
\end{align}
for some sufficiently small constant $c$ and all $t=O\big(\polylog(n)/\eta\big)$. This can be proved by the combination of Lemma \ref{lemma:update_sum_noise}, \eqref{eq:upperbound_sum_aboslute_noise}, and our previous characterizations  \eqref{eq:characterize1}, \eqref{eq:characterize2}, \eqref{eq:characterize3}. In particular, using the fact that $|\la\wb_{1,r}^{(t)}, \vb\ra|=\Omega(\sigma_0)$,
we have
\begin{align}\label{eq:bound_sum_noise_v}
&\bigg|\sum_{i=1}^n\sum_{p\in[P]}\gamma_1^{(t)}(\bxi_i^{(p)},\vb)\cdot\la\wb_{1,r}^{(t)},\bxi_i^{(p)}\ra\bigg|\notag\\
&\le O\bigg(\frac{d^{1/2}\sigma_0\sigma_p}{P^{1/2}n^{1/2}}\bigg) +O\bigg(\frac{\zeta}{Pn}\bigg)\cdot\sum_{s=1}^n\sum_{p\in[P]}|\la\wb_{1,r}^{(t)},\bxi_s^{(p)}\ra|+ O\bigg(\frac{\eta d\sigma_p^2}{Pn}\bigg)\cdot\sum_{\tau=0}^{t-1}\big[|\la\wb_{1,r}^{(\tau)},\vb\ra|+|\la\wb_{1,r}^{(\tau)},\ub\ra|\big] \notag\\
&\qquad+ O\bigg(\frac{\eta\rho d\sigma_p^2}{Pn}\bigg)\cdot\sum_{\tau=0}^{t-1}\big[|\la\wb_{1,r}^{(\tau)},\vb'\ra|+|\la\wb_{1,r}^{(\tau)},\ub'\ra|\big]+ O\bigg(\frac{\eta\zeta d\sigma_p^2}{n^2P}\bigg)\cdot\sum_{\tau=0}^{t-1}\sum_{p\in[P]}\sum_{s=1}^n |\la\wb_{1,r}^{(\tau)},\bxi_s^{(p)}\ra|\notag\\
& \le O\bigg(\frac{d^{1/2}\sigma_0\sigma_p}{P^{1/2}n^{1/2}}+\zeta d^{1/2}\sigma_0\sigma_p\bigg) + O\bigg(\frac{\eta d\sigma_p^2}{Pn}\bigg)\cdot\sum_{\tau=0}^{t-1}\big[|\la\wb_{1,r}^{(\tau)},\vb\ra|+|\la\wb_{1,r}^{(\tau)},\ub\ra|\big] \notag\\
&\qquad+ O\bigg(\frac{\eta\rho d\sigma_p^2}{Pn}\bigg)\cdot\sum_{\tau=0}^{t-1}\big[|\la\wb_{1,r}^{(\tau)},\vb'\ra|+|\la\wb_{1,r}^{(\tau)},\ub'\ra|\big]\notag\\
& \le O\bigg(\frac{d^{1/2}\sigma_0\sigma_p}{P^{1/2}n^{1/2}}+\zeta d^{1/2}\sigma_0\sigma_p\bigg) +  O\bigg(\frac{d\sigma_p^2}{Pn}\bigg)\cdot |\la\wb_{1,r}^{(t)},\vb\ra|\notag\\
&\le O\bigg(\bigg(\frac{d^{1/2}\sigma_p}{P^{1/2}n^{1/2}}+\zeta d^{1/2}\sigma_p+ \frac{d\sigma_p^2}{Pn}\bigg)\cdot |\la\wb_{1,r}^{(t)},\vb\ra|\bigg).
\end{align}
Then using the facts that $\zeta = o(d^{-1/2}\sigma_p^{-1})$ and $d\sigma_p^2 = o(n)$, we are able to complete the proof of \eqref{eq:verification_upperbound_allnoise_v}.

\end{proof}

\begin{lemma}\label{lemma:closeness_gamma_v_gamma_u}
Assume $\max_{k\in[2], (i,j)\in\cS} |F_k(\Wb^{(t)}; \xb_{i,j})|\le \zeta$ for some $\zeta =o\big(d^{-1/2}\sigma_p^{-1}\big)$. Then for any $t=O\big(\polylog(n)/\eta\big)$ that satisfies this condition, we have with probability at least $1-1/\poly(n)$, 
\begin{align*}
|\gamma_1^{(t)}(\vb,\vb) - \gamma_2^{(t)}(\ub,\ub)|\le o\bigg(\frac{1}{\polylog(n)}\bigg).
\end{align*}
\end{lemma}
\begin{proof}
Recall $\gamma_1^{(t)}(\vb,\vb)$, we have
\begin{align*}
\gamma_1^{(t)}(\vb, \vb) &=  \frac{1}{n^2}\sum_{i,j\in[n]}\ell_{1,(i,j)}^{(t)} \sum_{p\in[P]} [\theta_{i,j}^{(p)}(\vb)]^2\notag\\
& = \underbrace{\frac{1}{n^2}\sum_{i\in \cS_0^+\text{ or }j\in\cS_0^+}\ell_{1,(i,j)}^{(t)} \sum_{p\in[P]} [\theta_{i,j}^{(p)}(\vb)]^2}_{I_1} + \underbrace{\frac{1}{n^2}\sum_{i\not\in \cS_0^+\text{ and }j\not\in\cS_0^+}\ell_{1,(i,j)}^{(t)} \sum_{p\in[P]} [\theta_{i,j}^{(p)}(\vb)]^2}_{I_2}.  
\end{align*}
Regarding $I_2$, using the similar proof in Lemma \ref{lemma:feature_learning_coefficients_v_mixup}, we can obtain that $I_2 = o\big(1/\polylog(n)\big)$. For $I_1$, using the condition that $\max_{k\in[2],(i,j)\in\cS}|F_k(\Wb^{(t)};\xb_{i,j})|\le \zeta$, we have
\begin{align*}
I_1 = \frac{1}{n^2}\sum_{i\in \cS_0^+\text{ or }j\in\cS_0^+}l_{1,(i,j)} \sum_{p\in[P]} [\theta_{i,j}^{(p)}(\vb)]^2 \pm O(\zeta),
\end{align*}
where $l_{1,(i,j)}\in\{0.5,-0.5,0.5-\lambda,\lambda-0.5\}$ denotes the loss derivative of data $(\xb_{i,j},y_{i,j})$ when its neural network output is forced to be zero. To this end, using the similar decomposition for $\gamma_2^{(t)}(\ub, \ub)$ and noting $\zeta = o(1/\polylog(n))$, we can obtain
\begin{align}\label{eq:difference_gamma_v_gamma_u}
|\gamma_1^{(t)}(\vb, \vb)-\gamma_2^{(t)}(\ub, \ub)|&\le \bigg|\frac{1}{n^2}\sum_{i\in \cS_0^+\text{ or }j\in\cS_0^+}l_{1,(i,j)} \sum_{p\in[P]} [\theta_{i,j}^{(p)}(\vb)]^2-\frac{1}{n^2}\sum_{i\in \cS_0^-\text{ or }j\in\cS_0^-}l_{2,(i,j)} \sum_{p\in[P]} [\theta_{i,j}^{(p)}(\ub)]^2\bigg|\notag\\
&\qquad+ o\big(1/\polylog(n)\big).
\end{align}
Moreover, for any $i\in\cS_0^+$, note that
\begin{align*}
\sum_{j\in[n]}l_{1,(i,j)}\sum_{p\in[P]}[\Theta_{i,j}^{(p)}(\vb)]^2 &= \sum_{j\in[n]}l_{1,(i,j)}\sum_{p\in\cP^*_{i,j}(\vb)}[\Theta_{i,j}^{(p)}(\vb)]^2 + \sum_{j\in[n]}l_{1,(i,j)}\sum_{p\not\in\cP^*_{i,j}(\vb)}[\Theta_{i,j}^{(p)}(\vb)]^2\notag\\
&=\ell_{1,(i,i)}+\sum_{j\neq i}l_{1,(i,j)}\cdot z_{i,j}^2\pm o\big(n/\polylog(n)\big),
\end{align*}
where $z_{i,j}=(1-\lambda)^2$ if $j\in\cS_1^+\cup\cS_0^-\cup\cS_1^-$ and
\begin{align*}
z_{i,j}=\begin{cases}
1& \text{with probability } 1/P;\\
(1-\lambda)^2+\lambda^2& \text{with probability } (P-1)/P,
\end{cases}
\end{align*}
if $j\in\cS_0^+$.
Consequently, applying Hoeffeding's inequality regarding the random variable $z_{i,j}$ (when $j\in\cS_0^+$), we have with probability at least $1-1/\poly(n)$,
\begin{align*}
&\frac{1}{n^2}\sum_{i\in\cS_0^+,j\not\in\cS_0^+}l_{1,(i,j)} \sum_{p\in[P]} [\theta_{i,j}^{(p)}(\vb)]^2 = \frac{(1-\lambda)^2}{n^2}\sum_{i\in\cS_0^+,j\in\cS_1^+\cup\cS_0^-\cup\cS_1^-}l_{1,(i,j)}\pm o\bigg(\frac{1}{\polylog(n)}\bigg)\notag\\
&\frac{1}{n^2}\sum_{i\in\cS_0^+,j\in\cS_0^+}l_{1,(i,j)} \sum_{p\in[P]} [\theta_{i,j}^{(p)}(\vb)]^2 = \frac{\ell_{1,(i,i)}}{n^2} + \frac{1+[(1-\lambda)^2+\lambda^2](P-1)}{Pn^2}\cdot\sum_{i\in\cS_0^+,j\in\cS_0^+, j\neq i}l_{1,(i,j)} \pm o\bigg(\frac{1}{\polylog(n)}\bigg).
\end{align*}
Similarly, we can also obtain
\begin{align*}
\frac{1}{n^2}\sum_{ i\not\in\cS_0^+,j\in\cS_0^+}l_{1,(i,j)} \sum_{p\in[P]} [\theta_{i,j}^{(p)} = \frac{(1-\lambda)^2}{n^2}\sum_{i\in\cS_1^+\cup\cS_0^-\cup\cS_1^-,j\in\cS_0^+}l_{1,(i,j)}\pm o\big(1/\polylog(n)\big)
\end{align*}
Therefore, combining the above results, we can get
\begin{align*}
&\frac{1}{n^2}\sum_{i\in \cS_0^+\text{ or }j\in\cS_0^+}l_{1,(i,j)} \sum_{p\in[P]} [\theta_{i,j}^{(p)}(\vb)]^2 \notag\\
&=  \frac{1}{n^2}\sum_{i\in\cS_0^+,j\not\in\cS_0^+}l_{1,(i,j)} \sum_{p\in[P]} [\theta_{i,j}^{(p)}(\vb)]^2 + \frac{1}{n^2}\sum_{i\in\cS_0^+,j\in\cS_0^+}l_{1,(i,j)} \sum_{p\in[P]} [\theta_{i,j}^{(p)}(\vb)]^2 + \frac{1}{n^2}\sum_{ i\not\in\cS_0^+,j\in\cS_0^+}l_{1,(i,j)} \sum_{p\in[P]} [\theta_{i,j}^{(p)}(\vb)]^2\notag\\
& = \frac{(1-\lambda)^2}{n^2}\sum_{i\in\cS_0^+,j\not\in\cS_0^+ \text{ or } i\not\in\cS_0^+,j\in\cS_0^+ }l_{1,(i,j)} + \frac{\ell_{1,(i,i)}}{n^2} + \frac{1+[(1-\lambda)^2+\lambda^2](P-1)}{Pn^2}\cdot\sum_{i\in\cS_0^+,j\in\cS_0^+, j\neq i}l_{1,(i,j)} \notag\\
&\qquad \pm o\bigg(\frac{1}{\polylog(n)}\bigg).
\end{align*}
Similarly, we can get
\begin{align*}
&\frac{1}{n^2}\sum_{i\in \cS_0^-\text{ or }j\in\cS_0^-}l_{1,(i,j)} \sum_{p\in[P]} [\theta_{i,j}^{(p)}(\ub)]^2 \notag\\
&=  \frac{1}{n^2}\sum_{i\in\cS_0^-,j\not\in\cS_0^-}l_{1,(i,j)} \sum_{p\in[P]} [\theta_{i,j}^{(p)}(\ub)]^2 + \frac{1}{n^2}\sum_{i\in\cS_0^-,j\in\cS_0^-}l_{1,(i,j)} \sum_{p\in[P]} [\theta_{i,j}^{(p)}(\ub)]^2 + \frac{1}{n^2}\sum_{ i\not\in\cS_0^-,j\in\cS_0^-}l_{1,(i,j)} \sum_{p\in[P]} [\theta_{i,j}^{(p)}(\ub)]^2\notag\\
& = \frac{(1-\lambda)^2}{n^2}\sum_{i\in\cS_0^-,j\not\in\cS_0- \text{ or } i\not\in\cS_0^-,j\in\cS_0^- }l_{1,(i,j)} + \frac{\ell_{1,(i,i)}}{n^2} + \frac{1+[(1-\lambda)^2+\lambda^2](P-1)}{Pn^2}\cdot\sum_{i\in\cS_0^-,j\in\cS_0^-, j\neq i}l_{1,(i,j)} \notag\\
&\qquad \pm o\bigg(\frac{1}{\polylog(n)}\bigg).
\end{align*}
Then note that the positive and negative data are generated with equal probability, we have $|\cS_0^-|$ and $|\cS_0^+|$ are different by at most $o\big(1/\polylog(n)\big)$, therefore, it is easy to get that
\begin{align*}
\bigg|\frac{1}{n^2}\sum_{i\in \cS_0^+\text{ or }j\in\cS_0^+}l_{1,(i,j)} \sum_{p\in[P]} [\theta_{i,j}^{(p)}(\vb)]^2-\frac{1}{n^2}\sum_{i\in \cS_0^-\text{ or }j\in\cS_0^-}l_{2,(i,j)} \sum_{p\in[P]} [\theta_{i,j}^{(p)}(\ub)]^2\bigg|\le o\big(1/\polylog(n)\big).
\end{align*}
Plugging the above inequality into \eqref{eq:difference_gamma_v_gamma_u} we can conclude that
\begin{align*}
|\gamma_1^{(t)}(\vb, \vb)-\gamma_2^{(t)}(\ub, \ub)|\le o\bigg(\frac{1}{\polylog(n)}\bigg).
\end{align*}
This completes the proof.
\end{proof}

Finally, we state the outcome of noise learning, common feature learning, and rare feature learning in the following Lemma.
\begin{lemma}
Let $\zeta$ be a preset quantity satisfying $\zeta = [\omega(d\sigma_p^2/(Pn)),o(d^{-1/2}\sigma_p^{-1})]$ and $T$ be the smallest iteration number such that $\max_{k\in[2], (i,j)\in\cS} |F_k(\Wb^{(T)}; \xb_{i,j})|\ge \zeta/2$, then with probability at least $1-1/\poly(n)$, it holds that
\begin{align*}
&\max_{r}|\la\wb_{1,r}^{(T)},\vb\ra|, \max_{r}|\la\wb_{2,r}^{(T)},\ub\ra| = \Omega\bigg(\frac{\zeta^{1/2}}{m^{1/2}}\bigg),\quad \max_{r}|\la\wb_{1,r}^{(T)},\vb'\ra|, \max_{r}|\la\wb_{2,r}^{(T)},\ub'\ra| =\Omega\bigg(\frac{\rho\zeta^{1/2}}{P m^{1/2}}\bigg)\notag\\
&\max_{r}|\la\wb_{2,r}^{(T)},\vb\ra|, \max_{r}|\la\wb_{1,r}^{(T)},\ub\ra| = \tilde O(\zeta^{3/2}),\quad \max_{r}|\la\wb_{2,r}^{(T)},\vb'\ra|, \max_{r}|\la\wb_{1,r}^{(T)},\ub'\ra| =\tilde O(\zeta^{3/2})\notag\\
\end{align*}
\end{lemma}
\begin{proof}

We will only prove the results for the inner products $\la\wb_{1,r}^{(t)},\vb\ra$, 
$\la\wb_{2,r}^{(t)},\ub\ra$,
$\la\wb_{1,r}^{(t)},\vb'\ra$, $\la\wb_{1,r}^{(t)},\ub\ra$, and $\la\wb_{1,r}^{(t)},\ub'\ra$, as the proof for the remaining  inner products will be exactly the same.

We first recall the update of $\la\wb_{1,r},\vb'\ra$:
\begin{align*}
\la\wb_{1,r}^{(t+1)},\vb'\ra &= \big[1 + \eta\gamma_1^{(t)}(\vb',\vb')\big]\cdot\la\wb_{1,r}^{(t)},\vb'\ra + \eta\gamma_1^{(t)}(\vb,\vb')\cdot\la\wb_{1,r}^{(t)},\vb\ra + \eta\gamma_1^{(t)}(\ub,\vb')\cdot\la\wb_{1,r}^{(t)},\ub\ra \notag\\
&\qquad + \eta\gamma_1^{(t)}(\ub',\vb')\cdot\la\wb_{1,r}^{(t)}, \ub'\ra + \sum_{i=1}^n\sum_{p\in[P]}\eta\gamma_1^{(t)}(\bxi_i^{(p)},\vb')\cdot\la\wb_{1,r}^{(t)},\bxi_i^{(p)}\ra.
\end{align*}
The using Lemma \ref{lemma:feature_learning_coefficients_v'_mixup} and the similar proof of Lemma \ref{lemma:dominance_v}, we can get
\begin{align*}
&\big|\gamma_1^{(t)}(\ub,\vb')\cdot\la\wb_{1,r}^{(t)},\ub\ra\big| = O\bigg(\frac{\zeta\rho}{P}\bigg)\cdot |\la\wb_{1,r}^{(t)},\ub\ra| = O\bigg(\frac{\zeta\rho\log^2(n)}{P}\bigg)\cdot |\la\wb_{1,r}^{(t)},\vb\ra|\notag\\
&\big|\gamma_1^{(t)}(\ub',\vb')\cdot\la\wb_{1,r}^{(t)},\ub\ra\big| = O\bigg(\frac{\zeta\rho^2}{P}\bigg)\cdot |\la\wb_{1,r}^{(t)},\ub\ra| = O\bigg(\frac{\zeta\rho^2\log^2(n)}{P}\bigg)\cdot |\la\wb_{1,r}^{(t)},\vb\ra|\\
&\bigg|\sum_{i=1}^n\sum_{p\in[P]}\gamma_1^{(t)}(\bxi_i^{(p)},\vb')\cdot\la\wb_{1,r}^{(t)},\bxi_i^{(p)}\ra\bigg|\le O\bigg(\bigg(\frac{\rho d^{1/2}\sigma_p}{P^{1/2}n^{1/2}}+\rho\zeta d^{1/2}\sigma_p\bigg)\cdot |\la\wb_{1,r}^{(t)},\vb\ra|\bigg).
\end{align*}
Therefore, noting that we have assumed $d\sigma_p=o(n/P)$ and $\zeta = o\big(\frac{1}{Pd^{1/2}\sigma_p}\big)$, 
\begin{align*}
\big|\gamma_1^{(t)}(\ub,\vb')\cdot\la\wb_{1,r}^{(t)},\ub\ra\big|, \big|\gamma_1^{(t)}(\ub',\vb')\cdot\la\wb_{1,r}^{(t)}, \ub'\ra\big|,\bigg|\sum_{i=1}^n\sum_{p\in[P]}\gamma_1^{(t)}(\bxi_i^{(p)},\vb')\cdot\la\wb_{1,r}^{(t)},\bxi_i^{(p)}\ra\bigg|\le c\cdot |\gamma_1^{(t)}(\vb,\vb')\cdot\la\wb_{1,r}^{(t)},\vb\ra|
\end{align*}
for some sufficiently small constant $c<0.5$.
Therefore, further applying Lemma \ref{lemma:feature_learning_coefficients_v'_mixup}, we can get that
\begin{align}\label{eq:update_v'_simplified}
\la\wb_{1,r}^{(t+1)},\vb'\ra &= \big[1 + \Theta(\eta\rho)\big]\cdot\la\wb_{1,r}^{(t)},\vb'\ra + \Theta(\eta\rho/P)\cdot\la\wb_{1,r}^{(t)},\vb\ra
\end{align}
Given the above equation, we are able to complete the proof by combining it with Lemma \ref{lemma:dominance_v}:
\begin{align}\label{eq:update_v_proof_v'}
\la\wb_{1,r}^{(t+1)},\vb\ra = \big[1 + \Theta(\eta)\big]\cdot \la\wb_{1,r}^{(t)},\vb\ra.
\end{align}
In particular, given the fact that $|\la\wb_{1,r}^{(0)},\vb\ra|=\Omega(\sigma_0)$,  we can get the following 
\begin{align}\label{eq:bound_v_proof_v'}
|\la\wb_{1,r}^{(T)},\vb\ra| = \Omega\bigg(\frac{\zeta^{1/2}}{m^{1/2}}\bigg)
\end{align}
for some $T = O\big(\frac{\log(\zeta/(m\sigma_0))}{\eta}\big)$. Besides, by Lemma \ref{lemma:closeness_gamma_v_gamma_u} and \eqref{eq:updates_v_tmp}, we have for any $r'\in[m]$,
\begin{align*}
\frac{|\la\wb_{1,r}^{(t+1)},\ub\ra|}{|\la\wb_{2,r'}^{(t+1)},\ub\ra|} &= \frac{|\la\wb_{1,r}^{(t)},\vb\ra|}{|\la\wb_{2,r'}^{(t)},\ub\ra|}\cdot\bigg(\frac{1 + \eta \gamma_1^{(t)}(\vb,\vb)\pm o\big(\eta/\polylog(n)\big)}{1+\eta \gamma_2^{(t)}(\ub,\ub) \pm o\big(\eta/\polylog(n)\big)}\bigg)\notag\\
&= \frac{|\la\wb_{1,r}^{(t)},\vb\ra|}{|\la\wb_{2,r'}^{(t)},\ub\ra|}\cdot \big[1 + \eta \cdot\big(\gamma_1^{(t)}(\vb,\vb) - \gamma_2^{(t)}(\ub,\ub) \big)\pm o\big(\eta/\polylog(n)\big)\big]\notag\\
& = \frac{|\la\wb_{1,r}^{(0)},\vb\ra|}{|\la\wb_{2,r'}^{(0)},\ub\ra|}\cdot \big[1 \pm o\big(\eta/\polylog(n)\big)\big]^t.
\end{align*}
Note that $t \le T= \tilde O(1/\eta)$, we can further get $\frac{|\la\wb_{1,r}^{(t+1)},\ub\ra|}{|\la\wb_{2,r'}^{(t+1)},\ub\ra|}=\Theta(1)\cdot \frac{|\la\wb_{1,r}^{(0)},\vb\ra|}{|\la\wb_{2,r'}^{(0)},\ub\ra|}$. This immediately implies that $\max_r|\la\wb_{2,r}^{(T)},\ub\ra| = \Theta(\max_r|\la\wb_{1,r}^{(T)},\vb\ra|) = \Omega\big(\zeta^{1/2}/m^{1/2}\big)$.

Moreover, \eqref{eq:update_v'_simplified} implies that
\begin{align*}
\la\wb_{1,r}^{(T)},\vb'\ra & = \big[1 + \Theta(\eta\rho)\big]^T\cdot \la\wb_{1,r}^{(0)},\vb'\ra +  \Theta(\eta\rho/P)\cdot\sum_{t=0}^{T-1}\big[1 + \Theta(\eta\rho)\big]^t\cdot \la\wb_{1,r}^{(t)},\vb\ra.
\end{align*}
Further note that $\la\wb_{1,r}^{(t)},\vb\ra$ has the same sign for all $t\le T$ and $[1+\Theta(\eta\rho)]^t=\Theta(1)$ for all $t\le T$, then define $T' = T-\Theta(1/\eta)$, we have
\begin{align*}
|\la\wb_{1,r}^{(T)},\vb'\ra| &= \bigg|\Theta(1)\cdot \la\wb_{1,r}^{(0)},\vb'\ra + \Theta(\eta\rho/P)\cdot\sum_{t=0}^{T-1} \la\wb_{1,r}^{(t)},\vb\ra\bigg|\notag\\
& \ge \Theta(\eta\rho/P)\cdot\bigg|\sum_{t=0}^{T-1} \la\wb_{1,r}^{(t)},\vb\ra\bigg| - \Theta(1)\cdot \big|\la\wb_{1,r}^{(0)},\vb'\ra\big|\notag\\
& \ge \Theta(\eta\rho/P)\cdot\sum_{t=T'}^{T-1} \big|\la\wb_{1,r}^{(t)},\vb\ra\big| - \tilde O(\sigma_0).
\end{align*}
Then by \eqref{eq:update_v_proof_v'} and \eqref{eq:bound_v_proof_v'}, we have for all $t\in[T', T-1]$, it holds that
\begin{align*}
|\la\wb_{1,r}^{(t)},\vb\ra| = \Theta\big(|\la\wb_{1,r}^{(T)},\vb\ra|\big) = \Omega\bigg(\frac{\zeta^{1/2}}{m^{1/2}}\bigg).
\end{align*}
Therefore, we can finally get
\begin{align*}
|\la\wb_{1,r}^{(T)},\vb'\ra| &\ge \Theta\bigg(\frac{(T-T')\eta \rho}{P}\cdot \bigg)\cdot\Omega\bigg(\frac{\zeta^{1/2}}{m^{1/2}}\bigg) - \tilde O(\sigma_0)\notag\\
& = \Omega\bigg(\frac{\rho\zeta^{1/2}}{Pm^{1/2}}\bigg).
\end{align*}



The remaining part is to establish the upper bounds in terms of incorrect feature learning, i.e., $\la\wb_{1,r}^{(T)},\ub\ra$ and $\la\wb_{1,r}^{(T)},\ub'\ra$. Particularly, recall their update forms as follows:
\begin{align*}
\la\wb_{1,r}^{(t+1)},\ub\ra &= \big[1 - \eta\gamma_1^{(t)}(\ub,\ub)\big]\cdot\la\wb_{1,r}^{(t)},\ub\ra + \eta\gamma_1^{(t)}(\vb,\ub)\cdot\la\wb_{1,r}^{(t)},\vb\ra + \eta\gamma_1^{(t)}(\vb',\ub)\cdot\la\wb_{1,r}^{(t)},\vb'\ra \notag\\
&\qquad + \eta\gamma_1^{(t)}(\ub',\ub')\cdot\la\wb_{1,r}^{(t)}, \ub'\ra + \sum_{i=1}^n\sum_{p\in[P]}\eta\gamma_1^{(t)}(\bxi_i^{(p)},\ub)\cdot\la\wb_{1,r}^{(t)},\bxi_i^{(p)}\ra,\notag\\
\la\wb_{1,r}^{(t+1)},\ub'\ra &= \big[1 - \eta\gamma_1^{(t)}(\ub',\ub')\big]\cdot\la\wb_{1,r}^{(t)},\ub'\ra + \eta\gamma_1^{(t)}(\vb,\ub')\cdot\la\wb_{1,r}^{(t)},\vb\ra + \eta\gamma_1^{(t)}(\ub,\ub')\cdot\la\wb_{1,r}^{(t)},\ub\ra \notag\\
&\qquad + \eta\gamma_1^{(t)}(\vb',\ub')\cdot\la\wb_{1,r}^{(t)}, \vb'\ra + \sum_{i=1}^n\sum_{p\in[P]}\eta\gamma_1^{(t)}(\bxi_i^{(p)},\ub')\cdot\la\wb_{1,r}^{(t)},\bxi_i^{(p)}\ra.
\end{align*}
Then by Lemmas \ref{lemma:feature_learning_coefficients_u_incorrect_mixup} and \ref{lemma:feature_learning_coefficients_u'_incorrect_mixup}, we have
\begin{align*}
\max\big\{|\gamma_1^{(t)}(\ub',\ub)|, \gamma_1^{(t)}(\ub,\ub')|\big\}\le\min\big\{\gamma_1^{(t)}(\ub,\ub), \gamma_1^{(t)}(\ub',\ub')\big\}, 
\end{align*}
the above equations further yield
\begin{align*}
&|\la\wb_{1,r}^{(t+1)},\ub\ra| + |\la\wb_{1,r}^{(t+1)},\ub'\ra| \\&\le |\la\wb_{1,r}^{(t)},\ub\ra| + |\la\wb_{1,r}^{(t)},\ub'\ra| + O(\eta\zeta)\cdot |\la\wb_{1,r}^{(t)},\vb\ra| + O(\eta\zeta\rho/P) \cdot |\la\wb_{1,r}^{(t)},\vb'\ra|\notag\\
& \qquad + \bigg|\sum_{i=1}^n\sum_{p\in[P]}\eta\gamma_1^{(t)}(\bxi_i^{(p)},\ub)\cdot\la\wb_{1,r}^{(t)},\bxi_i^{(p)}\ra\bigg| + \bigg|\sum_{i=1}^n\sum_{p\in[P]}\eta\gamma_1^{(t)}(\bxi_i^{(p)},\ub')\cdot\la\wb_{1,r}^{(t)},\bxi_i^{(p)}\ra\bigg|.
\end{align*}
Then using the fact that $T\eta = O(\polylog(n))$, we can further obtain
\begin{align*}
&|\la\wb_{1,r}^{(T)},\ub\ra| + |\la\wb_{1,r}^{(T)},\ub'\ra| \le \tilde O(\sigma_0) + \tilde O(\zeta)\cdot \max_{t\in[T]}|\la\wb_{1,r}^{(t)},\vb\ra| + \tilde O(\zeta \rho/P)\cdot \max_{t\in[T]}|\la\wb_{1,r}^{(t)},\vb'\ra| \notag\\
&\qquad + \sum_{t=0}^{T-1}\bigg[\bigg|\sum_{i=1}^n\sum_{p\in[P]}\eta\gamma_1^{(t)}(\bxi_i^{(p)},\ub)\cdot\la\wb_{1,r}^{(t)},\bxi_i^{(p)}\ra\bigg| + \bigg|\sum_{i=1}^n\sum_{p\in[P]}\eta\gamma_1^{(t)}(\bxi_i^{(p)},\ub')\cdot\la\wb_{1,r}^{(t)},\bxi_i^{(p)}\ra\bigg|\bigg].
\end{align*}
Moreover, following the same procedure of \eqref{eq:bound_sum_noise_v}, we can get
\begin{align*}
&\sum_{t=0}^{T-1}\bigg|\sum_{i=1}^n\sum_{p\in[P]}\eta\gamma_1^{(t)}(\bxi_i^{(p)},\ub)\cdot\la\wb_{1,r}^{(t)},\bxi_i^{(p)}\ra\bigg|, \sum_{t=0}^{T-1}\bigg|\sum_{i=1}^n\sum_{p\in[P]}\eta\gamma_1^{(t)}(\bxi_i^{(p)},\ub')\cdot\la\wb_{1,r}^{(t)},\bxi_i^{(p)}\ra\bigg|\\
&\le \tilde O(\sigma_0) + \tilde O\bigg(\frac{d\sigma_p^2}{Pn}\bigg)\cdot\max_{t\in[T]}\big[|\la\wb_{1,r}^{(t)},\vb\ra|+|\la\wb_{1,r}^{(t)},\ub\ra|+|\la\wb_{1,r}^{(t)},\vb'\ra|+|\la\wb_{1,r}^{(t)},\ub'\ra|\big]
\end{align*}
Finally, using the assumption that $\zeta = \omega(d\sigma_p^2/(Pn))$, we can get that
\begin{align*}
&|\la\wb_{1,r}^{(T)},\ub\ra| + |\la\wb_{1,r}^{(T)},\ub'\ra| \notag\\
&\le \tilde O(\sigma_0) + \tilde O(\zeta)\cdot \Big[\max_{t\in[T]}|\la\wb_{1,r}^{(t)},\vb\ra|+\max_{t\in[T]}|\la\wb_{1,r}^{(t)},\vb'\ra|+\max_{t\in[T]}\big[|\la\wb_{1,r}^{(t)},\ub\ra|+|\la\wb_{1,r}^{(t)},\ub'\ra|\big]\Big].
\end{align*}
Besides, note that the above inequality actually holds for any $T'\le T$, thus
\begin{align*}
&|\la\wb_{1,r}^{(T')},\ub\ra| + |\la\wb_{1,r}^{(T')},\ub'\ra| \notag\\
&\le \tilde O(\sigma_0) + \tilde O(\zeta)\cdot \Big[\max_{t\in[T']}|\la\wb_{1,r}^{(t)},\vb\ra|+\max_{t\in[T']}|\la\wb_{1,r}^{(t)},\vb'\ra|+\max_{t\in[T']}\big[|\la\wb_{1,r}^{(t)},\ub\ra|+|\la\wb_{1,r}^{(t)},\ub'\ra|\big]\Big]\notag\\
&\le\tilde O(\sigma_0) + \tilde O(\zeta)\cdot \Big[\max_{t\in[T]}|\la\wb_{1,r}^{(t)},\vb\ra|+\max_{t\in[T]}|\la\wb_{1,r}^{(t)},\vb'\ra|+\max_{t\in[T]}\big[|\la\wb_{1,r}^{(t)},\ub\ra|+|\la\wb_{1,r}^{(t)},\ub'\ra|\big]\Big].
\end{align*}
This further implies that
\begin{align*} &\max_{t\in[T]}\big[|\la\wb_{1,r}^{(t)},\ub\ra| + |\la\wb_{1,r}^{(t)},\ub'\ra|\big]\notag\\
&\le \tilde O(\sigma_0) + \tilde O(\zeta)\cdot \Big[\max_{t\in[T]}|\la\wb_{1,r}^{(t)},\vb\ra|+\max_{t\in[T]}|\la\wb_{1,r}^{(t)},\vb'\ra|+\max_{t\in[T]}\big[|\la\wb_{1,r}^{(t)},\ub\ra|+|\la\wb_{1,r}^{(t)},\ub'\ra|\big]\Big].
\end{align*}
Then, rearranging terms will readily give the following result:
\begin{align*}
|\la\wb_{1,r}^{(T)},\ub\ra| + |\la\wb_{1,r}^{(T)},\ub'\ra| &\le \max_{t\in[T]}\big[|\la\wb_{1,r}^{(t)},\ub\ra| + |\la\wb_{1,r}^{(t)},\ub'\ra|\big]\notag\\
&\le \tilde O(\sigma_0) + \tilde O(\zeta)\cdot \big[\max_{t\in[T]}\big[|\la\wb_{1,r}^{(t)},\vb\ra|+\max_{t\in[T]}\big[|\la\wb_{1,r}^{(t)},\vb'\ra|\big]\notag\\
&\le \tilde O(\zeta^{3/2}),
\end{align*}
where the last inequality holds since we must have
\begin{align*}
\max_{t\in[T]}|\la\wb_{1,r}^{(t)},\vb\ra|,\max_{t\in[T]}|\la\wb_{1,r}^{(t)},\vb'\ra| = O\big(\log(n)\cdot\zeta^{1/2}\big)
\end{align*}
as otherwise, we cannot have
$\max_{k\in[2], (i,j)\in\cS} |F_k(\Wb^{(T)}; \xb_{i,j})|\le \zeta/2$ for all $t\le T$, which contradicts the condition made in this lemma. This completes the upper bounds of $|\la\wb_{1,r}^{(T)},\ub\ra|$  and $|\la\wb_{1,r}^{(T)},\ub'\ra|$.

\end{proof}

\subsection{Proof of Theorem \ref{thm:Mixup_training}}\label{sec:proof_mixup}
\begin{proof}[Proof of Theorem \ref{thm:Mixup_training}] 

We will evaluate the test error for common feature data and rare feature data separately. In particular, take the positive data $(\xb,1)$ as an example. Then note that the data $\xb$ consists of the common feature $\vb$, we can obtain the following by Lemma \ref{lemma:outcome_Mixup_main}:
\begin{align*}
F_1(\Wb^{(t)};\xb) = \sum_{r=1}^m \sum_{p=1}^P\big(\la\wb_{1,r}^{(t)},\xb^{(p)}\ra\big)^2 \ge \sum_{r=1}^m\sum_{p:\xb^{(p)}=\vb} \big(\la\wb_{1,r}^{(t)},\vb\ra\big)^2=\tilde \Omega(\zeta).
\end{align*}
On the other hand, we can follow the similar proof of Theorem \ref{thm:std_training} to show that $|\la\wb_{k,r}^{(T)},\bzeta\ra|^2=\tilde O(\sigma_p^2n^2)$ with probability at least $1-1/\poly(n)$, then it follows that
\begin{align*}
F_2(\Wb^{(t)};\xb) = \sum_{r=1}^m \sum_{p=1}^P\big(\la\wb_{2,r}^{(t)},\xb^{(p)}\ra\big)^2  \le \tilde O(b\alpha^2\zeta^3) + \tilde O(\sigma_p^2n^2) < F_1(\Wb^{(t)};\xb).
\end{align*}
where we use the fact that $b\alpha^2 = o(1/\polylog(n))$ and $d = \omega(n^3P)$. 
This clearly suggests that
\begin{align*}
\PP_{(\xb, y)\sim \cD_{\mathrm{common}}}[ \argmax_k F_k(\Wb^{(t)},\xb)\neq y] \le \frac{1}{\poly(n)}.
\end{align*}

Then let's move on to the rare feature data. In particular, consider the positive rare feature data $(\xb, 1)$, which contains the rare feature $\vb'$, we have
\begin{align*}
F_1(\Wb^{(t)};\xb) = \sum_{r=1}^m \sum_{p=1}^P\big(\la\wb_{1,r}^{(t)},\xb^{(p)}\ra\big)^2\ge \ge \sum_{r=1}^m\sum_{p:\xb^{(p)}=\vb'} \big(\la\wb_{1,r}^{(t)},\vb\ra\big)^2=\tilde \Omega(\rho^2\zeta).  
\end{align*}
On the other hand, it holds that
\begin{align*}
F_2(\Wb^{(t)};\xb) = \sum_{r=1}^m \sum_{p=1}^P\big(\la\wb_{2,r}^{(t)},\xb^{(p)}\ra\big)^2  \le \tilde O(b\alpha^2\zeta^3) + \tilde O(\sigma_p^2n^2) =o(\rho^2\zeta)< F_1(\Wb^{(t)};\xb),
\end{align*}
where we use the fact that $b\alpha^2\zeta^2 = o(\rho)$ and $d = \omega(n^3P^3/\rho^2)$. Therefore, this implies that 
\begin{align*}
\PP_{(\xb, y)\sim \cD_{\mathrm{rare}}}[ \argmax_k F_k(\Wb^{(t)},\xb)\neq y] \le \frac{1}{\poly(n)}.
\end{align*}
Putting the results for common feature data and rare feature data together, we are able to complete the proof.

\end{proof}

\bibliography{ref}
\bibliographystyle{ims}

\end{document}